%% file: thesis.tex
\documentclass[MS]{macro/neu_msthesis}

\usepackage{physics} 
\usepackage{siunitx}
\usepackage{graphics} 
\usepackage{graphicx} 
\usepackage{epsfig} 
\usepackage{lipsum} 
\usepackage{acronym}
\usepackage{caption}
\usepackage{subcaption}

\usepackage{amsmath, bm} 
\usepackage{amssymb}  
\usepackage{upgreek}
\usepackage{url}
\usepackage{breakurl}
\usepackage[breaklinks]{hyperref}
\usepackage{caption}
\usepackage{subcaption}
\usepackage{pgfplots}
\usepackage{floatrow}
\usepackage{longtable}
\usepackage{booktabs}
\usepackage{tabularx}
\newfloatcommand{capbtabbox}{table}[][\FBwidth]

\title{System Identification of Thrust and Torque Characteristics for a Bipedal Robot with Integrated Propulsion}

\author{Thomas Cahill}

\dept{Mechanical and Industrial Engineering}

\degree{Master of Science}

\degreename{Mechanical Engineering}

\field{Mechanical Engineering}

\submitdate{April 2025}

\numberofmembers{2}

\principaladviser{Dr. Alireza Ramezani}
\firstreader{Dr. Rifat Sipahi}

\chairman{Dr. Yingzi Lin}

\dean{Jacqueline Isaacs}

\input{macro/macro}

\begin{document}

\pdfbookmark[1]{Cover}{cover}

\titlepage

\begin{frontmatter}

\pdfbookmark[1]{Table of Contents}{contents}
\tableofcontents
\listoffigures
\newpage\ssp
\listoftables


\input{tex/acronyms}
\input{tex/acknowledgements.tex}

\input{tex/abstract.tex}

\end{frontmatter}


\pagestyle{headings}

\input{tex/TJ_Intro.tex}

\input{tex/Northeasterns_Harpy_Platform}

\input{tex/EDF_Thrust_Characterization}

\input{tex/Thruster_design_and_incorporation}

\input{tex/Harpy_Motor_Charaterization}

\input{tex/conclusion}

\bibliographystyle{IEEEtran}
\bibliography{thesis}

\end{document}

%% file: macro/macro.tex
\usepackage{amsfonts}			
\usepackage{times}		

\newcommand{\ifno}[1]{}

\usepackage{multirow}
\makeindex
\usepackage{makeidx}
\usepackage{acronym}

\usepackage{url}
\hypersetup{				
pdfauthor = {\authorRef},
pdftitle = {\titleRef},
pdfsubject = {\expandafter{\degreeRef} thesis submitted to Northeastern University},
pdfkeywords = {add keywords here}
pdfcreator = {LaTeX with hyperref package},
pdfproducer = {dvips + ps2pdf}}

%% file: tex/acronyms.tex

\chapter*{List of Acronyms}
\addcontentsline{toc}{chapter}{List of Acronyms}

\begin{acronym}
	
\acro{CAD}{Computer aided design}. 
\label{acro:CAD}Refers to the use of computer software to create, modify, analyze, and optimize detailed designs for physical objects. CAD software allows engineers, architects, and designers to build precise 2D drawings or 3D models, enabling better visualization, faster iteration, and accurate documentation before physical prototypes are made. It is widely used in fields such as mechanical engineering, architecture, aerospace, and manufacturing to improve design quality, support simulation and analysis, and streamline the production process.

\acro{EDF}{Electric ducted fan}.
\label{acro:EDF}
 A propulsion system that uses a high-speed electric motor to spin a multi-blade fan enclosed within a cylindrical duct, generating thrust by accelerating airflow in a controlled and efficient manner. The duct improves aerodynamic performance by reducing turbulence, making EDFs a compact, lightweight, and effective choice for applications requiring precise and powerful thrust, such as drones, robotic platforms, and small aerial vehicles.
  
\acro{ESC}{Electronic speed controller}.
\label{acro:ESC}
An electronic device that regulates the speed, direction, and sometimes the braking of an electric motor by adjusting the power delivered to it. In the context of electric ducted fans and other motor-driven systems, the ESC receives control signals—typically from a flight controller, computer, or remote input—and modulates the motor's voltage and current accordingly to achieve precise speed control. ESCs are essential for providing smooth motor operation, protecting the motor from electrical faults, and enabling responsive and efficient control in applications like drones, electric vehicles, and robotic systems.
    
\acro{PWM}{Pulse width modulation}.
\label{acro:PWM}
A technique used to control the amount of power delivered to an electrical device by rapidly switching a signal on and off at a fixed frequency while varying the proportion of time the signal stays "on" during each cycle. By adjusting this "on" time, known as the duty cycle, PWM effectively controls the average voltage and current supplied to devices like motors, LEDs, and heating elements. In motor control systems, such as those using electronic speed controllers (ESCs), PWM allows precise regulation of motor speed and torque while maintaining high efficiency and reducing energy losses.

\acro{BET}{Blade Element Theory}.
\label{acro:BET}
A modeling approach used to predict the aerodynamic forces on a rotating blade, such as a propeller or rotor. The method divides the blade into many small segments, or "elements," and analyzes each independently to calculate the local lift and drag forces based on airflow conditions. By summing the contributions of all elements along the blade span, the overall thrust and torque generated by the blade can be estimated. Blade Element Theory is widely used because it provides a relatively simple yet effective way to model complex blade aerodynamics, especially in propellers, turbines, and electric ducted fans.

\acro{CFD}{Computational Fluid Dynamics}.
\label{acro:CFD}
The use of numerical methods and algorithms to solve and analyze problems involving fluid flows (like air, water, or gases). In CFD modeling, the fluid domain is divided into a large number of small cells (a mesh), and the governing equations of fluid motion — such as the Navier-Stokes equations — are solved inside each cell to simulate how the fluid behaves.

\end{acronym}

%% file: tex/acknowledgements.tex

\begin{acknowledgements}

This work would not have been possible without the support of everyone at the Silicon Synapse Lab at Northeastern University who assisted me throughout my thesis. I am deeply grateful to my primary adviser, Professor Alireza Ramezani, whose support and guidance has been invaluable to my studies. His guidance and challenges have shaped my research career, and he has given me the incredible opportunity to work on fascinating and innovative projects. I would also like to express my gratitude to my lab mates —Shreyansh Pitroda, Kaushik Venkatesh, Chenghao Wang, Bibek Gupta, and Patty Meza, for their help with hardware, experiments, control design, simulation, and their continuous encouragement. Additionally, I sincerely appreciate Prof. Rifat Sipahi for serving as my mechanical engineering department co-adviser and thesis reader. Lastly, I am profoundly thankful for the unwavering support of my parents, friends, and sisters. They have always pushed me to be better and stood by me every step of the way in all my pursuits.

\end{acknowledgements}

%% file: tex/abstract.tex

\begin{abstract}

Bipedal robots represent a remarkable and sophisticated class of robotics, designed to emulate human form and movement. Their development marks a significant milestone in the field. However, even the most advanced bipedal robots face challenges related to terrain variation, obstacle negotiation, payload management, weight distribution, and recovering from stumbles. These challenges can be mitigated by incorporating thrusters, which enhance stability on uneven terrain, facilitate obstacle avoidance, and improve recovery after stumbling. Harpy is a bipedal robot equipped with six joints and two thrusters, serving as a hardware platform for implementing and testing advanced control algorithms. This thesis focuses on characterizing Harpy’s hardware to improve the system’s overall robustness, controllability, and predictability. It also examines simulation results for predicting thrust in propeller-based mechanisms, the integration of thrusters into the Harpy platform and associated testing, as well as an exploration of motor torque characterization methods and their application to hardware in relation to closed-loop force-based impedance control.

\end{abstract}

%% file: tex/TJ_Intro.tex
\chapter{Introduction}
\label{chap:TJ Intro}

\section{Literature Review}
\label{sec:litreview}

Bipedal robotic systems have long been a central research focus in legged locomotion due to their potential to navigate environments designed for humans. From stair climbing and rubble traversal to upright manipulation and dynamic balancing, the promise of bipedal robots lies in their versatility. However, this promise is countered by persistent challenges in balancing, disturbance rejection, and actuation efficiency, especially during transitions between motion primitives such as walking, jumping, and recovering from perturbations.

Early bipedal platforms like Honda’s ASIMO \cite{honda_what_2018} demonstrated full-body control and stable gait generation, but were limited in energy efficiency and agility. More recent advancements include Boston Dynamics’ Atlas platform \cite{dyanmics_atlas_2024}, which utilizes high-power actuation and reflexive control strategies to perform complex maneuvers including back flips, vaulting, and parkour-like behavior. While these capabilities demonstrate significant progress, they are primarily enabled by highly customized hardware, proprietary actuation, and carefully tuned controllers. In addition, these robots often require external planning support and are not fully able to recover from severe disturbances or falls without arms or significant support infrastructure.

\begin{figure}[H]
    \centering
    \includegraphics[width=0.75\linewidth]{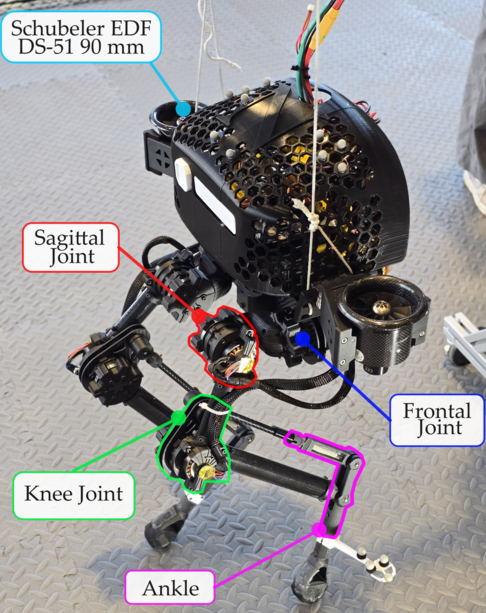}
    \caption{Harpy v2 designed and developed at Northeastern University}
    \label{fig:cover-image}
\end{figure}

Agility Robotics’ Cassie \cite{noauthor_agility_2018} and its successor Digit represent a move toward more scalable and deployable bipedal systems. Cassie features a spring-mass inspired design that allows passive energy recycling and dynamic locomotion over variable terrain. Despite its efficiency, Cassie is limited in roll stabilization and fall recovery due to its narrow footprint and lack of external actuation for body posture regulation. Similarly, ETH Zürich’s ANYmal \cite{eth_anymal_2024}, a quadruped with modular joints and perception-guided control, exemplifies performance in rough environments but lacks aerial capabilities for obstacle negotiation beyond stepping and climbing.

Recognizing the need for more adaptable robots, researchers have begun to explore \textit{thruster-assisted bipedal locomotion}, aiming to augment ground-based mobility with aerial control strategies. The most well-known example of this approach is \textit{LEONARDO} \cite{kim_leonardo_2021}, developed at UC Berkeley. LEONARDO combines legged locomotion with twin propeller arrays mounted on its arms, enabling the robot to modulate body pitch and roll during walking and leaping tasks. In particular, LEONARDO’s use of distributed propeller thrust allows it to stabilize during underactuated flight phases and provides roll correction during mid-air maneuvers. The integration of aerial and terrestrial modalities gives LEONARDO the ability to land more softly, jump higher, and recover more quickly, especially on unstructured or discontinuous terrain.

Another related system is \textit{Salto-1P} \cite{haldane_repetitive_2017}, a monopedal robot developed at Berkeley as well. Salto uses a single leg with a geared brushless DC motor and a lightweight carbon-composite frame. Stability is achieved using a flywheel and tail rotor to modulate orientation mid-air, drawing inspiration from the agile body control observed in animals like lizards and birds. Although not bipedal, Salto demonstrates the value of aerial torque and momentum control for rapid ground-contact transitions and agile hopping.

In contrast to thruster-based systems, alternative methods for enhancing stability in bipedal robots have explored \textit{passive buoyancy} and ultra-lightweight designs. A compelling example is \textit{Ballu} \cite{chae_ballu2_2021}, developed at UCLA. Ballu employs a helium balloon as a buoyant counterweight above a minimalist bipedal leg system, reducing the effective weight of the robot and mitigating fall impacts. While this passive system reduces energy cost and improves safety, it faces challenges with environmental disturbances, such as wind, and is not easily scalable for field use. Nevertheless, it highlights the value of body weight modulation in addressing the limitations of purely ground-based actuation.

Below are several examples of bipedal robots—some previously discussed and others newly introduced—that demonstrate how thrusters and aerial assistance can be integrated into robotic designs to enhance stability and enable more dynamic, agile movement during complex task execution.

\begin{figure}[H]
    \centering
    \includegraphics[width=0.9\linewidth]{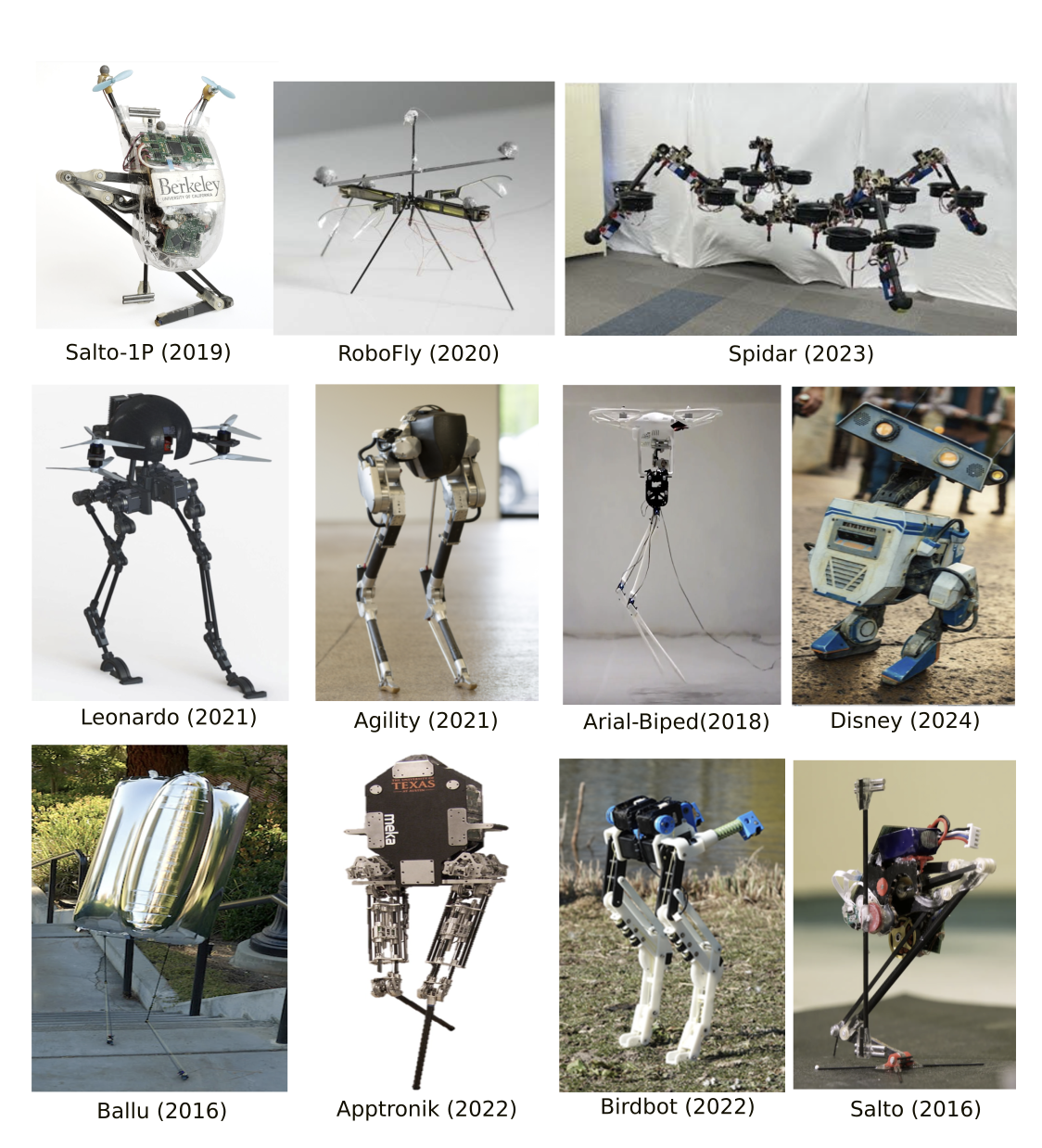}
    \caption{Bipedal robots with thrusters designed by Haldane et al. \cite{haldane_repetitive_2017}, X. Zeng et al. \cite{zeng_kou-iii_2024}, N. Kanagami et al. \cite{kanagamani_pdf_2023}, K. Kim et al. \cite{kim_leonardo_2021}, S. Crowe et al. \cite{noauthor_agility_2018}, H. Chae et al. \cite{chae_ballu2_2021}, R. Grandia et al. \cite{grandia_design_2024},Villani et al. \cite{villani_survey_2018}.A. Abourachid et al. \cite{abourachid_natural_2024}}
    \label{fig:lit-review}
\end{figure}

As legged platforms explore hybrid mobility, the \textit{design and characterization of propulsion systems} becomes critical. Thrusters, especially in the form of EDF's, offer compact, powerful, and modular solutions for generating aerial forces. Unlike open-blade propellers, EDFs encapsulate the fan blades within a cylindrical duct, reducing the risk of damage or injury and improving thrust efficiency via reduced tip vortices. However, integrating EDFs into robots requires careful modeling and testing to ensure stability, thrust-to-weight optimization, and proper force-vector placement relative to the center of mass.

Recent efforts in \textit{propeller thrust modeling} \cite{bryant_development_2020} emphasize combining theoretical, simulation-based, and empirical approaches. Theoretical models typically use momentum theory and BET \cite{ledoux_analysis_2021} to relate thrust to parameters such as RPM, pitch, diameter, and air density. While these models offer generalizable insights, they often fail to capture flow disturbances and real-world aerodynamic losses. Computational tools such as \textit{ANSYS Fluent} enable CFD-based modeling \cite{bryant_development_2020} that simulates fluid flow around rotating blades, capturing complex vortex interactions and pressure gradients. Such simulations, though accurate, are computationally expensive and require fine-tuned boundary conditions and meshing strategies.

To bridge this gap, hybrid models that incorporate \textit{empirical tuning constants} have been proposed \cite{lee_empirical_2020}. For instance, recent studies propose thrust equations augmented with empirically derived coefficients (e.g., $k_1$, $k_2$) to account for nonlinearities in power scaling and blade geometry. These coefficients are typically tuned via data collected from thrust stands equipped with load cells, where variations in blade geometry, RPM, and environmental conditions are systematically explored.

Parallel to thrust modeling is the critical task of \textit{motor torque characterization}, especially for robots relying on force-based control strategies. Understanding a motor’s torque constant $K_t$, defined as the output torque per ampere of input current, is essential for implementing impedance controllers that regulate joint behavior in response to contact forces. Common methods for determining $K_t$ include:

\begin{enumerate}
    \item \textbf{Datasheet-based estimation}, using relationships between motor speed constants (KV rating) and torque constants under ideal assumptions \cite{spong_robot_1989}.
    \item \textbf{Back-EMF measurement}, where the voltage generated by a spinning motor is used to infer $K_t$ via known relationships in brushless DC motor theory \cite{spong_robot_1989}.
    \item \textbf{Torque sensor experiments}, where a calibrated load cell or reaction torque sensor measures direct mechanical output under controlled current inputs \cite{suh_development_2020}.
\end{enumerate}

Each method has trade-offs in precision, cost, and required equipment. Back-EMF methods are susceptible to signal noise and phase ambiguity but are non-invasive and cost-effective. Sensor-based approaches are more accurate and allow closed-loop calibration, but require complex mechanical setups and integration with data acquisition systems.

In summary, the field of bipedal robotics is increasingly moving toward \textit{multimodal platforms} that unify legged agility with aerial control. Integrating thrusters presents both new opportunities and engineering challenges in modeling, actuation, and control. Accurate thrust modeling, motor characterization, and modular mechanical design are foundational elements that support this integration. The development of the Harpy robot contributes to this trajectory by combining empirically validated EDF systems, analytically grounded motor characterization techniques, and control frameworks for future agile hybrid robots.

In addition to the core literature already explored, a range of supplementary studies provide valuable insights into thruster-assisted legged robotics. These works examine a variety of platforms that integrate aerial propulsion to improve locomotion, balance, and maneuverability in complex environments. Several papers focus specifically on the modeling and simulation of thruster performance, including both physics-based hand calculations and high-fidelity software simulations. These methodologies help to predict system behavior under varying conditions, enabling robust design decisions. Furthermore, recent studies on motor constant characterization outline practical approaches for empirically deriving key parameters—such as torque constants and back-EMF coefficients—through both standalone experiments and integrated sensing techniques. Together, these sources contribute to a deeper understanding of how propulsion systems can be effectively modeled, tested, and implemented in dynamic robotic platforms.

\begin{center}
    \textbf{Thruster-Assisted Legged Robots}
\end{center}
\begin{longtable}{p{2cm} p{3.3cm} p{3.3cm} p{3.3cm}}
\toprule
\textbf{Paper} & \textbf{Problem Statement} & \textbf{Proposed Solution} & \textbf{Key Results / Significance} \\
\midrule

Kim et al., 2016 \cite{kim_leonardo_2021} & Numerous mobile robots in various forms specialize in either ground or aerial locomotion, whereas very few robots can perform complex locomotion tasks beyond simple walking and flying, or both. &   Leonardo bridges the gap between two different locomotion regimes of flying and walking using synchronized control of distributed electric thrusters and a pair of multijoint legs. & The mechanical design and synchronized control strategy achieve a unique multimodal locomotion capability that could potentially enable robotic missions and operations that would be difficult for single-modal locomotion robots. \\

  Haldane et al., 2017 \cite{haldane_repetitive_2017} & Mono pedal  robot that  jumps and has directional control on the ground and air. &  Salto-1P uses a small motor and a system of linkages and gears to jump, and  a rotating inertial tail and two little thrusters to stabilize and reorient itself  & Able to perform wall jumps: After jumping off of the ground, Salto-1P can jump off of a wall to get higher. Vertical jumping agility: 1.83 m/s. Maximum jump height 1.25 meters (from crouch to apex) \\

  Fadelli et al., 2020 \cite{fadelli_robofly_2020} &Insect-size robots could have numerous useful applications, for instance, aiding search and rescue (SAR) missions, but size and and weight provide  manufacturing hindrances &   researchers at the University of Washington have recently created RoboFly, a 74-mg flapping-wing robot that can move in the air, on the ground and on water surfaces. Built using fewer components than other micro sized robots.  & The fact that it can fly, walk and drift on water makes RoboFly unique, setting it apart from other insect-size robots. The robot could be far more effective than existing ones in avoiding obstacles, as it can simply switch to a different mode of locomotion (e.g. flying or moving on water if it detects obstacles on the ground).\\

Pitroda et al., 2024 \cite{pitroda_capture_2024} & bipedal robots are still susceptible to falling over and struggle to negotiate rough terrains& The authors developed a capture-point-based controller integrated with a quadratic programming (QP) solver to enable thruster-assisted dynamic bipedal walking on the Harpy robot.  & This work uniquely integrates thruster forces into the capture point control paradigm, introducing the concept of "virtual buoyancy" to modulate walking frequency and stability.\\

Huang et al., 2017 \cite{huang_jet-powered_2017} &  Many robots lack the ability to traverse tough terrains to provide assistance during natural disasters such as hurricanes and earthquakes. & The research team  developed a ducted-fan propulsion system that drives the legs of a bipedal robot, named Jet-HR1 (Jet Humanoid Robot version 1), over broad gaps.  & The effectiveness and practicality are validated by Jet-HR1 stepping over a wide ditch with 370 mm, using almost 80\% of its leg, both in simulation and experiment. \\

Li et al., 2022 \cite{li_design_2022} & Enabling short-distance flight enhances the mobility of humanoid robots in complex environments, such as traversing large obstacles or accessing elevated areas, which is especially valuable for rapid-response missions.&  To address the challenge of maintaining a stable attitude during takeoff under low thrust-to-weight conditions, the robot was designed with a thrust vectoring approach. Its propulsion system is composed of four ducted fans.  & The robot successfully executed takeoff with a thrust-to-weight ratio of 1.17 (17 kg thrust / 20 kg weight), maintaining a stable attitude and reaching a height exceeding 1000 mm.\\

IIOT et al., 2024 \cite{iiot_ironcub_2024} & The iRonCub project aims to overcome the mobility limitations of traditional humanoid robots by integrating jet propulsion to enable seamless transitions between walking and flying, allowing navigation through complex and unstructured environments. &  iRonCub is a humanoid robot equipped with four jet engines—two mounted on its arms and two on a backpack. This configuration enables the robot to perform controlled flight maneuvers and transitions between walking and flying.  & The project focuses on trajectory planning, flight control, and thrust estimation to achieve seamless integration of aerial and terrestrial locomotion proven through rigorous empirical testing. \\

Fadelli et al., 2025 \cite{fadelli_quadrotors_2025} &The KOU-III robot addresses the challenge of enhancing bipedal locomotion by integrating quadrotor assistance, enabling improved stability, agility, and adaptability in complex environments.  &  By incorporating rotors that provide lift and torque, KOU-III can perform dynamic movements such as higher jumps and rapid posture adjustments, inspired by the jumping behavior of the Red-Capped Manakin bird & The successful integration of quadrotor thrust to enhance bipedal jumping height, posture adjustment, and agility—enabling stable locomotion over uneven terrain inspired by bird-like movement dynamics proves the robots effectiveness.\\

\addlinespace
\bottomrule
\end{longtable}

\begin{center}
    \textbf{Thruster modeling papers}
\end{center}

\begin{longtable}{p{2cm} p{3.3cm} p{3.3cm} p{3.3cm}}
\toprule
\textbf{Paper} & \textbf{Problem Statement} & \textbf{Proposed Solution} & \textbf{Key Results / Significance} \\
\midrule
Gunel et al., 2016 \cite{gunel_modeling_2016} & The need for accurate measurement and control of thrust force and torque in micro propellers used in UAVs. &   Designed a PID controller for a basic propeller thrust test system under static thrust conditions.  & Provided a framework for determining micro propeller characteristics, crucial for UAV performance optimization. \\

Lam et al., 2023 \cite{lam_propeller_2023} & Understanding the performance characteristics of the Blue Robotics T200 Thruster. &  Conducted experimental tests to measure thrust and torque under various conditions and methods to test each methods effectiveness to hone in on performance characteristics. & Provided empirical data to aid in the effective application of the T200 Thruster in marine robotics. \\

Kosaet al., 2022 \cite{kosa_experimental_2022} & Developing a system for accurate measurement of UAV propeller parameters, including thrust and current consumption. &  Created a processing circuit based on an ATmega2560 microcontroller to acquire and store data.  & Offered a universal experimental measuring system for UAV propeller performance evaluation. \\

Bryant et al., 2016 \cite{bryant_development_2020} &Creating a propeller thrust model that accounts for factors like crosswinds and vehicle proximity effects. &  Combined experimental measurements with CFD simulations to develop the model. &  Provided a comprehensive model applicable to UAV operations near moving platforms. \\

Simmons et al., 2022 \cite{simmons_efficient_2022} & The need for accurate yet computationally efficient aerodynamic models of variable-pitch propellers in vectored-thrust electric vertical takeoff and landing (eVTOL) aircraft for use in simulation and control design. &  The authors developed a physics-informed surrogate model based on blade element momentum theory (BEMT), validated it against high-fidelity CFD simulations and wind tunnel data, and implemented it for fast performance estimation under varying flight conditions. & The study presents one of the first compact and computationally efficient aerodynamic models tailored specifically for variable-pitch propellers in eVTOL configurations, enabling real-time simulation and model-based control for complex aircraft geometries. \\

Zhang et al., 2018 \cite{zhang_investigation_2018} & The study examines the complex hydrodynamic behavior of ducted propellers operating under oblique flow conditions, which is critical for the design and safe operation of thruster-driven vessels and dynamic positioning (DP) systems. &  The researchers conducted numerical simulations using Reynolds-Averaged Navier-Stokes (RANS) models for quasi-steady analysis and hybrid RANS/LES models for transient computations. They analyzed a marine model thruster subjected to various oblique flow angles, focusing on hydrodynamic loads on propeller blades and the nozzle.  & This work provides a comprehensive analysis of the performance variations of ducted propellers under different oblique flow angles, revealing significant insights into thrust and torque fluctuations due to flow-induced imbalances. \\

Jakubowski et al., 2016 \cite{jakbouski_analysis_2015} & Continued need to increase test methods to empirically solve for thrust produced by RC motor-prop assemblies.  & The paper presents a design of test stand designed for thrust propel-
lers measurements. There are shown results of the thrust for different propel-lers. & Lots of empirical data from the test stand was taken, and analysis indicates that propeller diameter is its most important parameter and in some range thrust is proportional to diameter. \\

Lipovetsky et al., 2009\cite{lipovetsky_linear_2009} & There are many methods to linearize non-linear systems, but can always be improved for efficiency and accuracy based on application.  &  Considered regression models were constructed by nonlinear optimization techniques, yielding stable solutions and good quality of fit, conducted by having a simple structure of linear aggregate data, demonstrated high predictive ability, and suggested a convenient ways to identify the main predictors for modeling. & Linear regression methods discussed proved to be just as accurate for specific use cases than other traditional methods.\\

\addlinespace
\bottomrule
\end{longtable}

\begin{center}
    \textbf{Actuator Motor Constant Characterization Papers}
\end{center}
\begin{longtable}{p{2cm} p{3.3cm} p{3.3cm} p{3.3cm}}
\toprule
\textbf{Paper} & \textbf{Problem Statement} & \textbf{Proposed Solution} & \textbf{Key Results / Significance} \\
\midrule
Lee et al., 2016 \cite{lee_how_2023} & addresses the issue of inconsistent and unclear modeling practices for brushless DC (BLDC) motors, which often stem from vague manufacturer datasheets,  &  This approach provided a standardized set of governing equations and highlighting common errors to improve motor selection and design in lightweight robotic applications.  & The paper presents a unified and validated modeling approach for brushless motors that clarifies datasheet ambiguities, corrects common modeling errors, and enables accurate torque and power predictions for lightweight robotic systems.\\

Lin et al., 2010 \cite{lin_-depth_2010} & This paper addresses the ambiguity in defining torque and back-EMF constants in BLDC and PM AC motors. &  This was accomplished by analyzing their applicability across different motor types, offering a unified framework that accounts for current freewheeling and non-ideal waveforms. & They proved that their methods were successful through methods of empercial testing and modeling. \\

HBK et al., 2025 \cite{hbk_getting_2025} & Eliminate the need for complex test setups when calculating the back emf or inverse torque constant of a torque motor. &  This study introduces a rapid, non-invasive method to identify the back-EMF constant of PM motors by manually rotating the rotor and capturing voltage data with a Gen3i recorder. & In the end, a simple hands-on technique was shown that leverages the high-speed capture capabilities to extract the motor’s back-EMF constant from a single manual twist, making it ideal for quick diagnostics and field testing. \\

Schmidt et al., 2018 \cite{schmidt_practical_2018} & Addressing the lack of accurate motor characterization simulations available without emperically testing motors after fabrication to redefine the model.  & A hands-on methodology for actuator characterization was derived that yields motor models from standalone measurements, enabling accurate simulation of actuator dynamics prior to full system integration & The most critical aspect of this approach is a non-robot-integrated actuator to extract dynamic motor parameters—enabling reusable, system-independent models adaptable across multiple robotic platforms. \\

Colmenares et al., 2020 \cite{colmenares_characterization_2020} & As motors become smaller, they can be harder to predict and characterize. &  This paper tackles the thermal challenges in micro-scale resonant actuators by experimentally identifying thermal parameters and developing a lumped parameter thermal model to predict safe operating conditions under known torque coefficients and impedance control. & These ressearchers accounted for factors like variable winding resistance, bushing friction, and speed-dependent forced convection to accurately predict safe operating limits and optimize thermal and associated torque driven management in miniature robotic systems. \\

Suh et al., 2020 \cite{suh_development_2020} & The need for a cost-effective and adaptable method to measure torque in rotating shafts without relying on expensive multi-axis sensors. & Introduced a simple, single-axis reaction torque sensor that uses a load cell combined with adjustable lever arms to continuously measure applied torque. & The design allows for easy adjustment of measurement range and sensitivity by repositioning the load cell, making it versatile for various applications.\\

Hwang et al., 2018 \cite{hwang_virtual_2018} & The challenge of estimating torque in low-cost RC servo motors, which typically lack built-in torque sensing capabilities.& Developed a virtual torque sensor by deriving mathematical models based on the internal dynamics of RC servo motors, enabling torque estimation without additional hardware. & Provides a non-invasive and cost-effective solution for torque estimation in RC servo motors, facilitating their use in applications requiring torque control.\\

Wen et al., 2020 \cite{wen_investigation_2020} & The challenge of measuring output torque in permanent magnet spherical motors (PMSMs), which have complex structures. &  Proposed a novel method using a microelectromechanical system (MEMS) gyroscope to measure rotor motion acceleration, which is then used to calculate output torque based on rotor dynamics.
 & Offers a feasible and less invasive approach to torque measurement in PMSMs by leveraging MEMS technology. \\
\addlinespace
\bottomrule
\end{longtable}

\section{MS Thesis Objectives}
\label{sec:MStthesisobj}

This thesis specifically aims to develop and characterize a bipedal robotic platform called Harpy which integrates aerial propulsion to overcome traditional limitations in bipedal mobility. Specifically, the objectives of the thesis are:

\begin{itemize}
    \item To integrate and empirically characterize electric ducted fan (EDF) thrusters into a lightweight bipedal robot.
    
    \item To develop a theoretical thrust model for EDF performance and validate it through \textit{ANSYS Fluent} simulations and physical testing.
    
    \item To experimentally characterize the torque constant $K_t$ of custom brushless DC (BLDC) motors using three independent methods: datasheet estimation, back-EMF analysis, and direct torque sensing.
    
    \item To incorporate the resulting torque constant into an impedance-based control framework for more stable and responsive joint actuation.
\end{itemize}

\section{MS Thesis Contributions}
\label{sec:contributions}

This MS thesis contributes to the growing body of research on thruster-assisted locomotion and actuator characterization for agile robots. The thesis specifically contributes to ongoing research activities at the Silicon Synapse lab in the following ways:

\begin{itemize}

 \item \textbf{Thrust Modeling and Empirical Validation:}
 A comprehensive approach was used to design and validate a EDF thrust system, beginning with a theoretical thrust model that incorporates key aerodynamic parameters. This model was validated against both empirical measurements and CFD simulations, showing less than 6\% error—demonstrating its accuracy and reliability. The validated model enables rapid parametric exploration of thrust output under varying design conditions, significantly reducing development time for future configurations. The empirical testing process involved static load measurements and airflow characterization, providing a robust dataset for refining model coefficients.
 
 \item \textbf{Test Platform Development and Robotic Integration:}
 To ensure precise thrust evaluation and safe experimentation, a custom-built test stand was developed, allowing for controlled off-robot validation of the EDF-ESC system. Once performance was confirmed, the thrust module was integrated into the bipedal robot’s frame using rigid carbon fiber mounts, with careful mass distribution to preserve the robot’s center of gravity and dynamic balance. This staged process—from bench-top testing to full system integration—ensured that aerial actuation could be seamlessly incorporated into the robot, enhancing its ability to stabilize and execute dynamic movements during task completion.

    \item \textbf{Torque Constant Characterization:} Conducted a three-pronged investigation into the motor torque constant $K_t$ using (i) manufacturer datasheets, (ii) back-EMF voltage monitoring with oscilloscope measurements, and (iii) a calibrated torque sensor system. The sensor-based method provided the most consistent results, with a final value of $K_t = 0.0311~\text{Nm/A}$.
    
\item \textbf{Comprehensive Literature Review and System Down selection:}
     A detailed literature review was conducted encompassing three key domains: (1) thruster characterization methods, including analytical, empirical, and CFD-based techniques for modeling and validating thrust output in robotic systems; (2) motor characterization approaches, with a focus on extracting constants such as torque and back-EMF coefficients using standalone sensor-based tests, dynamic response analysis, and data-driven parameter estimation; and (3) control strategies used in current bipedal robots, particularly those implementing closed-loop control for gait stability, disturbance rejection, and aerial-assisted locomotion. This review informed the technical direction for the propulsion subsystem, guiding the down selection of EDF units based on thrust-to-weight ratio, electrical efficiency, mass, form factor, and integration potential. Alternative thrust-generation methods, such as cold gas thrusters and axial fans, were also evaluated. The synthesis of this research supports the goal of improving closed-loop stability through intelligent thrust integration, enabling bipedal robots to better respond to external disturbances and perform dynamically complex tasks with higher robustness.

\end{itemize}


%% file: tex/Northeasterns_Harpy_Platform.tex

\chapter{Northeastern's Harpy Platform}
\label{chap: Northeastern's Harpy platform}

This chapter provides an overview of Northeastern's Harpy hardware platform. Harpy consists of two main subsystems namely legged assembly and aerial assembly. The design of the Harpy's legged assembly was done by previous MS and PhD students and preliminary tests were performed by prototyping one leg. This chapter gives an overview of Harpy's legged and aerial assembly.


\section{Harpy Mechanical Overview}

\begin{figure}[H]
        \centering
        \includegraphics[width=0.7\linewidth]{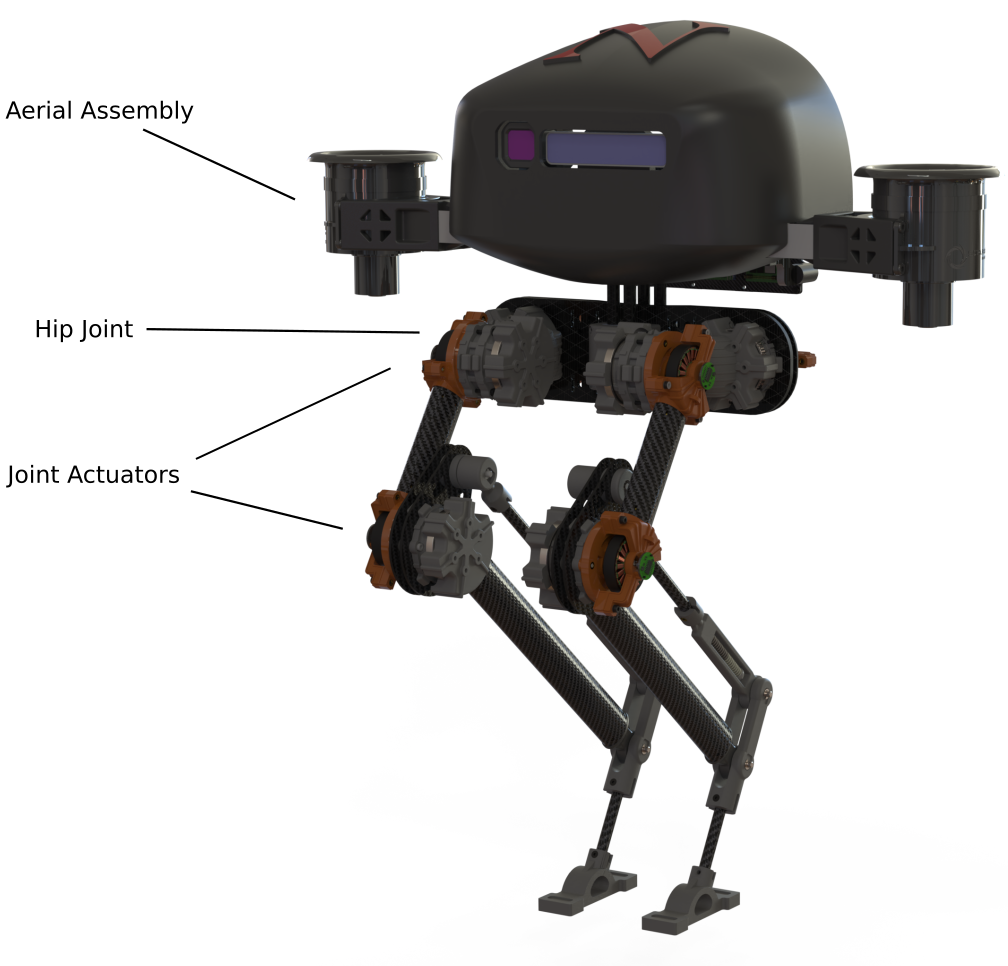}
        \caption{A visual breakdown of Harpy primary subsystems}
        \label{fig:harpy-breakdown}
    \end{figure}

Harpy~\cite{dangol_control_2021,pitroda_capture_2024,pitroda_dynamic_2023,kelly_design_2021,pitroda_enhanced_2024,dangol_feedback_2020,dangol_unilateral_2021,dangol_reduced-order-model-based_2021,sihite_posture_2024,liang_rough-terrain_2021,de_oliveira_thruster-assisted_2020,dhole_feedback_nodate,dangol_hzd-based_2021,dangol_performance_2020,dangol_towards_2020,pitroda_conjugate_2024,pitroda_quadratic_2024} was developed in the Silicon Synapse Lab at Northeastern University. A visual breakdown of its primary subsystems is shown in Fig.~\ref{fig:harpy-breakdown}. The robot stands approximately 30 inches tall, spans 24 inches in width, and has a total mass of approximately $6.5~\text{kg}$. Its mechanical structure is designed to be lightweight, energy-dense, and robust, enabling dynamic motions such as jumping and absorbing impacts while maintaining low inertia near the feet for agile foot placement. Additionally, other research projects at Northeastern University's Silicon Synapse lab including robots Husky~\cite{krishnamurthy_narrow-path_2024,sihite_optimization-free_2021, ramezani_generative_2021,noauthor_progress_nodate,salagame_quadrupedal_2023,krishnamurthy_towards_2023,krishnamurthy_enabling_2024}, Husky-B~\cite{wang_legged_2023}, and M4~\cite{sihite_efficient_2022,sihite_multi-modal_2023,sihite_demonstrating_nodate} have each contributed key technologies and design insights that helped advance the development of the Harpy project. Husky~\cite{krishnamurthy_thruster-assisted_2024,sihite_efficient_2022,salagame_letter_2022,gherold_self-supervised_2024-1} and Husky-B provided foundations in robust bipedal locomotion, control strategies for uneven terrain, and the introduction of aerial assist techniques via the integration of thruster-based stabilization. Lastly, M4~\cite{mandralis_minimum_2023,noauthor_this_nodate,sihite_multi-modal_2023-1,gherold_self-supervised_2024,sihite_dynamic_2024} contributed innovations in lightweight structural design, modular architecture, and high-agility actuation that directly informed Harpy’s final system design.

The leg architecture adopts a series-actuated pantograph configuration, chosen for its favorable mechanical advantage and efficient transmission of actuator torque. Each leg provides three degrees of freedom, achieved through the hip frontal joint (, hip sagittal joint, and knee joint . The thruster assembly is mounted to the upper body using a composite frame that ensures a high strength-to-weight ratio, critical for preserving balance and minimizing overhead mass.

At the core of the robot is the pelvis block, a key structural component that connects both legs and supports the aerial assembly. It plays a crucial role in maintaining balance and absorbing dynamic loads generated during locomotion. Harpy’s pelvis is constructed using two parallel carbon fiber plates spaced 8~mm apart. This configuration enhances transverse rigidity and prevents deflection. The plates are embedded into the frontal actuator assembly, and additional stiffness in the frontal direction is achieved through two oval-section carbon fiber rods, which anchor the central connector to the frontal motor assembly.

The pantograph leg design was selected based on its mechanical efficiency and the ability to avoid singularities and undesirable knee inversion. To achieve the optimal trade-off between weight and rigidity, the linkages use carbon fiber tubes with oval cross-sections. Unlike circular tubes that provide isotropic bending resistance, oval profiles offer enhanced stiffness in the sagittal plane—the primary direction of load during walking and jumping. This geometry reduces structural weight and spatial footprint while increasing directional stiffness.

To ensure robust mechanical coupling, the carbon fiber tubes are joined using a criss-cross fastener pattern that constrains rotation in all directions. Furthermore, the leg actuation incorporates series elastic elements, which provide several performance advantages. These elements enable high-fidelity force control, enhance robustness against external perturbations, and act as low-pass filters to suppress high-frequency disturbances. During cyclic behaviors such as walking, trotting, or jumping, the elastic elements can also store and release mechanical energy, contributing to smoother dynamics and improved energy efficiency.

\section{Custom-Designed Actuators}
\label{sec:actuator}

\begin{figure}[H]
    \centering
    \includegraphics[width=0.75\linewidth]{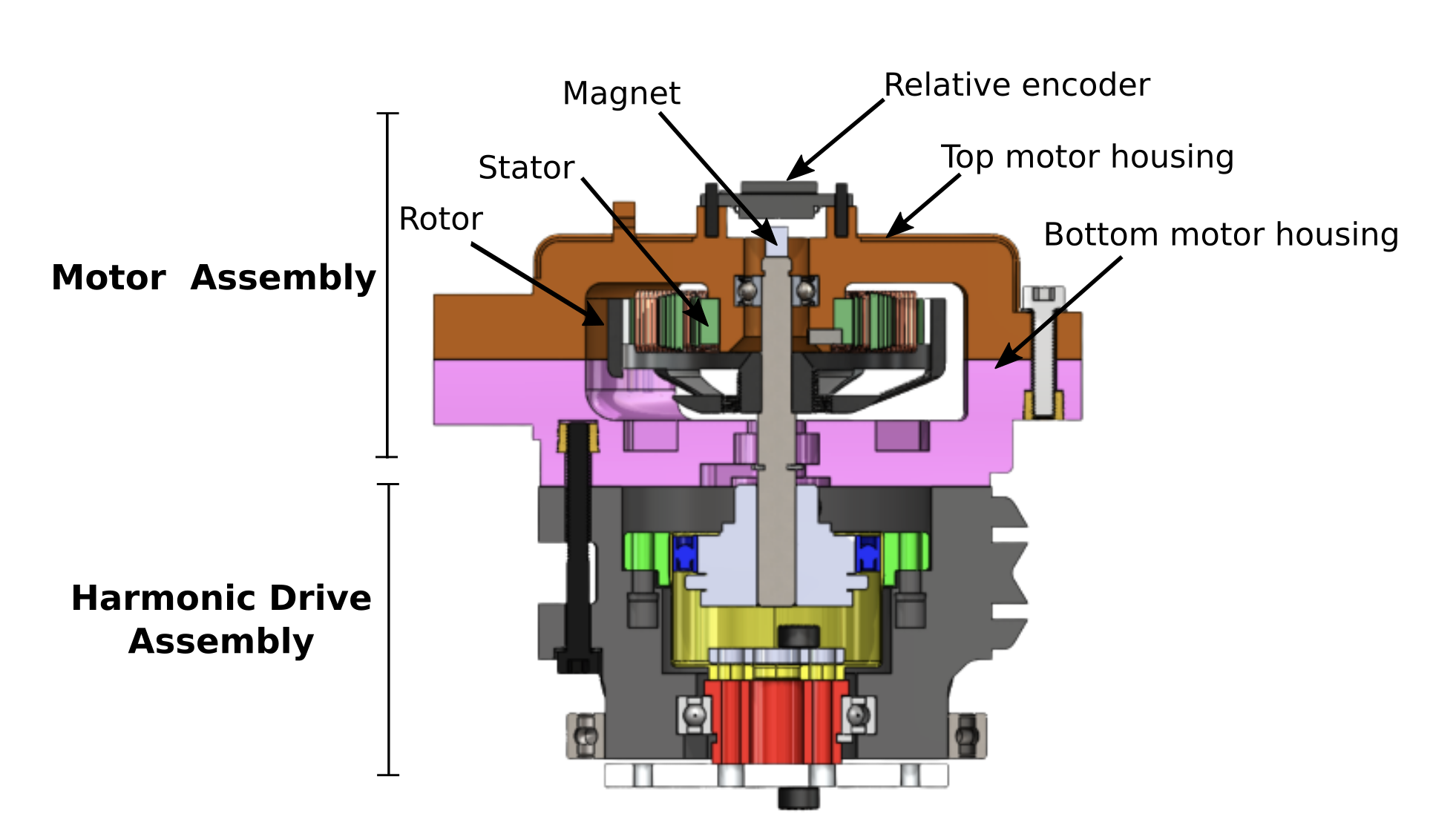}
    \caption{Harpy's Actuator Sectional View}
    \label{fig:enter-label}
\end{figure}

Legged robot joint actuators in general must be extremely power dense with high transparency, high efficiency, and minimal backlash. The importance of the joint lies in its fundamental role in enabling robots to mimic the bio-inspired movement and accomplish a wide variety of tasks with precision and adaptability. The methodology behind selecting components and materials for this robot’s leg joint actuators are discussed in this section. 

In the design of a Harpy's actuator transmission, several critical qualities must be considered, including mechanical transparency, torque density, back-drivability, and low backlash. High transparency actuators allow efficient energy flow from the motor to the output shaft, typically achieved with low gear ratios. 

Torque density is essential for achieving a high overall thrust-to-weight ratio in the robot, while back-drivability ensures better impact absorption and high transparency. Low backlash is crucial to minimize uncertainty in the robot's leg position. Direct drive systems offer high efficiency, precise control, and low backlash but may be limited by the size and weight of the motor. 

Planetary gear systems provide high torque density and can handle significant loads but might suffer from increased friction and complexity. Harmonic drives combine features from planetary gears and flexible components, offering high reduction ratios with minimal backlash and high torque transmission. They also offer modularity in robot actuators due to their interchangeable gear ratios. The selection of the transmission type affects the input torque developed by the motor's electromagnetic torque, which, in turn, determines the output torque for each joint actuator. Considering these factors, a harmonious drive system was chosen to strike a balance between efficiency, torque density, and low backlash for the legged robot's actuator design.

\section{Propulsion Assembly}
\label{sec:prop}

\begin{figure}[H]
    \centering
    \includegraphics[width=1\linewidth]{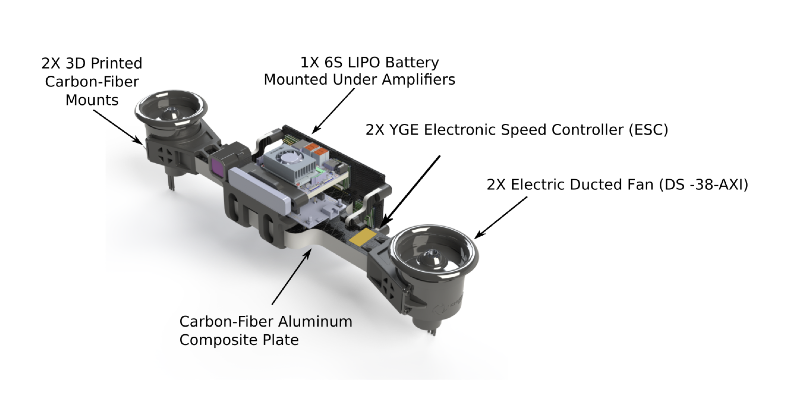}
    \caption{Harpy Thruster EDF ESC 6S LIPO Battery Assembly}
    \label{fig:Propulsion_Assembly}
\end{figure}

Harpy's thrusters as shown in Fig.~\ref{fig:Propulsion_Assembly} are its wings, enabling it to bypass large obstacles and rough terrain in addition to giving the robot additional degrees of freedom to increase its stability. Harpy's thruster assembly consists of Carbonfiber-aluminum composite mount, a Realsense camera T265, Jetson Orin, STM(F439ZI) $+$ NetXshield board, an EDF, an amplifier rack, and IMU. Further exploration and explanation of the Harpy's aerial assembly can be found in Chapter~\ref{chap:prop-hardware-integrate}.

\subsection{Schuebeler Electric Ducted Fans}

The process of selecting the appropriate type and size of thrusters for the robot involves evaluating numerous factors including efficiency, maximum thrust, safety, response time, energy density, weight, diameter, volume, heat emission, and noise. To ensure a thrust-to-weight ratio greater than one, the target total thrust capacity for the thrusters is set at 6 kgf, equivalent to approximately 58 N. Chapter three will delve into detail on deriving a thrust equation as well as CFD models to back up this thrust equation and EDF downselection.

EDFs stand out due to their capacity to generate significantly higher thrust outputs compared to conventional propellers of equivalent blade diameters. This feature is particularly advantageous in reducing the volume occupied by thrusters. Despite the added weight of the duct, the utilization of composite materials enables the creation of lightweight yet robust housing. Another significant advantage for ground operations, which constitute the majority of Harpy's activity, is the enhanced safety level provided by housing the blades within the duct. This design aspect reduces concerns regarding blade tip collisions, contributing to safer navigation in unknown environments and increased human safety in close proximity to the robot.

The efficiency enhancement that ducts bring to propellers is achieved by minimizing energy losses caused by blade tip vortices and converting that energy into additional thrust. The closely spaced relationship between the blade tip and the duct obstructs air passage around the blade tip due to the pressure differential between the upper and lower surfaces of the blade. Introducing a lip at the duct inlet further enhances efficiency by leveraging the Coandă effect \cite{circiu_pdf_2024}, which generates additional thrust at no extra cost to the motor. The Coandă effect is driven by the pressure difference between the ambient air pressure and the lower pressure air moving rapidly above the curved surface. The lip's mechanism for producing thrust is analogous to the generation of lift by an airplane wing according to Bernoulli's principle, where faster-moving air over the wing's top surface creates lower pressure compared to the slower-moving air beneath.

\begin{figure}[h]
\begin{floatrow}
\ffigbox{
  \includegraphics[width=0.42\textwidth]{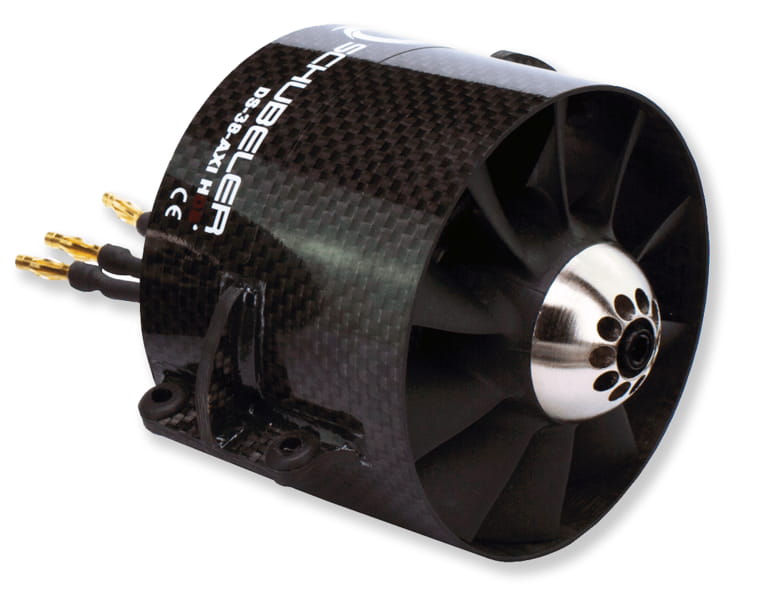}%
}{
  \caption[Schuebeler DS-30 AXI HDS]{Schuebeler DS-30 AXI HDS \cite{schubeler_schubeler_2025}}
  \label{fig:DS38edf}
}
\capbtabbox{%
  \begin{tabular}{cc}\hline
  \bf{Parameter(s)} & \bf{Value}\\ \hline \hline
  Max. Thrust & 27 N \\ \hline
  Max. Exhaust speed & 87.3 m/s \\ \hline
  Duct weight & 100 g \\ \hline
  Motor weight & 300 g \\ \hline
  Duct diameter & 70 mm \\ \hline
  Duct length & 60 mm \\ \hline
  \end{tabular}
}{%
  \caption[Schuebeler DS-30 AXI HDS EDF with HET 700-68-1400 motor specifications]{Schuebeler DS-30 AXI HDS EDF\cite{schubeler_schubeler_2025}}
  \label{table:DS38specs}
}
\end{floatrow}
\end{figure}

MAD Motors is a well-regarded manufacturer of high-performance motors that are frequently used by the lab for motor selection in various projects. Their motors are chosen for their reliability, precision, and versatility, making them ideal for applications involving thrust production and propeller designs. These motors are widely used as a baseline for research in the lab, particularly when studying the relationship between motor specifications and their capability to drive propellers or EDF's for thrust generation.

Research was conducted into the best RC motors to use for this application specifically focusing on commonly used actuators by the lab called MAD Motors’, to establish baseline data, helping to understand the potential of different motor and propeller combinations. This baseline data became essential in comparing various motor options and evaluating their performance in terms of thrust output and power requirements. The motors were tested with different propellers to observe the effects on performance, including factors such as thrust generation, efficiency, and operating conditions. In this context, the motors offered a consistent reference point for determining the best combinations for specific needs, such as creating propellers with the optimal thrust for a given application.

\begin{figure}[H]
    \centering
    \includegraphics[width=1\linewidth]{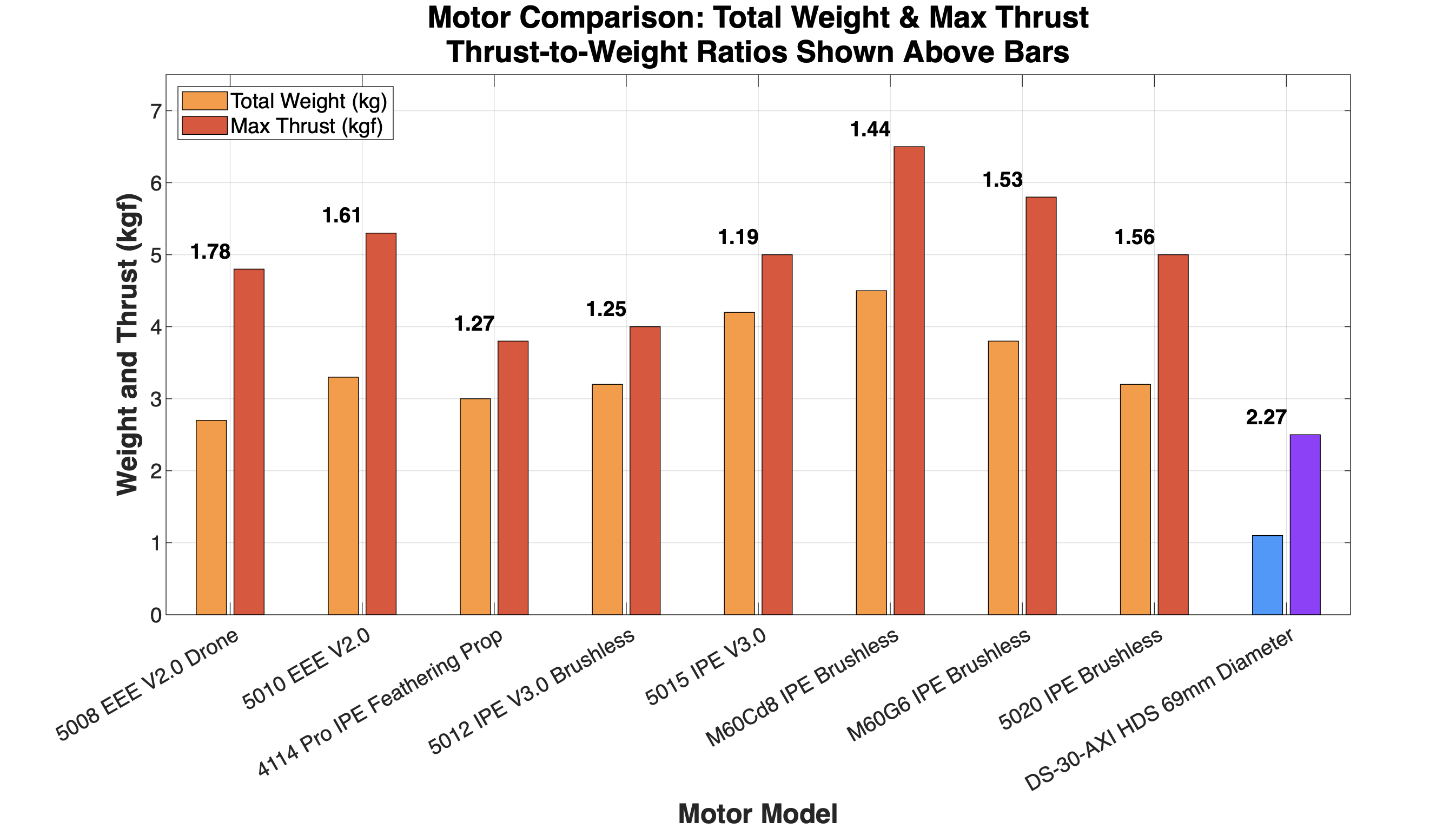}
    \caption{Motor Propeller Thrust Relationship}
    \label{fig:motor-weight-thrust}
\end{figure}

The Fig.~\ref{fig:motor-weight-thrust} below provides a summary of the data collected from this motor research, focusing on the performance of two different motors, each paired with a propeller. Notably, one of these motors was an EDF unit, chosen for its specific weight and the desired thrust output. The table showcases how these combinations were analyzed and compared, considering various parameters like thrust, RPM, and power consumption. This data helped refine the understanding of how to select motors and propellers that would meet the thrust requirements of different systems and how MAD Motors’ offerings fit into the larger picture of motor and propeller selection for lab experiments.

Through this research, MAD Motors’ products served as a valuable resource in exploring the capabilities of various motors and their applications in thrust generation, laying the groundwork for further developments in propeller design and EDF technology.

EDFs were chosen as the ideal thruster option, primarily due to their ability to deliver high thrust in compact volumes while also housing the blades. The Schuebeler HDS series, specifically the DS-30-AXI HDS model (Fig.~\ref{fig:DS38edf}), was selected for its remarkable energy density achieved through the combination of lightweight carbon fiber duct and blades, again backed up by figure 2.5 above as seen in the smallest weight to thrust ratio, called out with the blue EDF circle in the rightward legend.

For controlling the brushless DC motor connected to the EDF, an ESC is employed. Unlike the brushless motor drives, ESCs typically determine the rotor's position and speed by measuring back EMF during the inactive phase of the three-phase DC motor's rotation, rather than relying on position sensors like encoders or hall effect sensors. The chosen ESC for Harpy is the YGE 95A LV Telemetry ESC, known for its lightweight design. This ESC supports input voltages from 2 to 6S LiPo batteries and delivers a continuous current of 95A. It features a powerful adjustable BEC, which can be set from 5.5V to 8.4V, with some configurations supporting up to 9V. The BEC provides a continuous current of 8A and can peak up to 18A, making it capable of powering high-demand servos and onboard electronics. The “T” in 95LVT stands for telemetry, and the ESC supports a wide range of telemetry protocols.

Physically, the ESC measures 78 x 32 x 15 mm and weighs approximately 63 grams without cables and 93 grams with cables. The controller uses a sinusoidal startup for smooth motor engagement and automatically adjusts the PWM frequency to match the motor’s requirements. It includes built-in current limitation and is based on a 32-bit firmware architecture, which can be updated via an optional USB adapter.

Before integrating the EDF-ESC combo into Harpy, extensive testing was conducted to ensure reliable performance and compatibility. The combination was evaluated for power handling, responsiveness, and PWM signal response. This allowed for fine-tuning of the system to ensure smooth operation and prevent overheating or instability during flight. These tests also helped verify the proper integration of the ESCs with the EDFs, ensuring they worked seamlessly together for optimal thrust output. Further discussion into testing and EDF and ESC integration can be found in chapter \ref{chap:prop-hardware-integrate}.


%% file: tex/EDF_Thrust_Characterization.tex
\chapter{Thrust Characterization}
\label{chap: Thrust chacterization}

To understand the necessary characteristics of the Harpy EDF thruster, a thorough investigation into its thrust output was conducted. This investigation became necessary due to the limited available literature that specifically addresses the interplay between crucial parameters—such as pitch, diameter, RPM (revolutions per minute), and environmental factors—on the thrust performance of an RC (radio-controlled) type motor and blade system. The absence of direct empirical data on how these variables influence the thruster’s output in various conditions required a more systematic approach to derive the needed performance characteristics.

The first step in this process involved considering the fundamental principles of fluid dynamics and energy conservation to build a theoretical framework. Using the principle of conservation of energy, it was possible to estimate the amount of energy being converted into thrust by the EDF system. The conservation of mass flow principles was then applied to model the airflow characteristics through the fan blades most impacted by their diameter in relation to the cross sectional area of a circle. The relationship between the mass flow rate, pressure change, and velocity of the airflow was central to understanding how the EDF would perform under different operating conditions.

To refine this theoretical model, empirical data from existing EDF systems and RC motors were incorporated. This data helped to identify key performance trends related to variables such as the pitch of the blades, the diameter of the ducted fan, and the rotational speed of the motor. Through this empirical analysis, it was possible to estimate the expected thrust output under various operational conditions, including differences in atmospheric pressure, temperature, and altitude. These environmental variables are crucial as they directly affect air density and, therefore, the efficiency of the thrust production.

A specific equation was derived based on these principles that links the aforementioned parameters—pitch, diameter, RPM, and environmental conditions—to the thrust output of the system. This equation not only provides an analytical tool for predicting the performance of the Harpy EDF thruster in a variety of conditions but also serves as a foundation for future design optimizations. By understanding how each of these factors contributes to the overall thrust, future iterations of the Harpy robot can be fine-tuned to achieve optimal performance in different operational environments, whether it’s for short low-altitude flights or quick hopping or trotting menuvers to combat difficult terrain.

Through this process, a deeper understanding of the capabilities of the EDF thruster was achieved, offering insight into how variations in design and operating conditions can influence overall system performance. This lays the groundwork for making informed decisions on component selection and system configurations that will maximize the Harpy robot’s stability and functionality with relation to roll control, and thrustoutput for short term low altitude flights, trotting,and jumping.

\section{Theoretical Thrust Equation Derivation}
\label{sec:thrust-theo}

The derivation of the thrust equation, which incorporates blade diameter, pitch, rotational speed (RPM), and ambient air velocity, is fundamentally grounded in Newton’s Second Law of Motion. By analyzing the change in momentum imparted to the air by the rotating blades, a direct relationship is established between these physical parameters and the resulting thrust force.

Starting from Newton's Second Law:

\begin{equation}
F = m a = \dot{m} v,
\end{equation}

where:
\begin{itemize}
    \item $F$ is the thrust force [N],
    \item $m$ is the mass [kg],
    \item $a$ is the acceleration [m/s$^2$],
    \item $\dot{m}$ is the mass flow rate [kg/s],
    \item $v$ is the velocity of the air [m/s].
\end{itemize}

The thrust is caused by the change in momentum of the air:

\begin{equation}
F = \dot{m} \Delta v_e = \dot{m}(v_e - v_{ac}),
\end{equation}

where:
\begin{itemize}
    \item $v_e$ is the exit velocity of air accelerated by the propeller [m/s],
    \item $v_{ac}$ is the velocity of ambient (incoming) air [m/s].
\end{itemize}

The mass flow rate $\dot{m}$ can be expressed as:

\begin{equation}
\dot{m} = \rho A v_e,
\end{equation}

where:
\begin{itemize}
    \item $\rho$ is the air density [kg/m$^3$],
    \item $A$ is the cross-sectional area swept by the propeller blades [m$^2$], which depends on blade diameter $d$ as $A = \frac{\pi d^2}{4}$.
\end{itemize}

Substituting $\dot{m}$ into the thrust equation gives:

\begin{equation}
F = \rho A v_e (v_e - v_{ac}),
\label{eq:sub_equation}
\end{equation}

or more compactly:

\begin{equation}
F = \rho A v_e v_r,
\end{equation}

where $v_r = v_e - v_{ac}$ is the relative velocity between exit flow and ambient air. By substituting the cross-sectional area, which is proportional to the square of the propeller blade diameter, and incorporating the mass flow rate into the larger thrust equation, we can establish a more comprehensive relationship that accounts for the key factors influencing thrust generation.

\begin{equation}
F = \rho \frac{\pi d^2}{{4}}(v_e^2-v_ev_{ac})
\label{eq:3.6}
\end{equation} 
To calculate the pitch, the following relationship between RPM and its conversion into inches was derived and integrated into Eq.~\ref{eq:sub_equation}, allowing for a more accurate representation of the propeller's geometry and its impact on thrust production.

\begin{equation}
v_{\text{pitch}} = \text{RPM} \times \text{Pitch}_{\text{in}} \times \frac{1~\text{ft}}{12~\text{in}} \times \frac{1~\text{mile}}{5280~\text{ft}} \times \frac{60~\text{min}}{1~\text{hr}},
\end{equation}

In order to get the output in KGF, the propeller diameter and pitch outputs must be converted from inches to meters, and this relationship is 1 inch = 0.0254 meters. Substituting this in to Eqs.~\ref{eq:sub_equation} and \ref{eq:3.6} while also simplifying, the equation becomes:

\begin{equation}
F_{\text{thrust}} = 1.225 \cdot \frac{\pi \cdot (0.0254 \cdot d)^2}{4} \cdot \left[\frac{\text{RPM}_{\text{prop}} \cdot 0.0254 \cdot \text{Pitch}_{\text{prop}}}{60}\right]^2 \cdot \left(K_1 \cdot \frac{d}{\text{Pitch}_{\text{prop}}}\right)^{K_2},
\label{eq:K1K2Equation}
\end{equation}

The diameter to pitch ratio for a propeller blade generally falls between 0.5 and 2.5, with the most ideal radio being between 0.8 and 1.8. In propeller design, the blade area ratio (BAR), defined as the ratio of the blade's surface area to the total area swept by the propeller, significantly influences performance characteristics. Maintaining an optimal BAR is crucial for balancing thrust generation and efficiency. According to AB Marine, propellers should have a minimum surface area to prevent excessive loads on individual blades, which could lead to cavitation\cite{ab_marine_ab_2025}. To incorporate pitch and this ratio into the thrust equation, two coefficients were added,represented by \( K_1 \)and \( K_2 \) seen in equation \ref{eq:K1K2Equation}. This was based off of research conducted by the RC electric guy, but slightly modified the coefficients to increase fidelity and accuracy of the model. 

The values of \( K_1 \)and \( K_2 \)were determined through a process of trial and error, utilizing data from vendors that provide motor and propeller blades with known thrust outputs. \( K_2 \), the ``power constant,'' represents how power input is related to thrust output. The value of \( K_2 \) was chosen to be 1.5 after experimenting with various options. This particular value of 1.5, or 3/2, seemed to fit the data well, although its precise analytical basis is still unclear. It might be related to the power law dynamics inherent in the EDF system, where thrust is likely to follow a non-linear relationship with power.

Additionally, the relationship between thrust and propeller diameter often resembles a logarithmic curve, where increases in diameter lead to large gains in thrust at smaller sizes but gradually level off as diameter continues to grow. Through associated research and empirical testing, it was found that a value of approximately 1.5 provided a good fit for modeling this behavior, accurately capturing the way thrust scales with propeller size. This reflects how the thrust increase is not purely linear but instead follows a curve that matches the characteristic shape of a logarithmic relationship.

Once \( K_2 \) was set to 1.5, the next step was to solve for \( K_1 \), the "coefficient constant," which is associated with the fan's diameter-to-pitch ratio. The goal was to minimize the error between the predicted thrust values and the actual measured thrust across a set of 149 data points. To find \( K_1 \), a goal-seeking method was used. This involved incrementally adjusting \( K_1 \) and observing how the predicted thrust values compared to the actual data. After a series of trials, \( K_1 \) was determined to be approximately 0.30, which resulted in the smallest average error between the predicted and actual thrust values. This process was conducted by other researchers, and repeated with additional datsets found online via RC motor manufacturer sites. The plot that was created for this dataset to hone in on the \( K_1 \) value can be seen below. All of these predicted blade sizes all came from MAD motors, and Amazon, where expected thrust outputs were defined for motors at specific RPM's, blade diameters, and pitch.

\begin{figure}[H]
    \centering
    \includegraphics[width=0.75\linewidth]{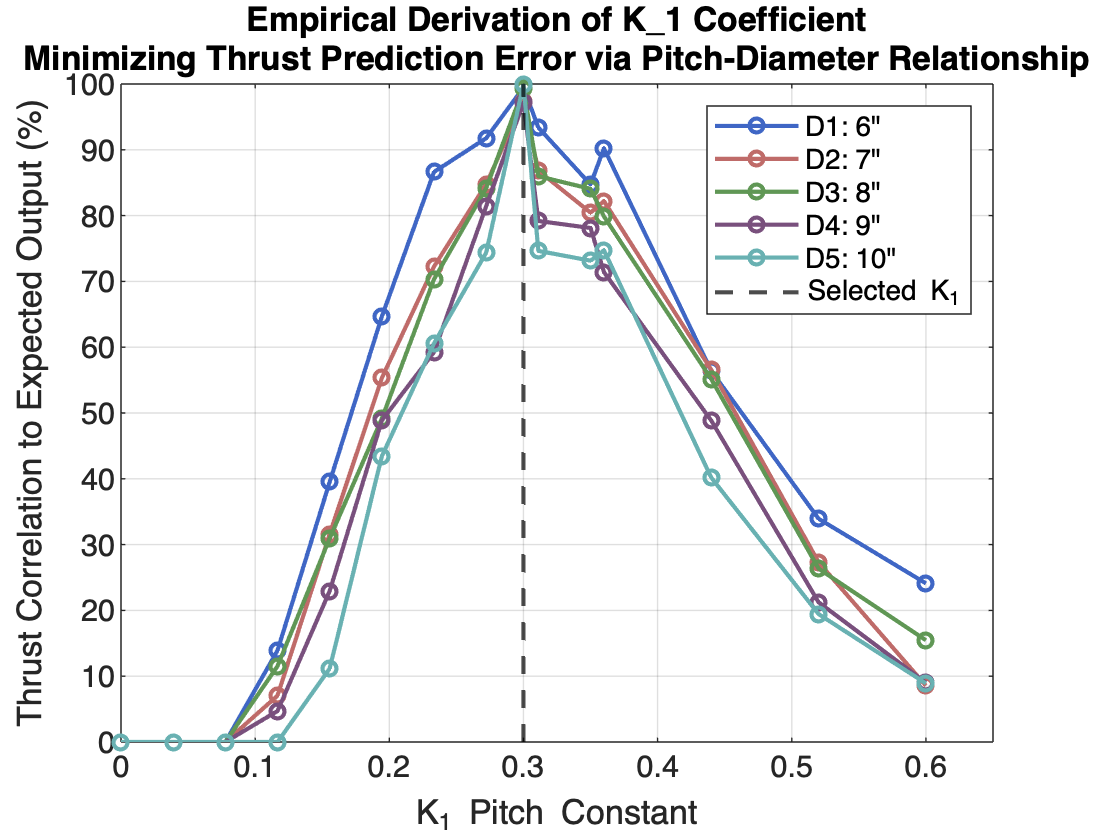}
    \caption{Illustration of data while solving for pitch-diameter coefficient \( K_1 \)}
    \label{fig:k1-pitch-diam-coeff}
\end{figure}

Through this iterative process, \( K_2 \) was first set based on experimentation, and then \( K_1 \) was fine-tuned to achieve an optimal fit. Although the value of 1.5 for \( K_2 \) may seem somewhat mysterious, it proved effective, and the choice of 0.30
for \( K_1 \) allowed for the best agreement between the predicted and actual thrust values, minimizing the error.

The plot compares the predicted thrust values (using \( K_1 = 0.30 \) and \( K_2 = 1.5 \)) against actual thrust values. The red dashed line represents the ideal case where predicted thrust equals actual thrust. The data points cluster around this line, indicating that the chosen constants provide a reasonable fit.

\subsection{Results}

The theoretical thrust equation will be used to verify the output of the EDFs by comparing the calculated thrust with the measured performance during testing. This equation takes into account factors such as motor power, propeller efficiency, and air density to predict the thrust generated by each EDF. Additionally, the desired location for mounting the EDFs will be carefully chosen to ensure the correct moment arm, allowing for proper distribution of thrust and achieving balanced control of Harpy. By integrating these theoretical calculations with empirical testing, we can fine-tune the robot's flight dynamics and ensure accurate thrust production for stable, efficient movement.

After solving for the \( K_1 \)and \( K_2 \) coefficients, the final theoretical thrust equation that incorporates blade diameter, pitch, RPM, as well as a static or dynamic external environment is as followed.

\begin{equation}
\begin{split}
F_{\text{thrust}} &= 1.225 \cdot \frac{\pi \cdot (0.0254 \cdot d)^2}{4} \cdot \left( \frac{\text{RPM} \cdot 0.0254 \cdot \text{Pitch}}{60} \right)^2 \\
&\quad - \left( \frac{\text{RPM} \cdot 0.0254 \cdot \text{Pitch}}{60} \cdot v_o \right) \cdot \left( \frac{d}{3.29546 \cdot \text{Pitch}} \right)^{1.5}
\end{split}
\label{eq:FinalEquation}
\end{equation}

\section{Numerical Modeling Using ANSYS}
\label{sec:ansys]}

An ANSYS Fluent CFD simulation was conducted to verify the theoretical hand calculation summarized in Section \ref{sec:thrust-theo} involving a rotating propeller within a cylindrical domain of the same diameter. The simulation setup included two distinct cell zones: Body 1, representing the propeller, was assigned a constant rotational velocity of 20,000 RPM, while Body 2, the surrounding air, was modeled as a stationary medium. Boundary conditions were carefully defined, with input and output velocity directions strictly controlled to reflect realistic operating conditions. Mesh motion was applied along the input velocity axis in coordination with the propeller's rotation. A computational mesh consisting of 200 nodes provided a sufficient number of data points for analysis. The turbulence model utilized was k-epsilon with realizable and scalable near-wall treatment, ensuring reliable predictions of turbulent behavior near the propeller surface. Additionally, a viscous-laminar flow model was tested to assess the influence of different flow regimes. This simulation setup enabled a direct comparison between the numerical results and the theoretical calculations, validating the accuracy of the analytical approach. A head on image of the dual-blade simulation at an arbitrary velocity can be seen below, where each blue sphere is a velocity node.

\begin{table}[htbp]
\centering
\footnotesize
\caption{Summary of ANSYS Fluent CFD Simulation Parameters}
\begin{tabularx}{\textwidth}{>{\raggedright\arraybackslash}p{4.5cm} X}
\toprule
\textbf{Simulation Parameter} & \textbf{Description} \\
\midrule
Simulation Software & ANSYS Fluent (CFD) \\
Objective & Validation of theoretical hand-calculated thrust model \\
Domain Geometry & Cylindrical domain with diameter equal to that of the propeller \\
Cell Zones & Body 1 (rotating propeller at 20,000 RPM), Body 2 (stationary surrounding air) \\
Boundary Conditions & Inlet and outlet velocity directions explicitly defined \\
Mesh Motion & Aligned with input velocity axis, coordinated with propeller rotation \\
Mesh Resolution & 200 velocity nodes used for flow sampling \\
Turbulence Model & $k$–$\varepsilon$ (realizable) with scalable near-wall treatment \\
Flow Regimes Tested & Turbulent and laminar (viscous) flow regimes \\
Validation Goal & Direct comparison of CFD thrust results against theoretical predictions \\
\bottomrule
\end{tabularx}
\end{table}

\begin{figure}[H]
    \centering
    \includegraphics[width=1\linewidth]{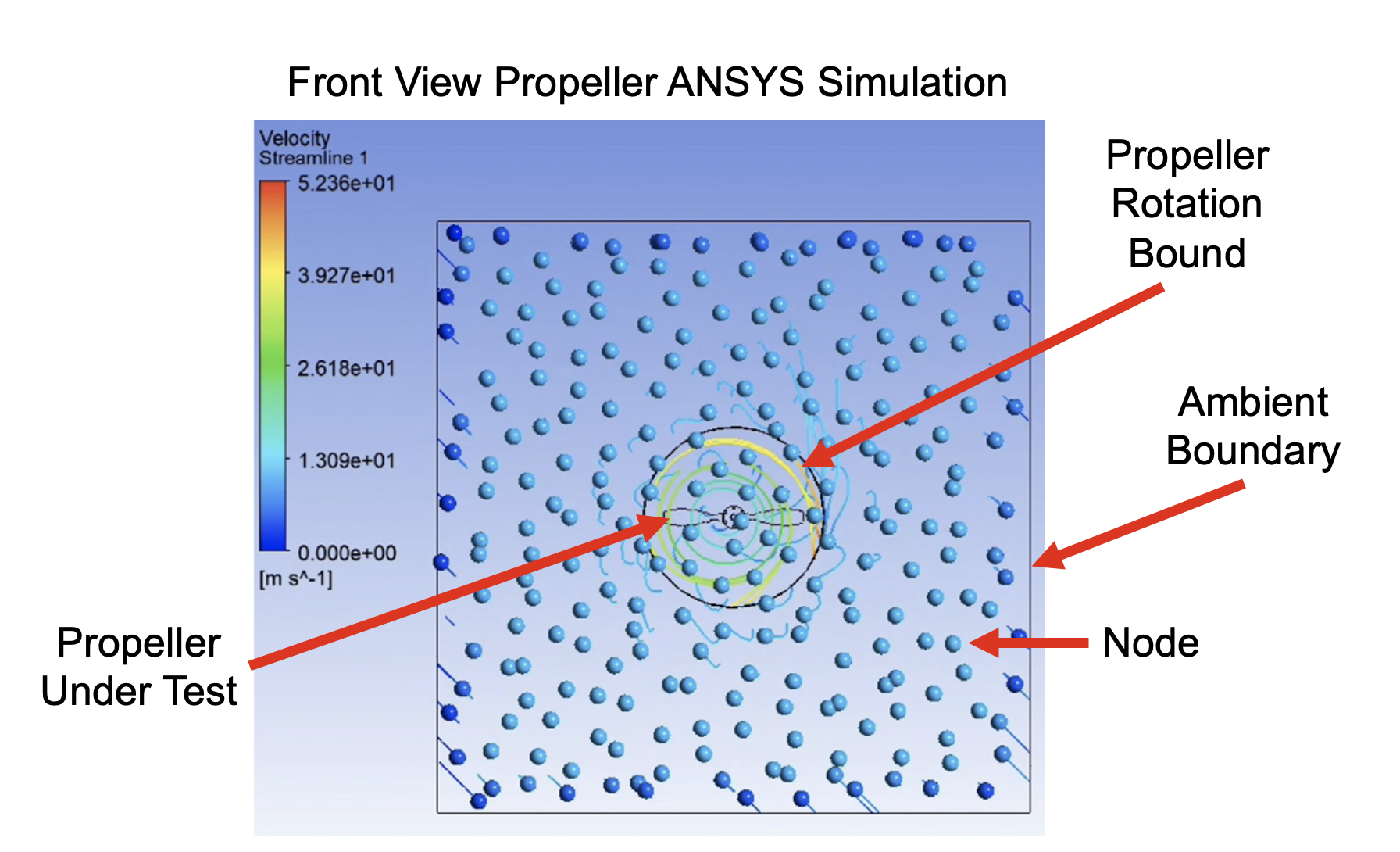}
    \caption{Head on ANSYS CFD Dual-Prop Simulation}
    \label{fig:ansys}
\end{figure}

In the simulation shown below, the flow dynamics can be observed with the inlet velocity directed towards the right side of the image and the outlet positioned on the left. The simulation includes 200 nodes representing the air velocity, with their respective trajectories visible throughout the image. These nodes help track the path of the airflow as it moves through the system. Notably, both in the image above and the one below, increased disturbances in the air velocity can be seen when the airflow interacts with the propellers. This disturbance is particularly evident in the yellow and green nodes, whose trajectories show significant deviations due to the effects of the rotating blades. These disturbances indicate the interaction between the airflow and the props, which contribute to changes in the velocity profile and overall flow characteristics in the vicinity of the rotor.

\begin{figure}[H]
    \centering
    \includegraphics[width=0.8\linewidth]{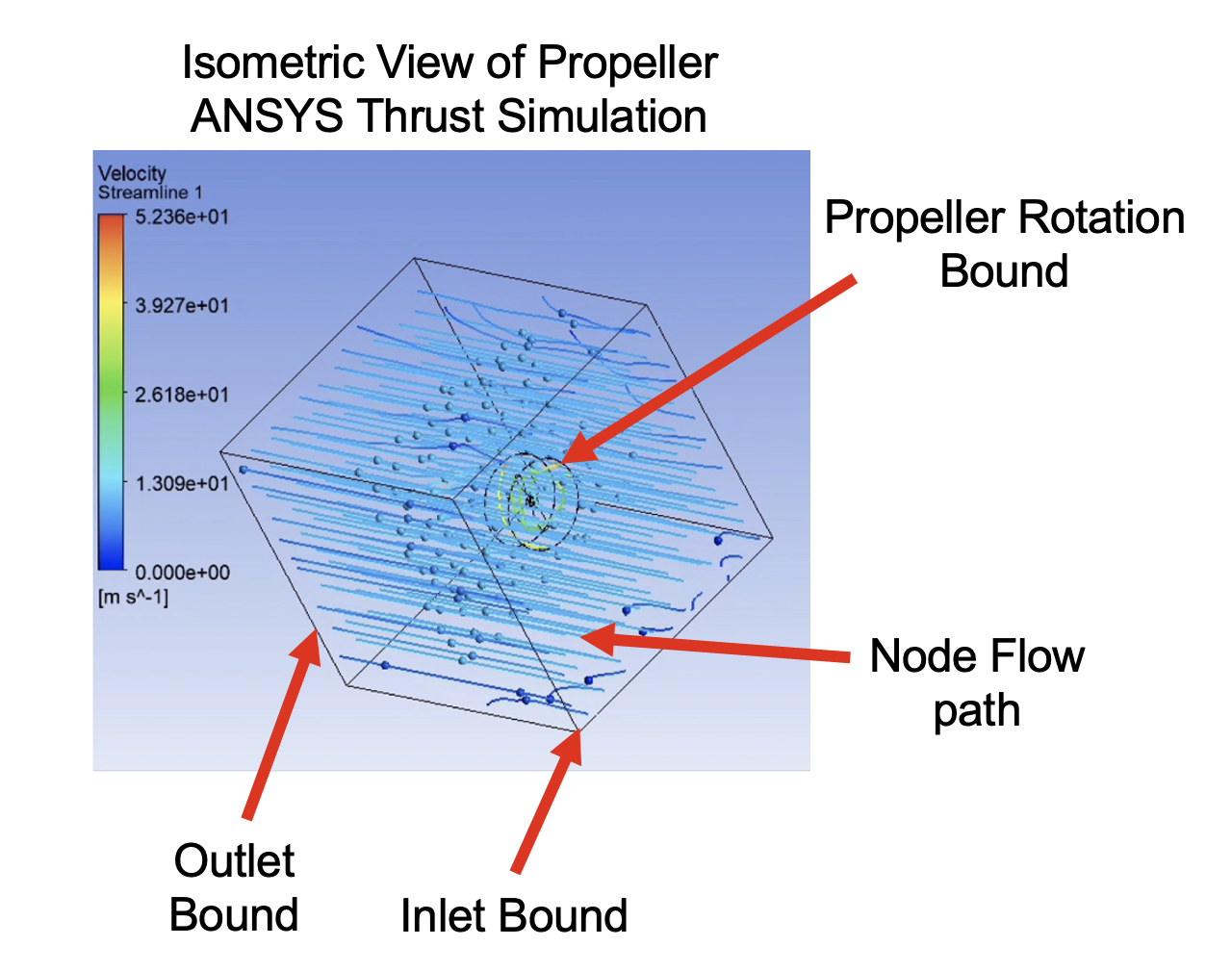}
    \caption{Isometric View of ANSYS Propeller Thrust Simulation}
    \label{fig:ansys-sim}
\end{figure}

\begin{figure}[H]
    \centering
    \includegraphics[width=0.75\linewidth]{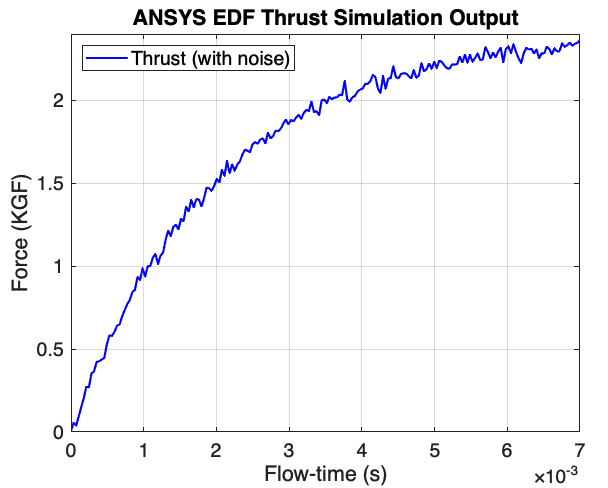}
    \caption{Harpy ANSYS Thrust Output at Observed Node}
    \label{fig:Ansys_outputThrust}
\end{figure}

A single node, located directly at the output of the propeller's volumetric area, was used as the analysis point for determining the total thrust produced by the propeller. This analysis was conducted multiple times to ensure consistency and accuracy in the results. The thrust values captured by this thrust node were recorded for each test, and the data was subsequently plotted to visualize the relationship between flow time and force. This is similar to the "ANSYS EDF Thrust Simulation Output" plot shown in Fig.~\ref{fig:Ansys_outputThrust} produced for the Harpy EDF motor that was integrated into the upper assembly, where the thrust output is correlated with varying conditions and time intervals, providing valuable insights into the EDF's performance over the course of the tests, with a max thrust output of about 2.4 KGF exported from the simulation.

In the lab, the EDF was tested by varying the PWM signal to control motor speed and measure the resulting thrust. The EDF unit was securely mounted on a thrust stand, ensuring that accurate force measurements could be taken without interference. A high-precision load cell or thrust sensor recorded the thrust output, while a microcontroller, such as an Arduino or an ESC tester, generated the PWM signals. These signals controlled the electronic speed controller (ESC), which in turn regulated the EDF motor speed. A stable power supply provided consistent voltage and current, and in some cases, a tachometer was used to monitor the fan's rotational speed.

To evaluate the EDF's performance, the PWM duty cycle was incrementally increased, typically from 1000 ms (representing the minimum throttle) to 2000 ms (maximum throttle). At each step, the EDF was allowed to stabilize before thrust measurements were recorded. This process ensured that data collection was accurate and minimized the effects of transient responses. Voltage and current were also monitored to analyze power consumption and efficiency.

Once the data was collected, it was analyzed to understand the relationship between the PWM signal and the resulting thrust. The data was plotted, typically with the PWM signal in microseconds on the x-axis and thrust output in grams or Newtons on the y-axis. Additional plots could be generated to compare power consumption versus thrust output, providing insights into efficiency. Repeating the experiment multiple times ensured consistency and reliability in the results.

\begin{figure}
    \centering
    \includegraphics[width=0.75\linewidth]{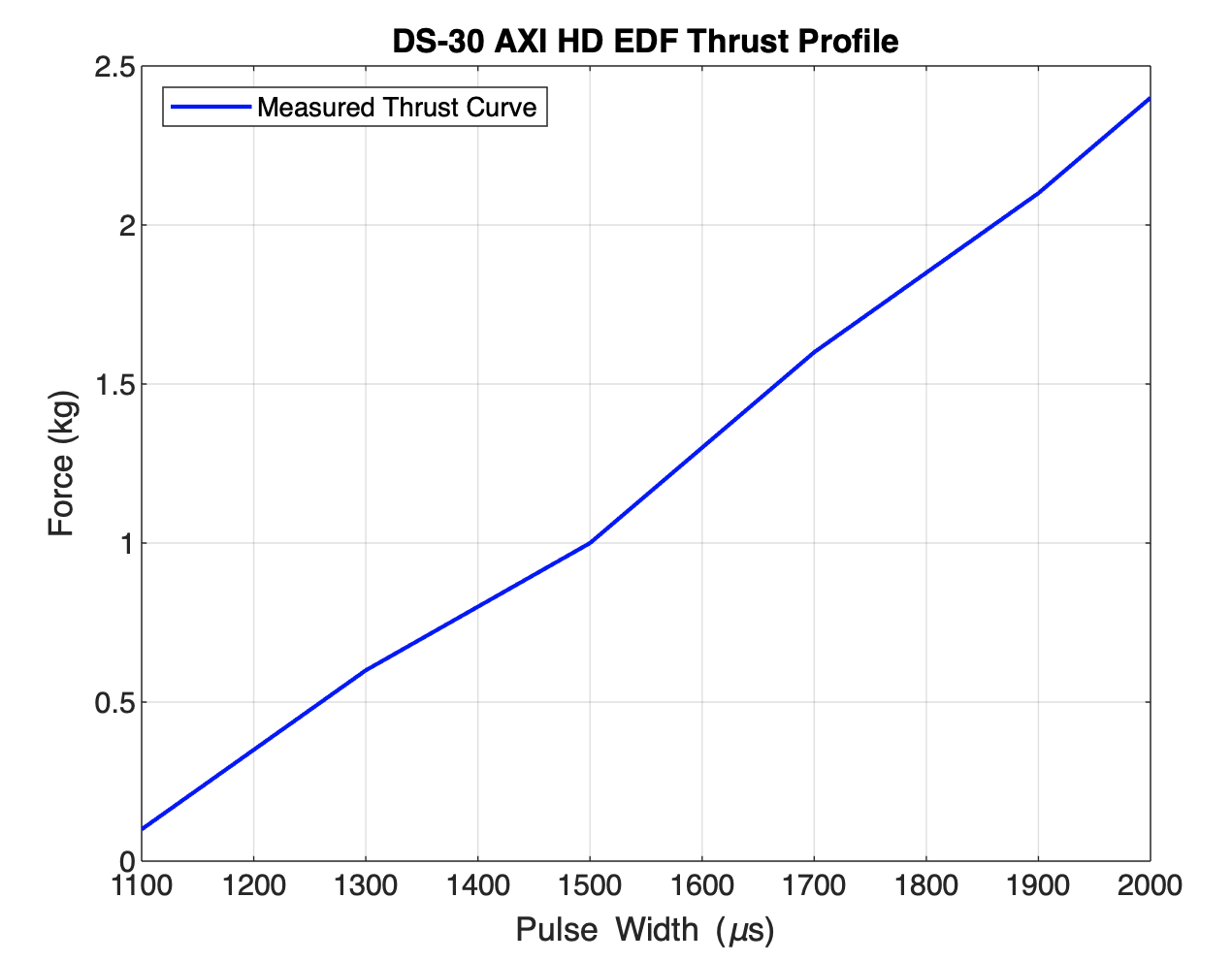}
    \caption{Loadcell measurements}
    \label{fig:loadcell}
\end{figure}

This empirical output is consistent with the predicted ANSYS analysis from Fig.~\ref{fig:loadcell} where the total thrust maxed out around 2.4 KGF. Although this plot has a x-axis labled as pulse width, it shares the same y-axis scale, as time to max thrust is not a critical comparative variable in this investigation. The EDF tested to produce figure \ref{fig:loadcell} was tested on a tabletop platform at UC Berkely and the pulse width was toggled between 1-2 ms to modulate the thrust output. This empirical output and the ANSYS CFD output and the theoretical thrust output can all be compared to one another for the Harpy thruster to validate its predicted performance and also provide confidence for methods of validation for future prop and thruster testing. Section~\ref{sec:theo-vs-ansys-vs-loadcell} will delve more into the Harpy thruster test methods and compare them to one another. 

\section{Validation of Theoretical Thrust Model via ANSYS CFD}
\label{sec:theo-vs-ansys-vs-loadcell}

To validate the hand-derived theoretical thrust model, a comparative analysis was conducted using ANSYS Fluent CFD simulations. The same set of input parameters—blade diameter, pitch, a constant rotational speed of 20,000 RPM, and a static ambient environment—was used across both the theoretical and numerical models to ensure a one-to-one comparison. These parameter values were selected based on standard dimensions commonly observed in EDF-equipped robots, including those in the Silicon Synapse Lab and potential configurations for the Harpy platform.

\begin{figure}[H]
    \centering
    \includegraphics[width=0.75\linewidth]{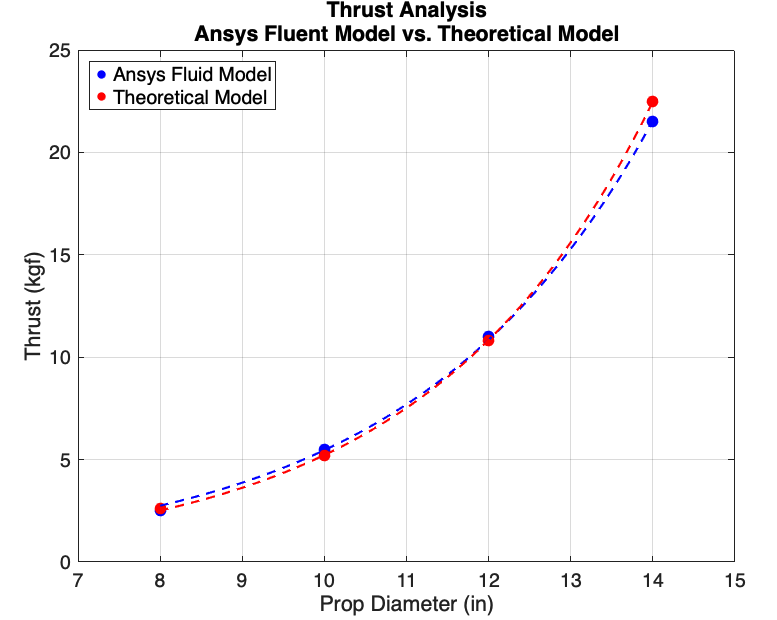}
    \caption{Comparison of ANSYS Fluent and hand-derived theoretical thrust calculations.}
    \label{fig:ansys-vs-theory}
\end{figure}

Four propeller configurations were analyzed, each combining a distinct blade diameter and corresponding pitch. The pitch values were selected based on typical vendor-reported specifications for each blade size. The plot in Fig.~\ref{fig:ansys-vs-theory} shows the comparison of the thrust values predicted by the theoretical model and those obtained from ANSYS Fluent simulations.

The average error between the ANSYS CFD results and the theoretical model was found to be 5.22\%, which is acceptably small. This level of agreement confirms that the theoretical model captures the core dynamics of thrust production and serves as an efficient tool for quick estimations without the computational cost of full CFD simulations.

\begin{table}[H]
\centering
\begin{tabular}{|c|c|c|c|c|}
\hline
\textbf{Diameter [in]} & 
\textbf{Pitch [in]} & 
\textbf{ANSYS Thrust [kgf]} & 
\textbf{Theoretical Thrust [kgf]} & 
\textbf{Error [\%]} \\
\hline
8  & 5 & 2.60  & 2.455  & 5.57 \\
10 & 5 & 5.80  & 5.36   & 7.58 \\
12 & 6 & 10.70 & 11.11  & 3.83 \\
14 & 8 & 21.20 & 22.017 & 3.85 \\
\hline
\end{tabular}
\caption{Comparison between ANSYS Fluent and hand-derived theoretical thrust values.}
\end{table}

The strong agreement between both methods validates the theoretical model as a practical and reliable alternative for estimating thrust in early-stage design and analysis workflows.

\vspace{1em}
\noindent The Schübeler DS-30-AXI-HDS EDF motor was selected for Harpy due to its optimal balance between thrust capacity and motor mass. With a maximum thrust output of 27~N (approximately 2.7~kgf), this EDF is well-suited for compensating Harpy's body weight during roll maneuvers and contributing to hybrid ground-aerial control. This selection is further justified through combined theoretical analysis and ANSYS-based simulation, reinforcing the model's predictive validity.

The DS-30-AXI-HDS operates at up to 55,000 RPM and features an outer diameter of 69~mm. However, due to a central hub occupying approximately 25.4~mm (1~inch) of diameter, only the annular region contributes to effective thrust generation. Accordingly, thrust calculations account for this reduced cross-sectional area.

Using the methodology presented in Eq.~\ref{eq:FinalEquation}, which integrates this correction, the expected thrust is calculated as approximately 2.64~kgf, closely matching the manufacturer’s specification. This outcome provides confidence in using the thrust equation for performance prediction.

\begin{table}[H]
\centering
\begin{tabular}{|c|c|}
\hline
\textbf{Analysis Method} & \textbf{Result [kgf]} \\
\hline
ANSYS CFD Simulation    & 2.4 \\
Theoretical Thrust Model & 2.64 \\
Empirical Bench Testing  & 2.4 \\
Manufacturer Specification & 2.75 \\
\hline
\end{tabular}
\caption{EDF thrust outputs across different analysis methods for the DS-30-AXI-HDS motor.}
\end{table}

These findings validate the use of the hand-derived thrust equation for rapid design evaluations of EDF-based propulsion in lightweight legged robotic systems. The consistency between theoretical, computational, and experimental results underscores the reliability of this modeling approach in control-critical applications, such as roll stabilization and low-altitude locomotion in hybrid platforms like Harpy.

%% file: tex/Thruster_design_and_incorporation.tex
\chapter{Propulsion Hardware Overview and Integration}
\label{chap:prop-hardware-integrate}

\section{Background}
\label{chap:prop-background}

As discussed in the introduction, the challenge of achieving increased roll control and stability in bipedal robots led to the incorporation of two EDF  thrusters, a solution explored in Chapter~\ref{sec:edf-mounts} of this thesis. After extensive research into various propeller-based thrusters, the Schübeler DS-30-AXI-HDS EDF motor emerged as the optimal choice for this application. This motor offers a high thrust-to-weight ratio, efficient power consumption, and a durable design, making it ideal for the precise and reliable control required in bipedal robotics. The DS-30-AXI-HDS motor’s compact yet powerful design is paired with its ability to provide substantial thrust while minimizing weight and power draw—key factors in maintaining the robot’s overall efficiency and balance.

To control the two EDF motors, two YGE 35 speed controllers were selected for their reliable and high-performance handling of brushless motors. These controllers allow for precise adjustments to the motor speed, ensuring accurate roll control. The motors and speed controllers are powered through a Y-harness, which efficiently distributes power from a single 6S LiPo battery, providing consistent voltage to each motor. Custom-designed mounts were fabricated to securely integrate the motors into the existing robot framework, ensuring both stability and ease of maintenance. Wires were carefully routed to avoid interference with any moving parts, preventing potential damage or signal disruption. Furthermore, considerations were made for heat dissipation, with strategic placement of the YGE speed controllers to ensure proper ventilation and to prevent overheating during operation. This integration of EDF motors, speed controllers, and power management systems significantly enhanced the robot’s ability to execute precise maneuvers and maintain stability in dynamic environments.

\section{Tethered Propulsion Tests}
\label{sec:edf-theted-tests}

\begin{figure}[H]
    \centering
    \includegraphics[width=0.75\linewidth]{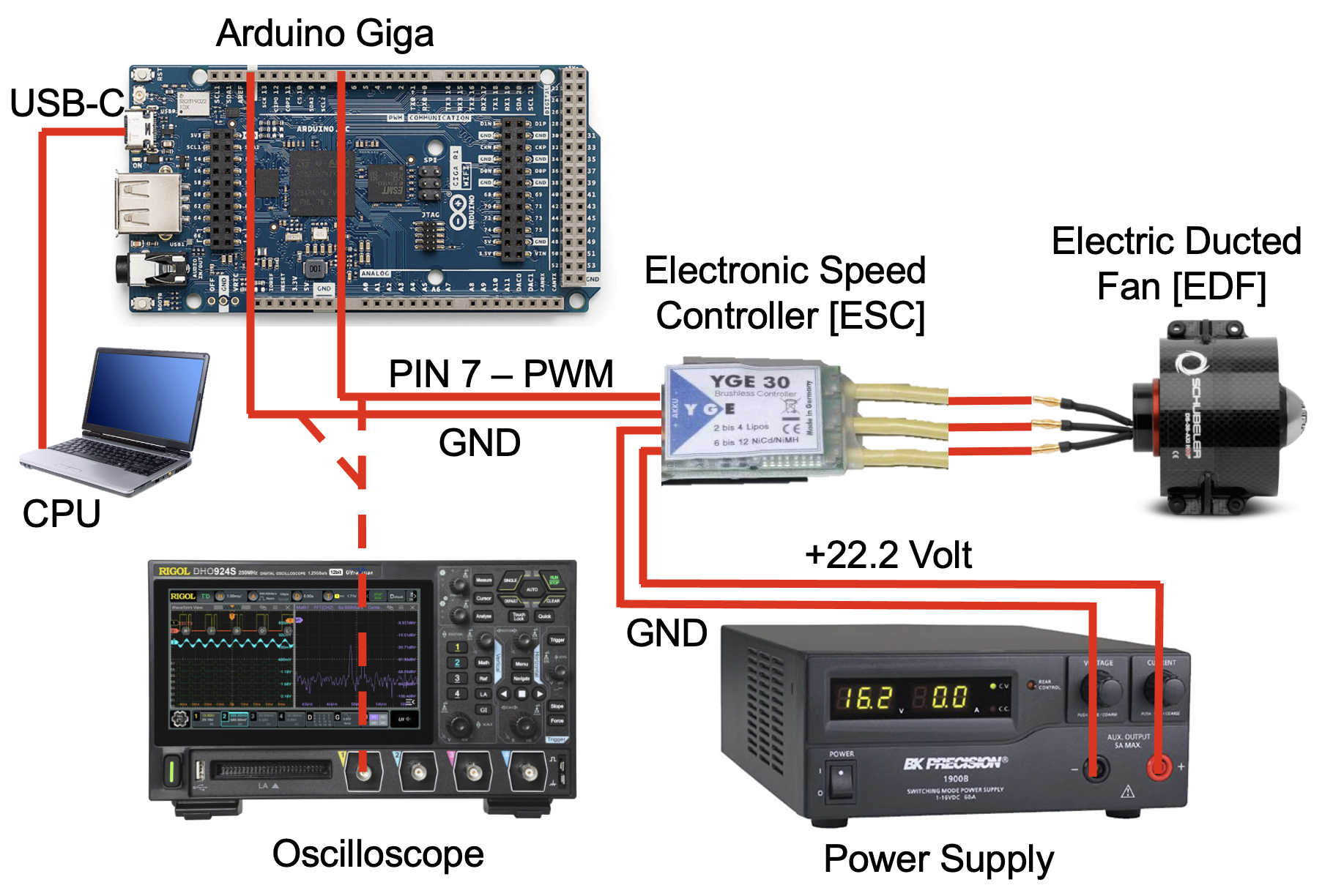}
    \caption{EDF Test Platform Schematic}
    \label{fig:edf-tethered-testbed}
\end{figure}

A dedicated testing platform (shown in Fig.~\ref{fig:edf-tethered-testbed}) was developed to evaluate a single YGE ESC in conjunction with an EDF motor before integrating it onto the Harpy platform. This precaution was taken to prevent potential damage to the existing hardware by ensuring proper ESC functionality and power delivery. The testing setup included an Arduino running a script that generated a steady 50Hz PWM signal with a pulse width modulated between 1ms and 2ms. These pulse widths corresponded to 0 \% and 100 \% thrust, respectively, effectively allowing full control over the EDF motor’s output. The ESC was supplied with a constant input voltage, replicating the conditions it would experience on the robot when powered by a battery capable of delivering sufficient current.

Through experimentation and thorough review of the ESC documentation, the 50Hz signal was identified as the optimal frequency for proper throttle response. The Arduino script facilitated precise adjustments to the EDF motor’s thrust by varying the pulse width, which was simultaneously observed using an oscilloscope. This oscilloscope verification ensured that the PWM signal was accurately transmitted to the ESC, aligning with expected output voltages. Additionally, each ESC required proper mode configuration to correctly interpret the incoming PWM signal and generate the necessary three-phase voltage to drive the EDF motor.

Once the testing process validated that the ESC responded correctly to PWM commands and delivered appropriate power to the EDF, the entire system—including the ESC, EDF, wiring, and power source—was securely mounted onto the robot. A battery with a Y-harness was used to distribute the same voltage to each ESC, ensuring uniform power delivery across the system. This careful validation process minimized potential integration risks and confirmed the compatibility of the ESC and EDF with the final robotic platform. 

The schematic shown in Fig.~\ref{fig:edf-tethered-testbed} illustrates a single EDF test platform setup designed to run the ESC and EDF safely through an Arduino GIGA, isolated from the main robot. This allows for a controlled environment to better understand and test the PWM signal control discussed earlier. In the full robot system, however, the ESC will receive its control signals from the Nucleo board, which serves as the primary micro controller responsible for orchestrating the robot's movement and control logic. Unlike the test bench setup, which relies on an external power supply, the integrated robot system will use a single onboard battery. This battery not only delivers consistent voltage across the system but can also supply more current than the test power supply, ensuring sufficient power for high-demand operations. 

\begin{figure}[H]
    \centering
    \includegraphics[width=0.75\linewidth]{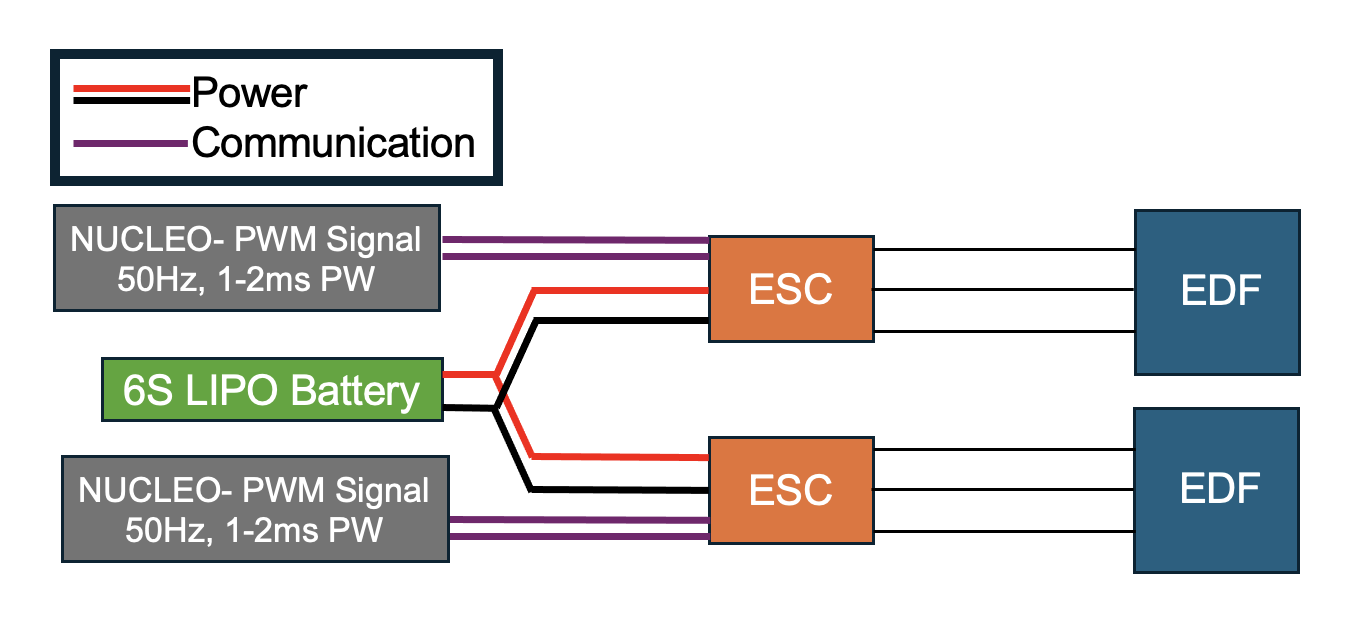}
    \caption{ESC-EDF Harpy Integration Schematic}
    \label{fig:ESC-EDF-schematic}
\end{figure}

The schematic shown in Fig.~\ref{fig:ESC-EDF-schematic} provides a detailed view of the EDF and ESC wiring and can be directly contrasted with the test configuration shown above for how it is integrated into Harpy on the robot.

\section{Thruster Mounts}
\label{sec:edf-mounts}

\begin{figure}[H]
    \centering
    \includegraphics[width=0.9\linewidth]{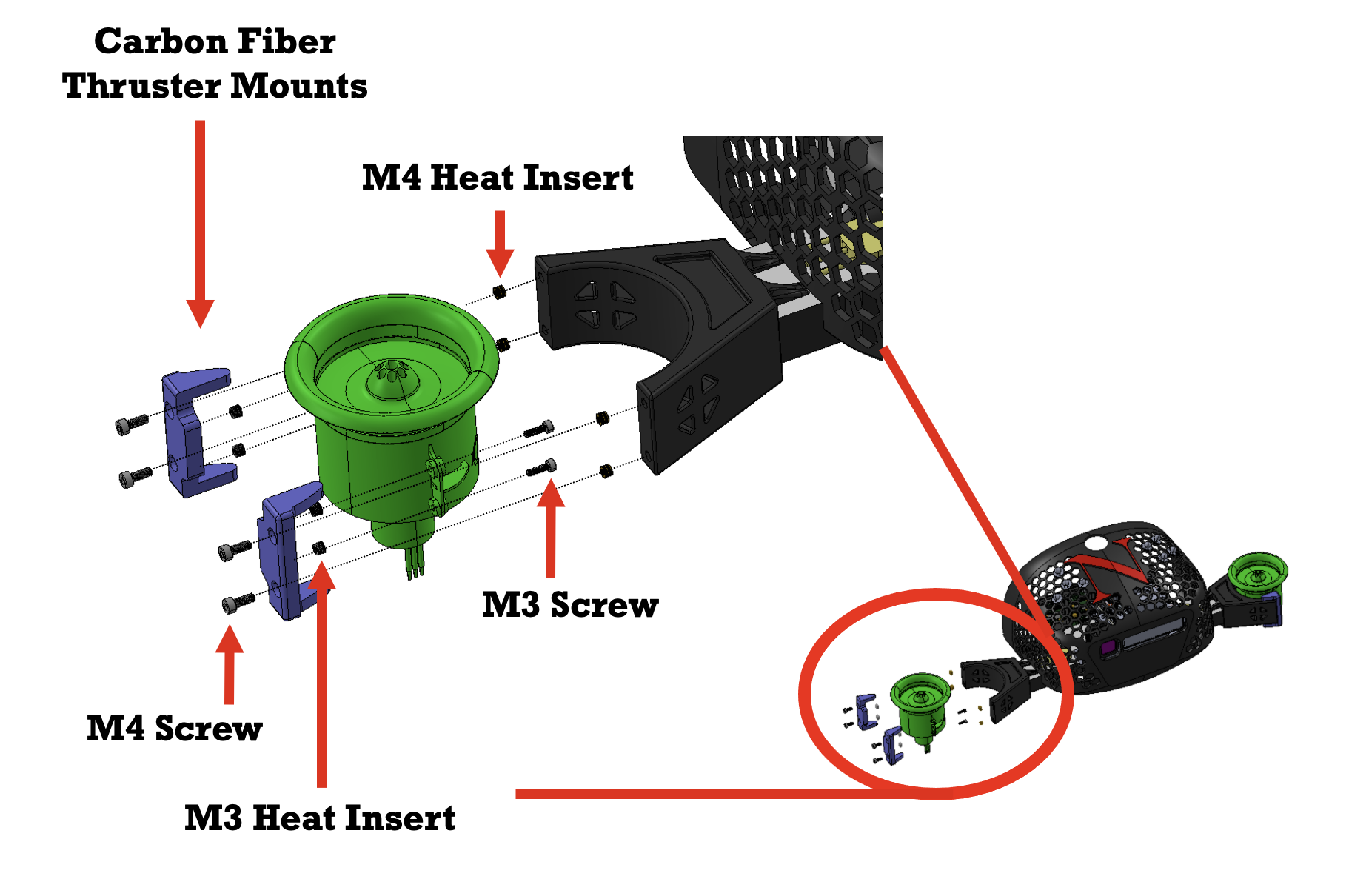}
    \caption{Illustration of EDF fixation onto the torso structure.}
    \label{fig:edf-mounds}
\end{figure}

The EDFs are rigidly mounted, as shown in Fig.~\ref{fig:edf-mounds}, to the robot using four M4 screws and custom-designed carbon fiber-reinforced mounts made from Onyx, a high-strength composite material suitable for structural applications. Threaded heat-set inserts were embedded into the mounts to allow seamless integration with the robot's frame and to enable easy removal and re installation when needed. A dedicated battery mount was also designed to secure the battery tightly to the robot’s frame near its center of mass, minimizing any unwanted shift in mass distribution or moment of inertia during operation. The ESC's, which regulate EDF output based on PWM signals, were rigidly fixed to Harpy’s arms using 3D printed clips. This ensured both a symmetric mass distribution and effective wire management—keeping the power and signal wires clear of Harpy’s moving legs to prevent entanglement or interference during locomotion. A brief schematic of this configuration can be seen below.

A more detailed CAD model of the thruster mounting configuration can be seen below that shows where the threaded heat inserts live as well as the assembly stackup. Additionally, figure 2.3 inn section two displays the aerial assembly of Harpy with the location of the EDF's, ESC's, and the 6S LIPO battery. 

The plot below illustrates the maximum moment generated by the EDF thrusters relative to the center of mass (COM) of the robot, highlighting the strategic placement of the thrusters. This visualization emphasizes how the location of the thrusters contributes to the overall moment arm, aiding in roll stabilization and improving the robot’s balance and maneuverability.

\begin{figure}[H]
    \centering
    \includegraphics[width=0.75\linewidth]{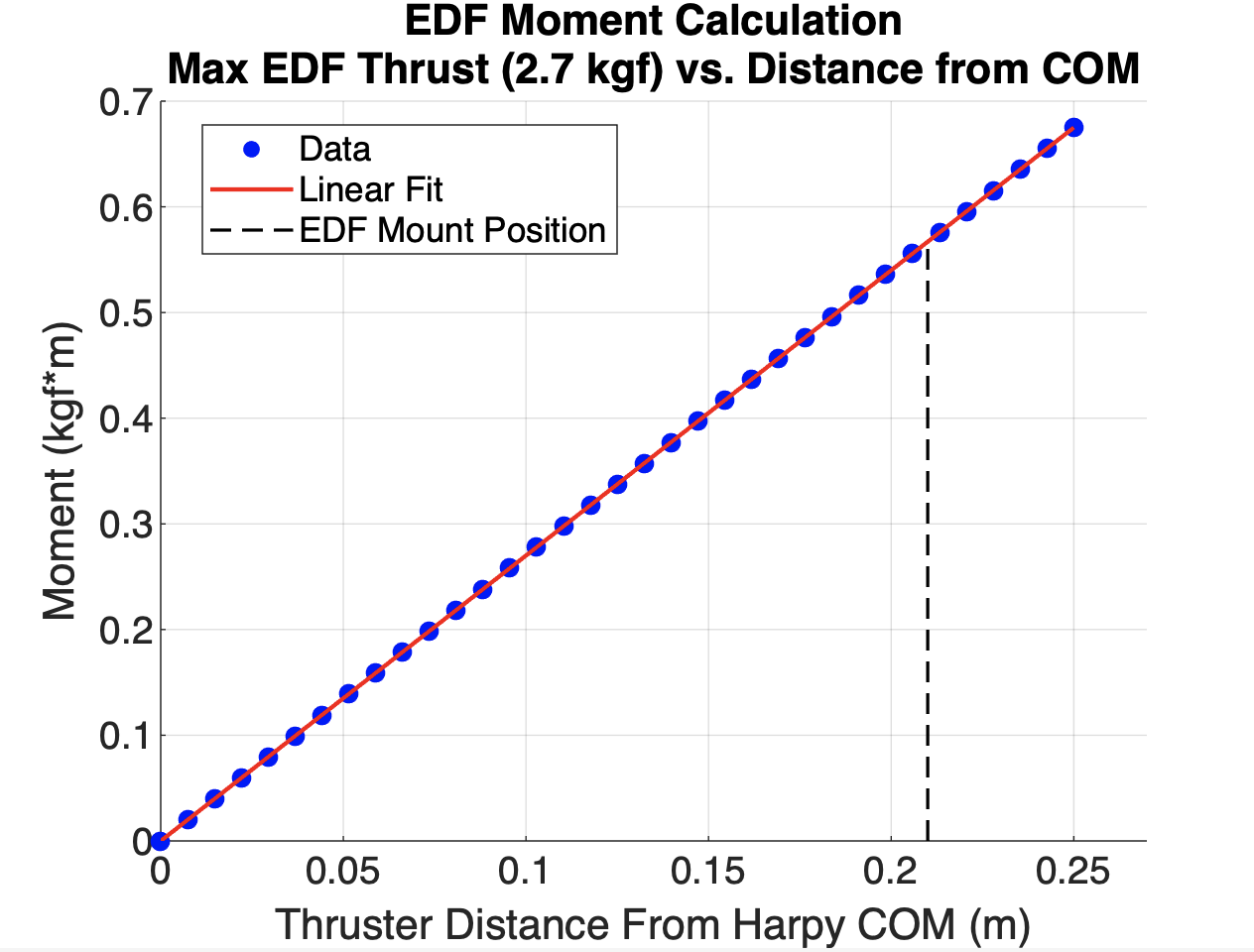}
    \caption{Moment vs. EDF Location from Harpy Center of Mass}
    \label{fig:enter-label}
\end{figure}

Using the maximum thrust of the EDF, which is 2.7~kgf (approximately 27~N), applied at a distance of 0.2~m from the robot's center of mass, the resulting moment is slightly over 0.5~kgf$\cdot$m. This moment is sufficient to counteract Harpy’s body torque during roll stabilization scenarios.

%% file: tex/Harpy_Motor_Charaterization.tex
\chapter{Harpy Motor Characterization}
\label{chap:motor-char}

To expand Harpy’s control capabilities beyond basic positional commands, impedance-based control was incorporated into the closed-loop system to allow for force-responsive interaction with the environment. A key step in this integration was determining the torque constant (N·m/A) of the KV380 motors, enabling current-based force control. This was achieved through a three-pronged investigation: analyzing the motor datasheet, conducting empirical tests with a torque sensor, and using the back-EMF method, where one motor is driven and the voltage output of another is measured via oscilloscope.

\section{Torque Study Overview}
\label{sec:torque-overview}

To expand the control capabilities of Harpy beyond its current reliance on purely positional control, it became necessary to incorporate impedance-based control into its closed-loop functionality. Impedance control enables the robot to interact more dynamically and compliantly with its environment by modulating force in response to position and velocity deviations. This control strategy is particularly valuable for applications requiring responsive mobility and nuanced physical interaction, such as terrain adaptation or contact-sensitive maneuvers. To implement this, a crucial step involved determining the torque constant (in N·m/A) of the KV380 motors used in Harpy. This torque constant acts as a bridge between electrical input and mechanical output, enabling force-based control utilizing current modulation. A comprehensive torque investigation was conducted using three distinct methods: analysis of the motor’s datasheet specifications, empirical testing using a torque sensor, and the back-EMF method, which involves driving one motor while measuring the generated voltage in another through its three-phase leads with an oscilloscope. Among these methods, the most accurate will yield the torque coefficient used in the final implementation. Incorporating this constant into the current control of the motors allows it to be seamlessly integrated into the closed-loop kinematic system, enabling full impedance-based control and enhancing Harpy’s interaction with its physical environment.

The T-Motor MN4006 KV380 motor used in the Harpy project is a high-performance UAV-grade brushless motor, well-suited for precision control applications. It features a 5mm shaft, which allows for straightforward coupling with a torque sensor and a second motor during back-EMF investigations. Structurally, the motor is composed of two main components: the stator, which remains stationary and houses the windings, and the rotor, which rotates and contains the permanent magnets. This motor is delta-wound, meaning the windings are connected in a triangular configuration, which offers higher current capacity and greater torque at lower speeds—ideal for responsive, high-load applications like Harpy.

Harpy’s design further enhances the integration of this motor through custom top and bottom housings that are 3D printed using carbon fiber filament. This not only provides exceptional strength-to-weight ratio but also enables advanced manufacturing techniques such as embedding bearings directly into the part during printing. This approach significantly improves structural rigidity while reducing overall system mass. These features together support high mechanical performance and precise control, making the motor an excellent candidate for torque-based impedance control. The MN4006 motor itself supports operation on 4–6S LiPo batteries, offers smooth low-noise operation, and is designed for stability and reliability in multi-rotor environments—making it a strong foundation for advanced robotic actuation.

After motor assembly was completed, this torque constant investigation was accomplished through a three-part investigation: examining the motor’s datasheet for manufacturer-provided specifications, performing hands-on measurements using a torque sensor, and applying the back-EMF method—where one motor is actively driven while the resulting voltage is captured from a second motor’s three-phase leads using an oscilloscope.

\subsection{Method 1: Spec Sheet Torque Estimation}
\label{subsec:method-1}

The KV380 motor is a type of brushless DC motor characterized by its speed constant, which defines how quickly the motor spins per volt applied under no-load conditions. This speed constant, typically expressed in revolutions per minute per volt (RPM/V), serves as a useful metric for estimating motor performance. However, for theoretical analysis and modeling, it is often converted into angular velocity units to better align with standard physical equations. This conversion facilitates a more precise understanding of the motor's dynamic behavior in control and simulation environments.

The torque constant, which represents the amount of torque a motor produces per ampere of current, is closely related to the speed constant. In an ideal system, the torque constant is the inverse of the speed constant when both are expressed in appropriate SI units. This relationship forms the basis for estimating torque output from electrical input and is central to motor modeling and performance prediction. However, this simplified relationship holds only under idealized assumptions and neglects several practical factors, including the motor's electromagnetic characteristics, winding configuration, and losses due to resistance and friction.

\begin{equation}
K_v = 380~\left[\frac{\text{RPM}}{\text{V}}\right] = 39.79~\left[\frac{\text{rad}}{\text{s} \cdot \text{V}}\right]
\end{equation}

\noindent where:
\begin{itemize}
    \item $K_v$ is the motor velocity constant, representing the no-load angular velocity per applied volt.
    \item $\text{RPM}$ is the rotational speed in revolutions per minute.
    \item $\text{V}$ is the input voltage applied to the motor.
    \item $\text{rad/s}$ is the angular velocity in radians per second, where $1~\text{RPM} = \frac{2\pi}{60}~\text{rad/s}$.
\end{itemize}

To obtain a more accurate theoretical value for the torque constant, detailed motor design parameters must be considered. These include the winding configuration, number of pole pairs, magnetic flux linkage, and core material properties. A series of foundational electromagnetic equations are used to describe these relationships, linking the motor's physical construction to its electrical characteristics. This theoretical framework provides a deeper understanding of how the motor generates torque and how its design influences performance.

In order to validate the theoretical torque constant, empirical testing is performed using two distinct methods. The first involves directly measuring the torque output at various current levels using a test bench or dynamometer. This provides a practical, real-world value of the torque constant that includes losses due to friction, resistance, and other non-ideal factors. The second method involves measuring the back electromotive force (back-EMF) generated when the motor is spun at a known speed. From this measurement, the back-EMF constant is calculated and used as an indirect estimate of the torque constant. Comparing these empirical results with the theoretical model allows for an evaluation of the motor's efficiency and accuracy of the assumptions made during theoretical derivation.

In order to convert the three phase motor quantities into a two-axis system(direct and quadrature, one can do so by correlating the rotors rotating magnetic field. The Clark transformation converts 3-phase quantities into two orthogonal components. Furthermore, the park transformation then rotates these two orthogonal components obtained from the Clarke transformation to align with the rotors magnetic field. These transformation methods and math were derived using methods from a well known textbook written by M. Spong \cite{spong_robot_1989}. 

\begin{equation}
C =
\begin{bmatrix}
\cos(\theta) & \sin(\theta) \\
-\sin(\theta) & \cos(\theta)
\end{bmatrix}
\end{equation}

\noindent where:
\begin{itemize}
    \item $C$ is the 2D rotational transformation matrix (also known as the direction cosine matrix),
    \item $\theta$ is the angle of rotation, measured in radians,
    \item $\cos(\theta)$ and $\sin(\theta)$ represent the cosine and sine of the rotation angle, respectively.
\end{itemize}

The Clarke transformation, shown in Eq.~\ref{eq:clarke}, is widely used in control systems for electric motors, especially in three-phase configurations such as brushless DC (BLDC) and AC induction motors. Its primary purpose is to project three-phase quantities $(a, b, c)$ onto a two-axis orthogonal coordinate system $(\alpha, \beta)$, which simplifies the analysis and control of motor behavior. This transformation reduces the dimensional complexity and enables vector control techniques by working within a stationary reference frame, facilitating tasks such as torque regulation and field-oriented control (FOC)~\cite{spong_robot_1989}.

\begin{equation}
\label{eq:clarke}
P = \sqrt{\frac{2}{3}}
\begin{bmatrix}
1 & -\frac{1}{2} & -\frac{1}{2} \\
0 & \frac{\sqrt{3}}{2} & -\frac{\sqrt{3}}{2}
\end{bmatrix}
\end{equation}

The Park transformation, shown in Equation~\ref{eq:park}, maps the stationary two-phase system $(\alpha, \beta)$—typically obtained via the Clarke transformation—into a rotating reference frame $(d, q)$ aligned with the rotor’s magnetic field. This approach transforms AC quantities into DC-like signals in the rotating frame, greatly simplifying control tasks and improving dynamic performance in FOC applications~\cite{spong_robot_1989}.

When both transformations are applied in sequence, the result is a composite transformation matrix $P \cdot C$ that maps three-phase inputs directly into the rotating $(d, q)$ frame. The Clarke matrix $P$ projects three-phase signals onto the $(\alpha, \beta)$ plane, while the Park matrix $C$ rotates this plane into alignment with the rotor flux. This combined operation simplifies control of torque-producing currents and enables the use of PI controllers for DC signals in FOC.

\begin{equation}
\label{eq:park}
P \cdot C = \sqrt{\frac{2}{3}}
\begin{bmatrix}
\cos(\theta) & \cos(\theta - \frac{2\pi}{3}) & \cos(\theta + \frac{2\pi}{3}) \\
-\sin(\theta) & -\sin(\theta - \frac{2\pi}{3}) & -\sin(\theta + \frac{2\pi}{3})
\end{bmatrix}
\end{equation}

This transformation can be applied to either the phase currents or phase voltages. In the context of motor characterization, especially when determining the torque constant using back-EMF or datasheet values, the phase voltages are typically used.

\begin{equation}
\begin{bmatrix}
I^d \\
I^q
\end{bmatrix}
= P \cdot C \cdot I^{\phi}
= \sqrt{\frac{2}{3}}
\begin{bmatrix}
\cos(\theta) & \cos(\theta - \frac{2\pi}{3}) & \cos(\theta + \frac{2\pi}{3}) \\
-\sin(\theta) & -\sin(\theta - \frac{2\pi}{3}) & -\sin(\theta + \frac{2\pi}{3})
\end{bmatrix}
\begin{bmatrix}
I^\phi_A \\
I^\phi_B \\
I^\phi_C
\end{bmatrix}
\end{equation}

When focusing exclusively on the $q$-axis current contribution to torque, the transformation simplifies as:

\begin{equation}
I^q = \sqrt{\frac{3}{2}} I^\phi
\end{equation}

This leads to the expression for BLDC motor torque in the $q$-axis as follows:

\begin{equation}
\tau_q = K_t \cdot \sqrt{\frac{3}{2}} I^\phi \cdot \sqrt{\frac{3}{2}} j = \sqrt{\frac{3}{2}} I^\phi
\end{equation}

Thus, the $q$-axis torque constant can be expressed as:

\begin{equation}
K^q_t = \sqrt{\frac{3}{2}} K_t
\end{equation}

From the motor's datasheet, we can relate the torque constant $K_t$ to the back-EMF constant $K_v$ using:

\begin{equation}
K_b = \frac{1}{K_v} = 0.025~\left[\frac{\text{s} \cdot \text{V}}{\text{rad}}\right]
\end{equation}

In the $q$-axis reference frame, the torque constant for a delta-wound BLDC motor becomes:

\begin{equation}
K_{qt} = \sqrt{\frac{3}{2}} K_b = 0.0307~\left[\frac{\text{N} \cdot \text{m}}{\text{A}}\right]
\end{equation}

This $K_{qt}$ value is used in subsequent analyses (Section~\ref{subsec:Results}) to determine the torque constant experimentally via the back-EMF method.

\subsection{Method 2: Back EMF Torque Coefficient Method}
\label{subsec:method-2}

In three-phase motors, the back electromotive force (EMF) is intrinsically linked to the motor's torque constant, providing a valuable method for torque measurement. The back EMF constant (Ke) and the torque constant (Kt) are numerically equal when expressed in consistent units, differing only in their dimensional representations; both are measured in volt-seconds per radian (V·s/rad) and newton-meters per ampere (N·m/A), respectively.

In the experimental setup, two KV380 motors were mounted on the same fixture utilized in the torque sensor method seen in Fig.\ref{fig:back-emf-testbed} below. Using oscilloscope probes, the peak-to-peak voltage amplitude of the three-phase output signals from the motors was measured. By applying the known relationship between back EMF and torque constant, and considering the constant rotational velocity set by the Elmo Studio system,  the torque constants for the observed motor at various speeds was calculated. This approach allowed for a comprehensive analysis of the motors' performance characteristics across different operational conditions.

\begin{figure}[H]
    \centering
    \begin{subfigure}{0.75\linewidth}
        \centering
        \includegraphics[width=0.7\linewidth]{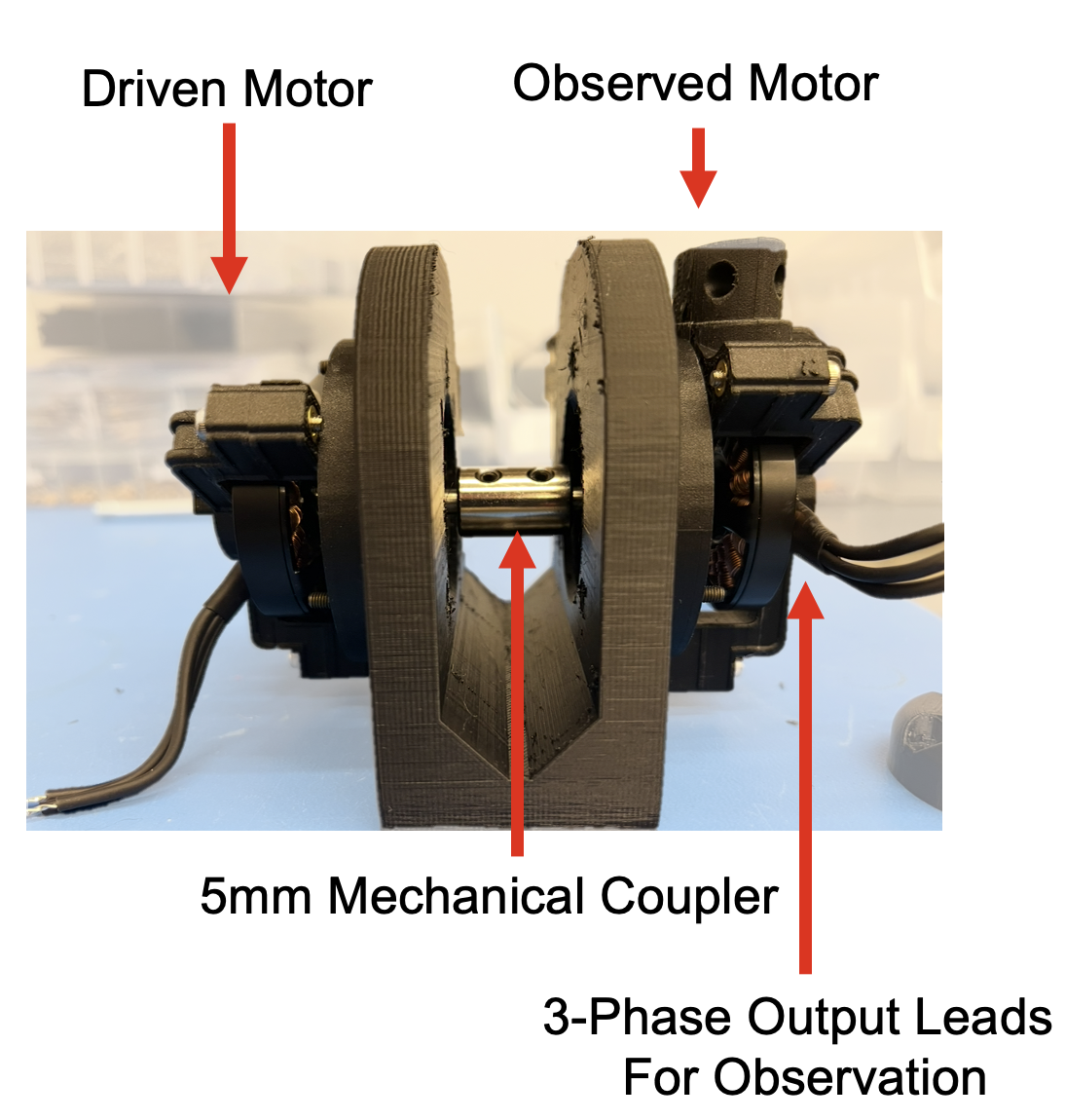}
        \caption{Actual testbed used for evaluating the motor constant via the Back EMF method.}
        \label{fig:Dual_motor_testbed_Image}
    \end{subfigure}
    
    \vspace{0.5em} 

    \begin{subfigure}{0.75\linewidth}
        \centering
        \includegraphics[width=0.99\linewidth]{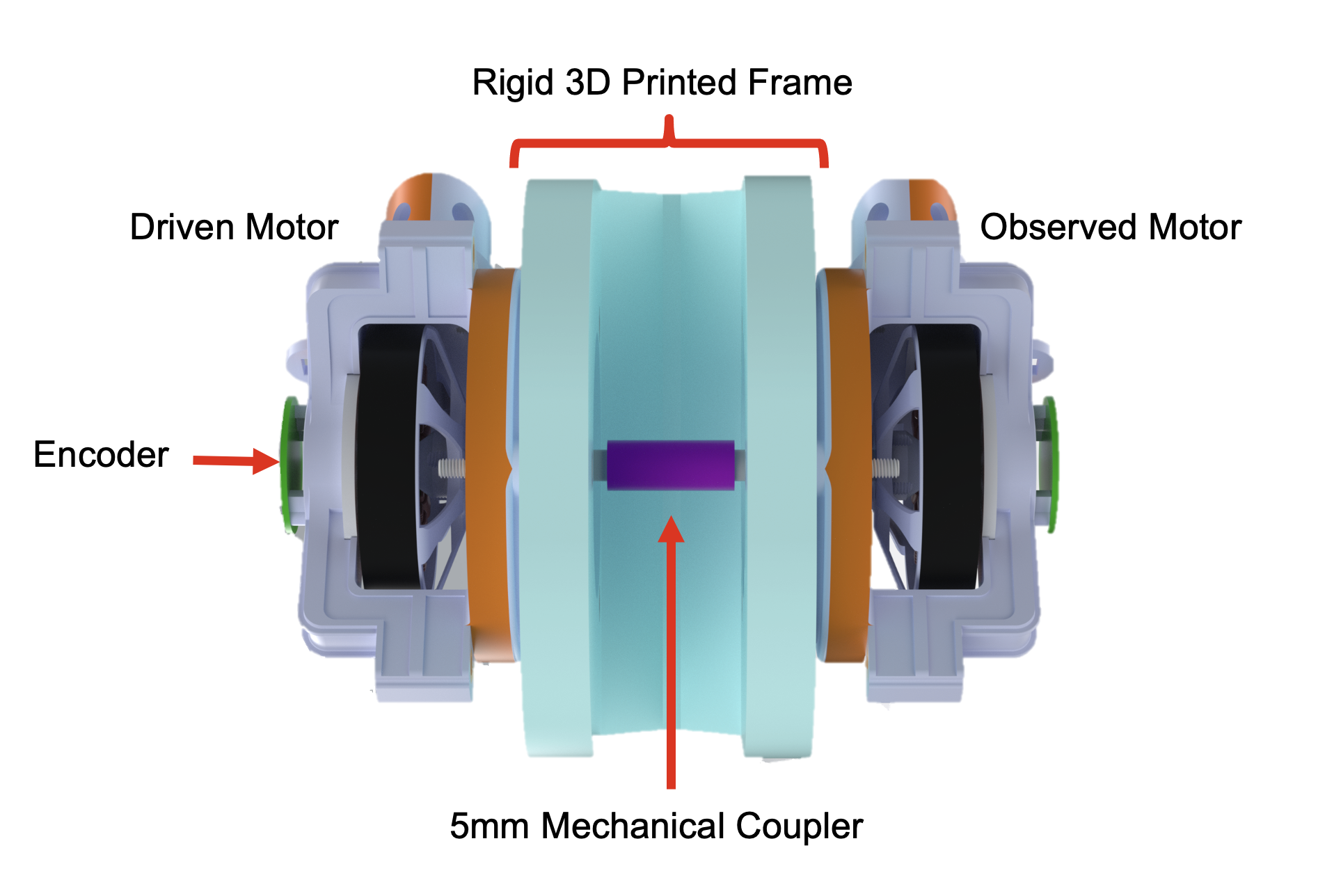}
        \caption{CAD model of the testbed used for Back EMF-based motor constant evaluation.}
        \label{fig:back-emf-testbed}
    \end{subfigure}
    
    \caption{Back EMF motor constant evaluation setup: Physical testbed (top) and CAD representation (bottom).}
    \label{fig:back_emf_combined}
\end{figure}

As described through Eqs.~\ref{eq:clarke} and \ref{eq:park} previously to convert three phase motor quantities into a two-axis system,  the back-EMF constant can be defined using the same clarke matrix transformation. The constant out from the Kb term is again defined by taking a single direction of the matrix.

\begin{equation}
\label{eq:FinalForm}
K^q_b = \sqrt{\frac{3}{2}}K_b
\end{equation} 

In Eq.~\ref{eq:FinalForm} above, the phase back-EMF constant and phase torque constant are identical to one another.

By controlling the driven motor at a known input velocity, and monitoring the peak to peak voltage of the three phase output on the oscilloscope, those two inputs can be used to calculate the back EMF constant, which is inversely proportional to the motor torque constant. An example of a peak to peak voltage observation from a three phase motor, similar to what the output of this motor showed, can be seen in \ref{fig:osc-observation} below.

\begin{figure}[htbp]
    \centering
\includegraphics[width=0.5\linewidth]{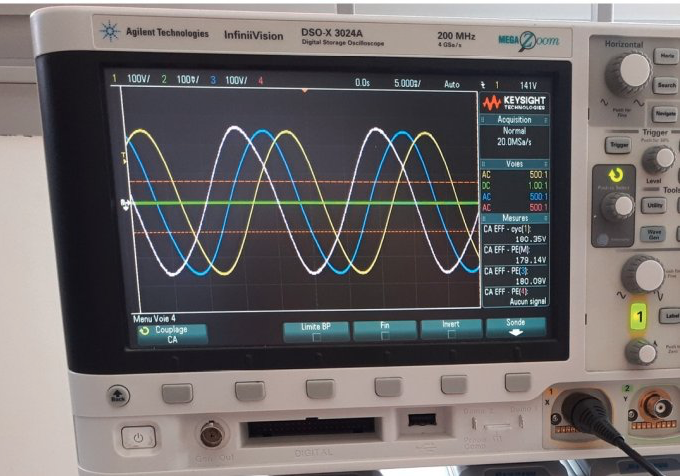}
    \caption{Three Phase Motor Oscilloscope Observation.}
    \label{fig:osc-observation}
\end{figure}

The motor's torque constant, denoted as Kt is a fundamental parameter that describes the relationship between the input current and the resulting torque output. In a three-phase motor, the Kt value is often derived from the motor's back EMF, which is proportional to both the velocity of the motor and the torque being generated. Essentially, kt serves as a scaling factor that allows us to convert the motor's input current (measured in amps) into the corresponding torque output (measured in Newton-meters).

In our setup, the input to this calculation is the peak-to-peak voltage observed from the motor’s three-phase output, which is measured using an oscilloscope. This voltage is directly related to the motor's back EMF, which increases with the motor's speed. By knowing the motor’s rotational velocity (which can be calculated from the encoder data), we can use this voltage measurement as part of the process to determine the torque constant. Since the back EMF constant is proportional to the motor’s torque constant, the peak-to-peak voltage is used as a relative term in the calculation of Kt.

To calculate the torque constant kt, the voltage amplitude (peak-to-peak) observed on the oscilloscope is used in combination with the motor’s velocity. This velocity, which is measured in radians per second, is provided by the encoder, and the relationship between voltage, velocity, and torque is used to derive the torque constant in N·m/amp. By performing this calculation at different velocities and corresponding voltage amplitudes, we can determine the value of Kt for the motor, providing us with a direct conversion factor that links the input current to the output torque.

Thus,Kt becomes the key parameter that translates the observed voltage (a representation of the motor's back EMF) into the torque generated by the motor in response to that voltage, but ultimately expressed in terms of current (amps). It serves as a conversion factor that relates the motor's electrical input to its mechanical output, allowing for the torque constant to be determined as N·m/amp.

There were 10 trials that were done using the back EMF method, some repeating at the same input amperage. This plot can be seen below in Fig.\ref{fig:back-emf-plot}.

\begin{figure}[H]
    \centering
    \includegraphics[width=0.75\linewidth]{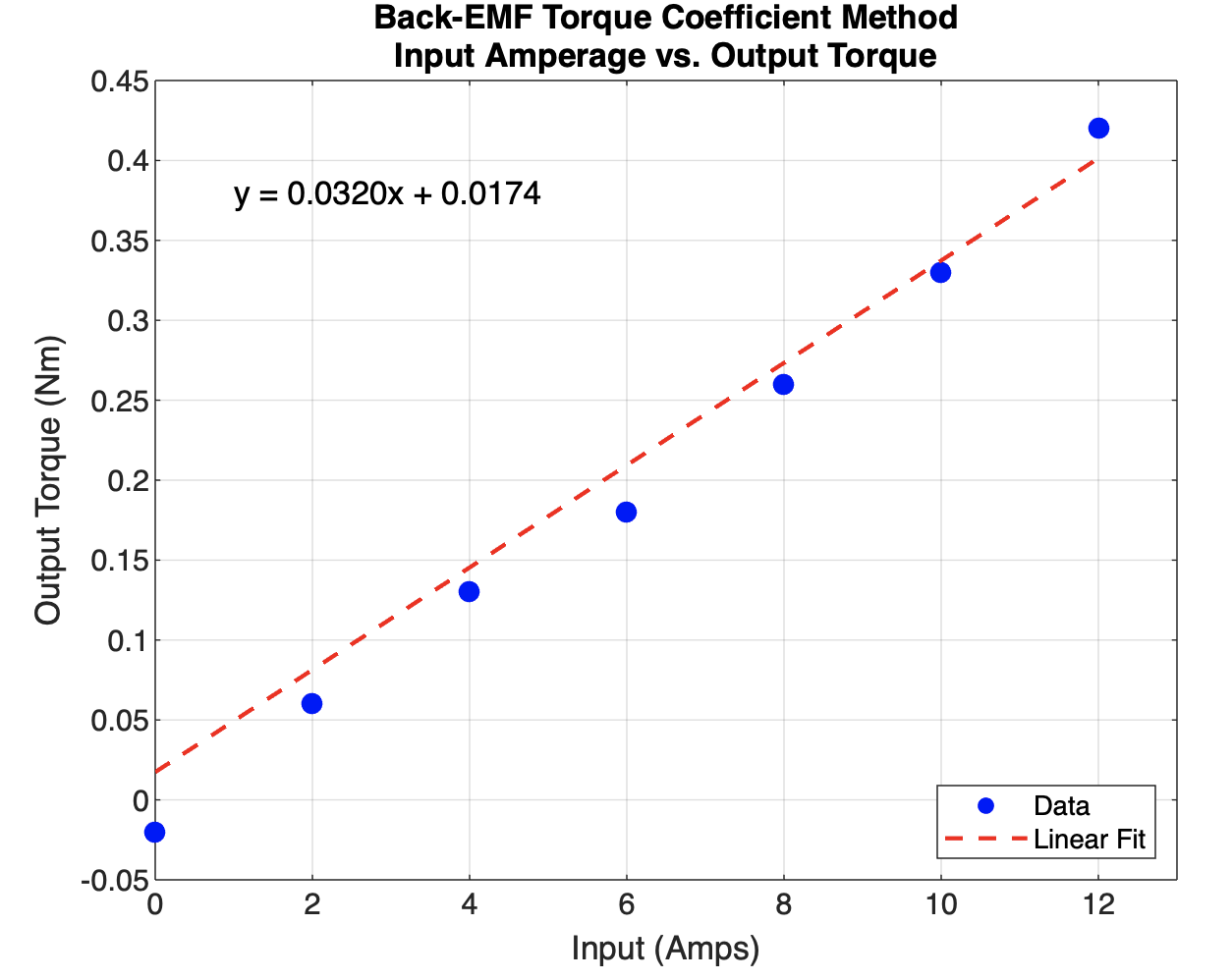}
    \caption{Back EMF Torque Constant Plot}
    \label{fig:back-emf-plot}
\end{figure}

The plot, along with the summary of the Back EMF proportional trials in relation to the torque constant, will be discussed in more detail in the results Section~\ref{subsec:Results}, where the findings and their implications will be further elaborated and analyzed.

\subsection{Method 3: Torque Sensor}
\label{subsec:method-3}

To empirically measure the torque output of the KV380 motor, a precision micro reaction torque sensor was used, selected specifically for its 0–2 N·m range. This range was chosen to align closely with the motor’s expected maximum torque output of approximately 0.5 N·m, providing both adequate resolution and measurement headroom. The sensor, which comes factory-calibrated, offers excellent accuracy and repeatability, making it highly suitable for capturing small, consistent torque variations in experimental setups. It is designed to provide reliable measurements in real time and has a proven calibration curve that relates its electrical output directly to physical torque values.

\begin{figure}[H]
    \centering
    \begin{subfigure}{0.75\linewidth}
        \centering        \includegraphics[width=0.6\linewidth]{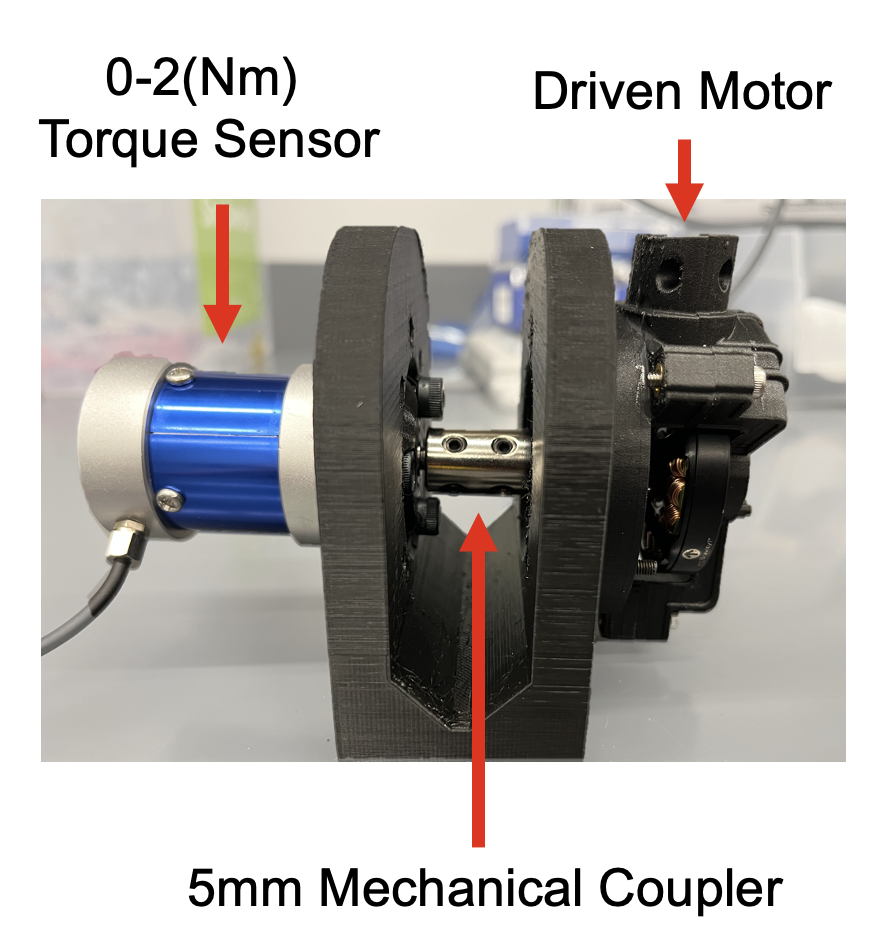}
        \caption{Torque Sensor and Motor Setup (Actual testbed).}
        \label{fig:actual_torque_sensor_Setup}
    \end{subfigure}
    
    \vspace{0.5em} 

    \begin{subfigure}{0.75\linewidth}
        \centering
        \includegraphics[width=0.99\linewidth]{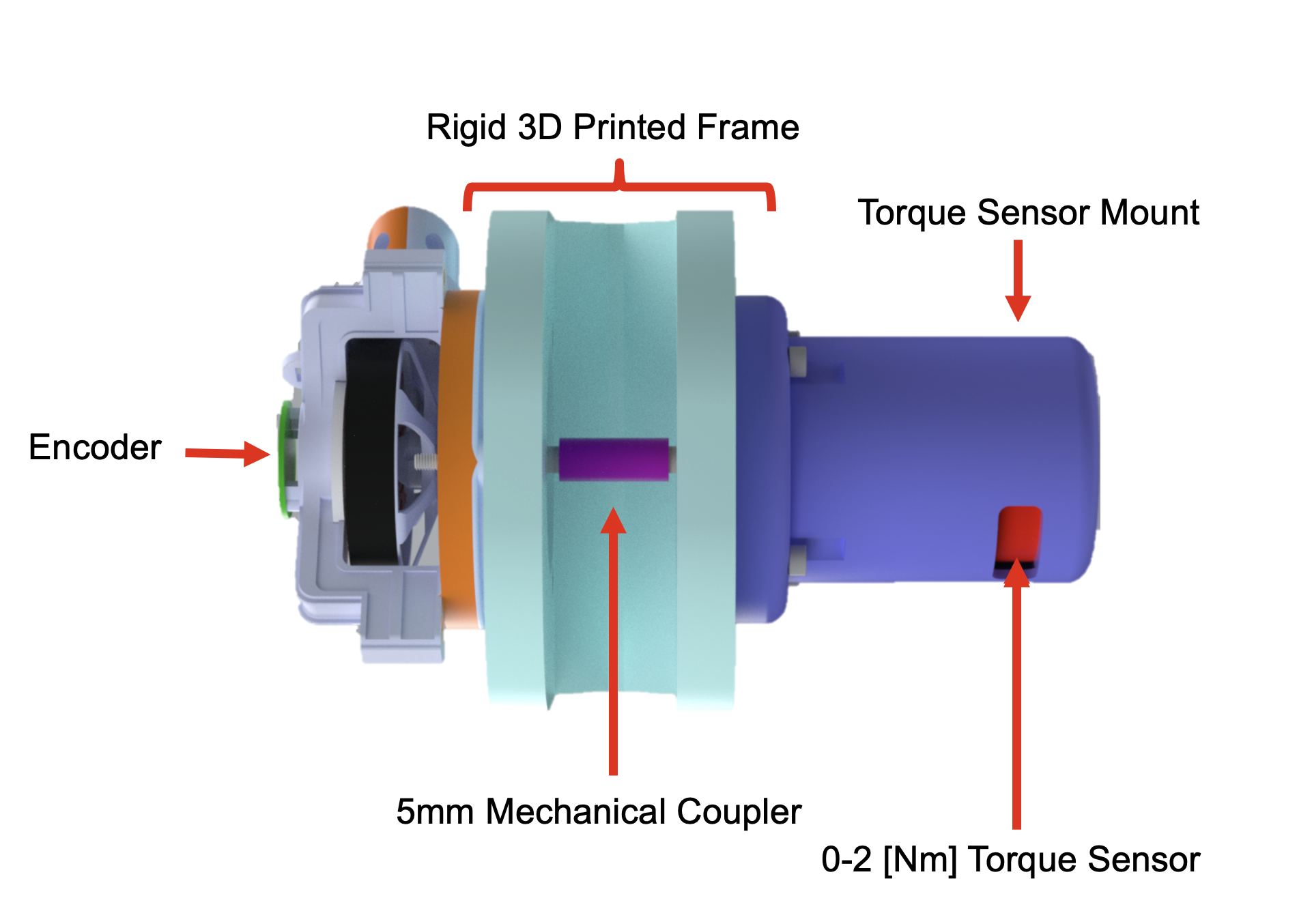}
        \caption{Schematic of the Torque Sensor and Motor Setup.}
        \label{fig:torque-sens-testbed}
    \end{subfigure}
    
    \caption{Torque coefficient testing setup: Actual hardware (top) and Schematic (bottom).}
    \label{fig:torque_sensor_combined}
\end{figure}

To integrate the sensor into the data acquisition system, a load cell transmitter was paired with it—specifically, a unit that converts the torque sensor’s signal into a standardized 4–20 mA current output. This transmitter ensures stable and noise-resistant signal transmission, particularly valuable in electrically noisy environments such as motor testing setups. The transmitter is also compatible with the sensor's output specifications and provides a clear analog signal proportional to the torque applied. Using the manufacturer-provided calibration curve, this current signal can be back-calculated to determine the corresponding torque in N·m.

\begin{figure}[htbp]
  \centering
  \begin{subfigure}[b]{0.45\textwidth}
    \raggedright
    \includegraphics[width=1.2\linewidth]{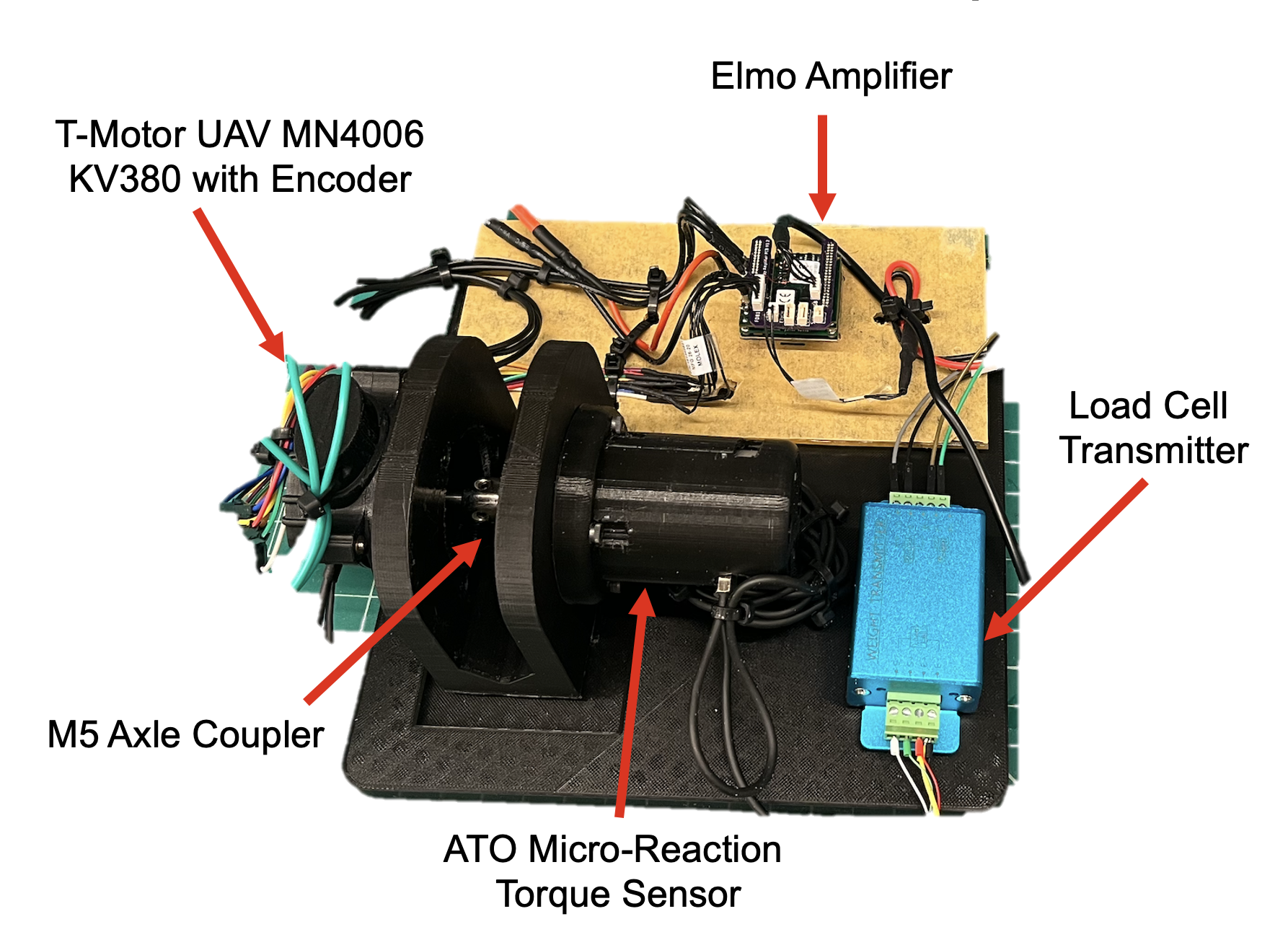}
    \caption{Actual Torque Sensor Setup}
    \label{fig:actual-torque-sensor-setup}
  \end{subfigure}
  \hfill
  \begin{subfigure}[b]{0.45\textwidth}
    \raggedleft
    \includegraphics[width=0.9\linewidth]{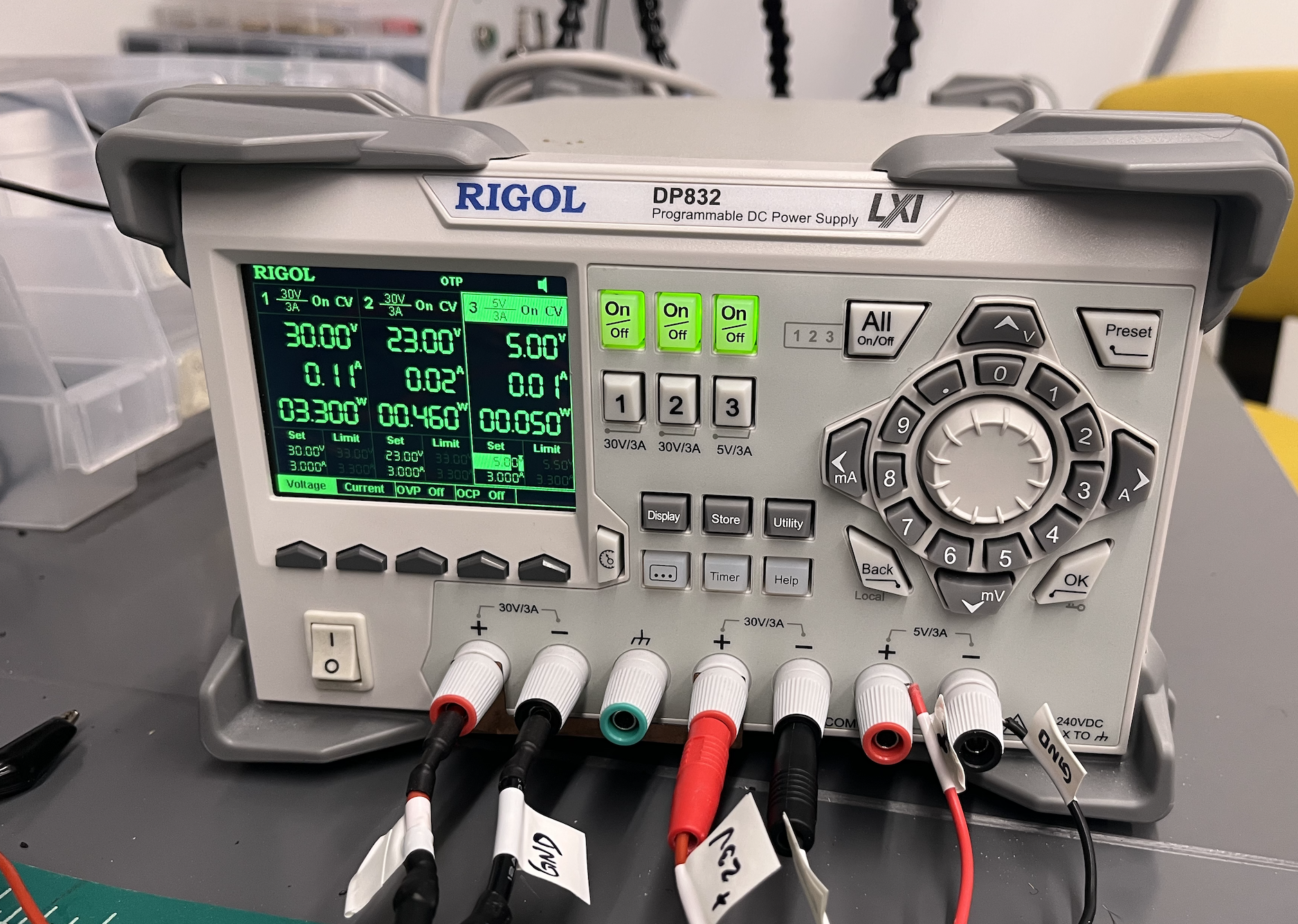}
    \caption{Power Supply Inputs Overview}
    \label{fig:power-supply-inputs}
  \end{subfigure}
  \caption{Overview of Torque Sensor Setup and Power Supply Inputs.}
  \label{fig:torque-sensor-overview}
\end{figure}

In this setup, three components require power. Two of these components are connected directly to the Elmo amplifier, which provides the necessary power to the system. Specifically, the amplifier supplies 5V to the emergency stop (e-stop) circuit, and 30V to the amplifier itself, which in turn powers the three-phase motor. The third power input is dedicated to the torque sensor, which operates within a voltage range of 15V to 30V. After reviewing the specifications provided by the vendor, a mid-range value of 23V was selected for optimal performance.

To manage these power requirements, a three-channel power supply was employed, as shown in Fig.~\ref{fig:torque-sensor-overview}. The power supply used is a Rigol three-channel model, known for its reliability and precision. Each channel is responsible for powering a specific component of the system. For ease of maintenance and to ensure stable connections, all components were securely mounted on a custom 3D-printed plate. This rigid mounting setup not only facilitates easier wire management but also ensures that all electrical connections remain intact and reliable throughout testing. Figure \ref{fig:torque-sens-testbed} below displays similar information but in a schematic format for ease of following connections and outputs.

\begin{figure}[H]
    \centering
    \includegraphics[width=0.95\linewidth]{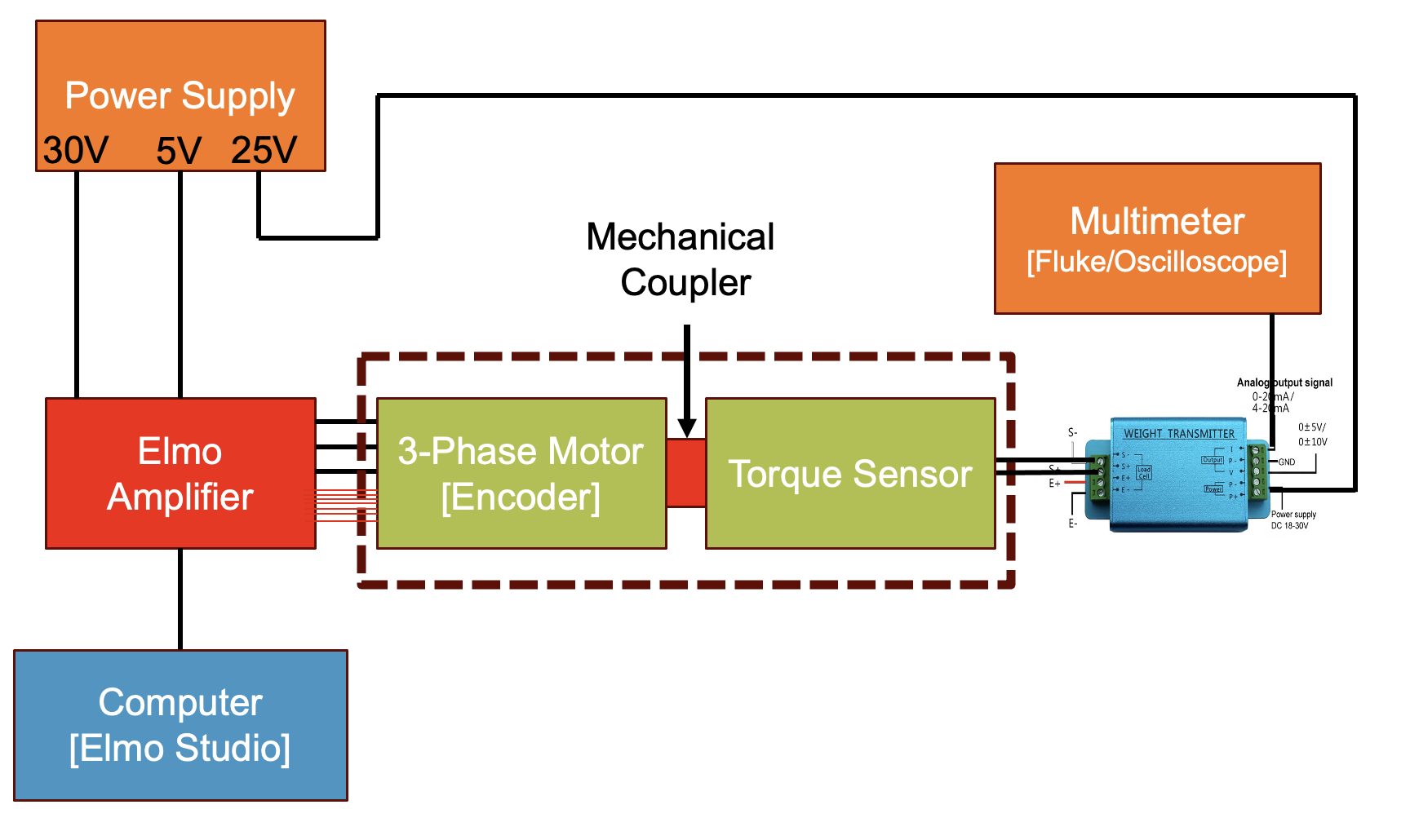}
    \caption{Torque Sensor Motor Characterization Schematic}
    \label{fig:torque-sens-testbed}
\end{figure}

To measure the voltage output from the load cell transmitter, a resistive probe and a Rigol oscilloscope were employed due to their superior measurement resolution compared to other tools in the lab. The Fluke multimeters available did not provide the necessary precision to accurately capture the low-range voltage fluctuations from the load cell output. The Rigol oscilloscope, with its high resolution, was the ideal choice for this task. By connecting a resistive probe to the oscilloscope and knowing its resistance, a simple calculation using Ohm's Law, $V = IR$, was performed to determine the output current. The voltage measured by the oscilloscope was monitored as the output current varied within the 4--20~mA range of the load cell transmitter. With the known resistance and the voltage readings from the oscilloscope, the corresponding current could be accurately calculated, allowing for precise tracking of the load cell's output as it responded to changing torque. This method ensured a high degree of accuracy in measuring the current and analyzing the sensor's performance.

With this setup, the real-time torque generated by the motor was recorded as a function of the input current supplied by the motor controller. The input current values were obtained directly from the Elmo Studio motor control software, which logs precise current draw data during operation. By pairing each torque measurement from the sensor system with the corresponding motor input current, a torque-versus-current curve was generated. The slope of this curve represents the motor’s torque constant, expressed in N$\cdot$m/A. This method provided a practical and accurate way to determine the empirical torque constant of the KV380 motor under controlled conditions, enabling direct comparison with theoretical values derived from the motor’s electrical parameters. The data in Figure~5.7 below is the calibration information provided by ATO, the company from which the torque sensor and load cell transmitter were purchased. Since the output of the load transmitter is given in amps, utilizing the calibration curve below, the output torque of the motor can be calculated in N$\cdot$m.

\begin{figure}[H]
    \centering
    \includegraphics[width=0.75\linewidth]{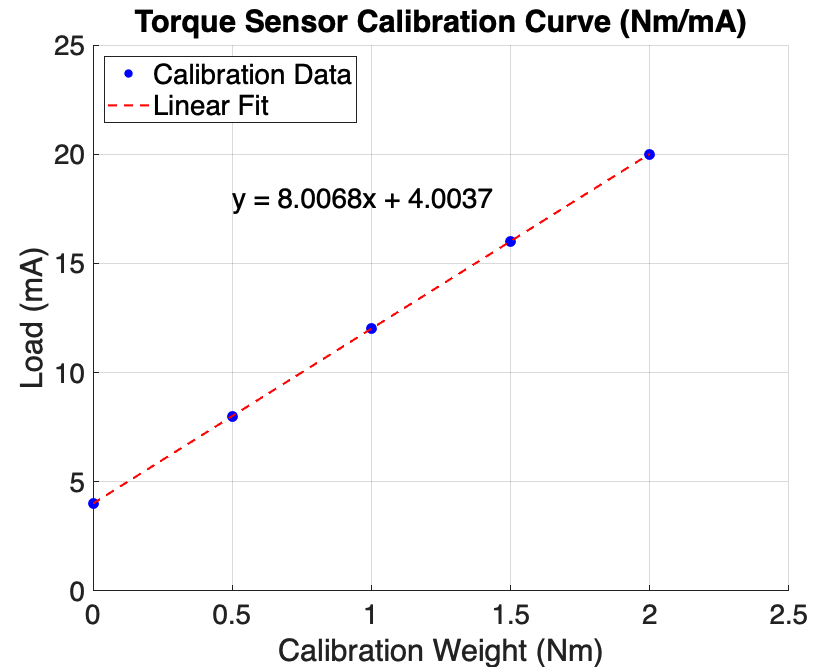}
    \caption{Torque Sensor Calibration Curve}
    \label{fig:torque-sens-curve}
\end{figure}
Once the torque was back-calculated for each trial of the 28 trials of input currents and output torques, the plot below was created for each trial, where the input motor current can be see on the x-axis and the output torque in n*m from the torque sensor is plotted. From there, a linear relationship can be seen describing the plot, where the slope is the average relation between the input motor amperage and output torque. Again this was done for all 28 trials, and the average slope value of all of these trials was utilized for a final result comparison for all torque methods used.

\begin{figure}[H]
    \centering
    \includegraphics[width=0.75\linewidth]{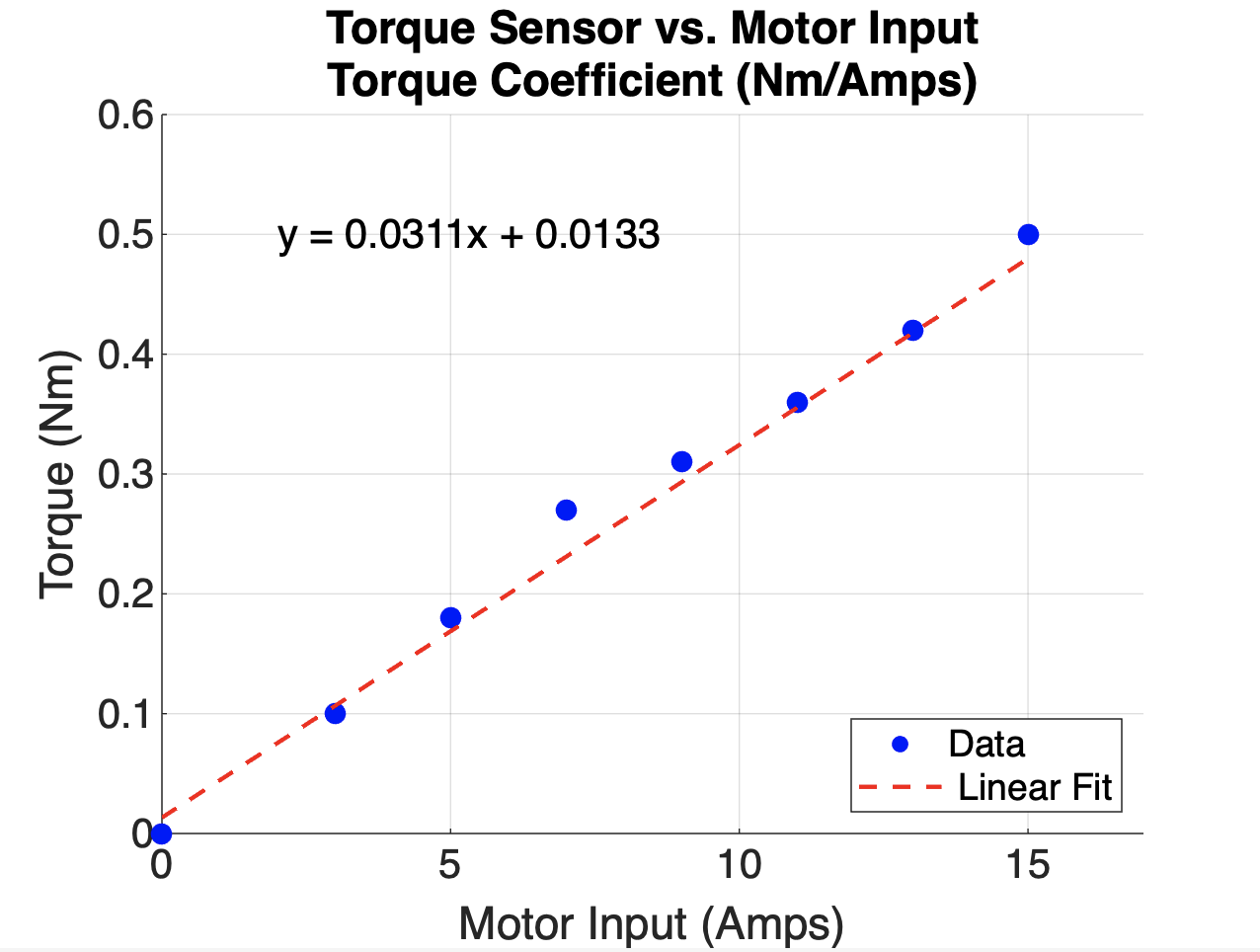}
    \caption{Torque Sensor Data (Coefficient investigation Method 3)}
    \label{fig:torque-sens-data}
\end{figure}

The results of all 28 trials and the comparison of the torque sensor method trials to the two other investigation methods are described in Section~\ref{subsec:Results}. 

\section{Motor Torque Coefficient Results}
\label{subsec:Results}

In this investigation, three methods were employed to measure and analyze the torque of the KV380 motor. The first method involved utilizing the motor's specification sheet, which provided key details on the motor's performance characteristics, including its rated torque and voltage parameters. This information served as a baseline for understanding the motor's expected behavior under specific conditions. The second method, known as the Back EMF method, relied on the fact that the back EMF constant of the motor is directly proportional to the torque constant at a given velocity or acceleration. By observing the peak-to-peak voltage of the motor's output and knowing the motor's velocity, the torque constant was derived for different operational speeds. The third method used an off-the-shelf calibrated torque sensor and a weight transmitter to measure the load cell outputs. 

\begin{figure}[H]
    \centering
    \includegraphics[width=0.75\linewidth]{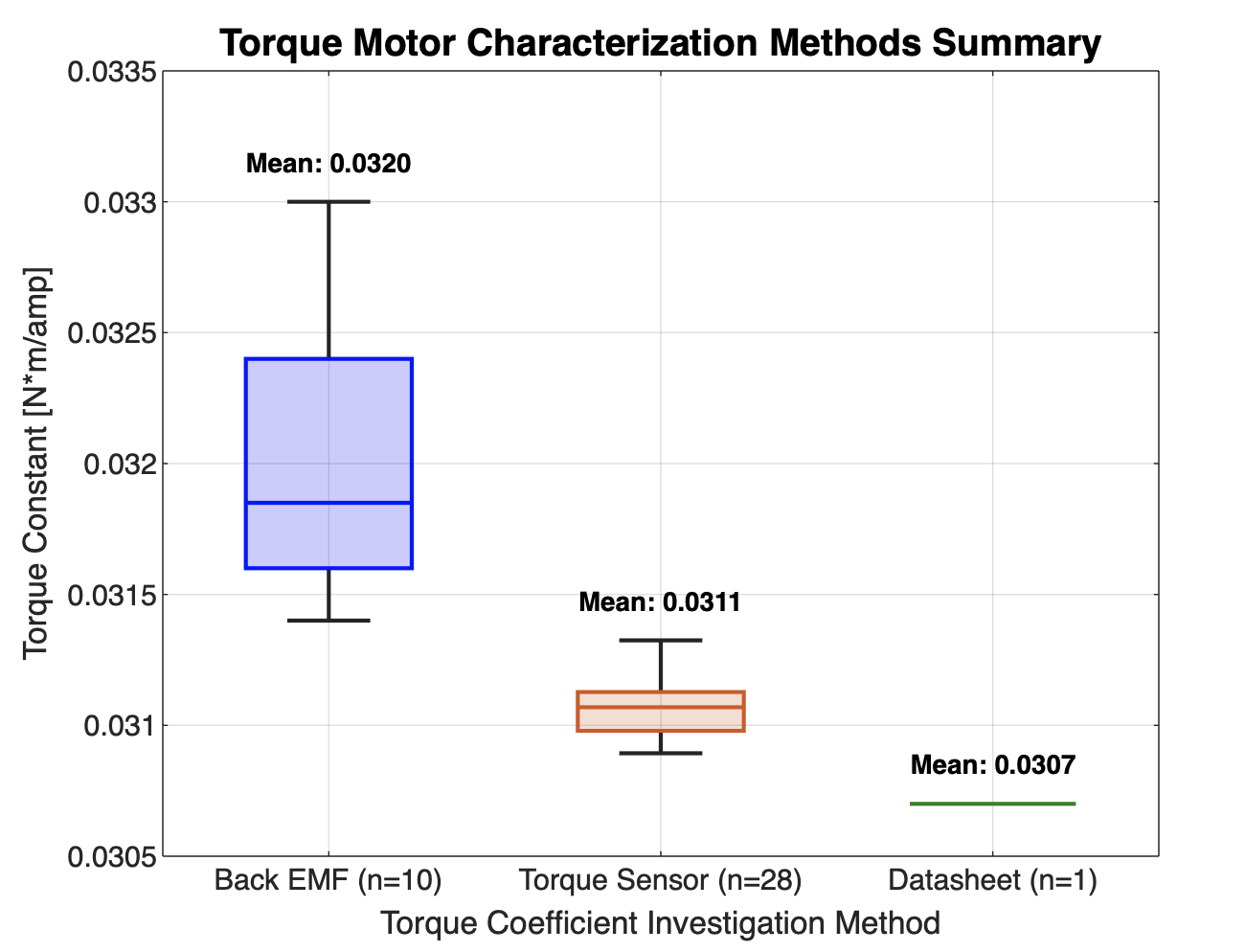}
    \caption{Torque Constant Investigation Methods Comparison}
    \label{fig:torque-coeff-comparison}
\end{figure}

This provided empirical data that could be directly compared to the other methods. To validate the findings, 10 trials using the Back EMF method were compared to 28 trials using the torque sensor method. The results of these trials, including the sensor's single output as specified in the datasheet, are summarized in Fig.~\ref{fig:torque-coeff-comparison}, highlighting the effectiveness and accuracy of each measurement technique.

The three methods used to measure the torque of the KV380 motor each have their own inherent sources of error, with the Back EMF method being particularly susceptible to inaccuracies. This method relies on measuring the peak-to-peak voltage of the motor’s back electromotive force (EMF) to estimate the torque. However, several factors can introduce errors in this process. One of the main issues is human error in reading the oscilloscope measurements, which can lead to inconsistencies in the data. Furthermore, noise in the three-phase signal complicates the task of obtaining consistent peak-to-peak voltage readings. This problem is exacerbated by the manual nature of the measurements, where fluctuations in the back EMF signal make it difficult to achieve stable results. Additionally, research has shown that back EMF measurements can be prone to errors arising from sensor nonlinearity, thermal variations, and inaccuracies in analog-to-digital conversion, all of which can distort rotor position and speed estimates, thus impacting torque calculations.

In contrast, the torque sensor method, which uses a calibrated sensor and weight transmitter, offers a more direct and reliable way to measure torque. This approach significantly reduces the measurement errors associated with the Back EMF method, as the sensor is calibrated to provide more consistent and accurate results. The torque sensor method also shows much smaller three-sigma error bars, indicating higher precision and less variability in the data. This is reflected in the boxcar plot, where the torque sensor method yields more accurate measurements, even though all three methods fall within five sigma of the Back EMF method. The superior accuracy of the torque sensor method is largely due to the larger number of trials conducted, which helps mitigate errors, as well as the reduced impact of signal noise compared to the Back EMF method. Overall, while the Back EMF method is useful for estimating torque, it is more prone to errors, and the torque sensor method proves to be the most reliable and precise for this application. The mean value of the torque sensor, which was determined to be 0.0311~N$\cdot$m/A, will be used in the next step of the motor characterization project and incorporated into the force-impedance-based control. This will serve as a crucial parameter in improving the motor's performance and control strategies moving forward.

%% file: tex/conclusion.tex
\chapter{Conclusion}
\label{chap:conclusion}

Bipedal robots continue to push the boundaries of robotic mobility, enabling machines to traverse complex, uneven terrains that would challenge traditional wheeled or tracked systems. Among these, Harpy stands out as a uniquely capable platform that integrates thrust-based augmentation with legged locomotion to overcome obstacles that have historically hindered bipedal robots—particularly rapid traversal, balance on inclines, and recovery from instability. One of the major challenges in implementing thrust augmentation was the lack of a comprehensive theoretical thrust model that could accurately relate RPM, pitch, blade diameter, and external ambient conditions to real-world thrust output. This gap in literature motivated a detailed thrust investigation, combining theoretical modeling, ANSYS simulations, and empirical testing. These efforts converged to validate the EDF configuration now implemented on Harpy.

To ensure reliable and efficient integration, off-robot testing of the EDF and ESC was conducted to characterize performance under controlled conditions. A custom carbon fiber mounting system was developed, providing a rigid and lightweight interface between the EDFs and the robot chassis. Furthermore, strategic placement of the ESCs and battery was carefully planned to preserve Harpy's center of mass, ensuring the robot's dynamic balance and agility were not compromised.

In addition to thrust characterization, a comprehensive analysis of Harpy’s joint motor torques was performed using multiple methods. This included leveraging manufacturer motor spec sheets, direct measurements via a torque sensor, and indirect estimation using back EMF analysis. All three methods were compared for consistency and accuracy, offering robust validation across different torque coefficient models. To support this, a dedicated testing platform was constructed to facilitate precise and repeatable measurements, enabling further refinements in motor control and torque prediction.

\section{Future Work}

Looking ahead, there are several promising directions to expand and refine the current research. The thrust model developed for Harpy, while validated through simulation and experimental testing, can benefit from further characterization. By collecting and analyzing a broader set of empirical data across varying environmental conditions and EDF configurations, the model’s coefficients can be fine-tuned to improve accuracy and predictive capability. This would allow the thrust control system to adapt more effectively to changes in external conditions such as air density, wind interference, or temperature gradients. Additionally, further investigation into torque characterization methods presents an opportunity to deepen our understanding of joint motor performance. Expanding the comparative analysis of spec sheet-based estimates, torque sensor measurements, and back EMF methods could help identify not only the most accurate but also the most reliable technique in terms of repeatability and noise sensitivity.

Another key area of future work involves integrating torque sensor feedback directly into the motor’s closed-loop control system. By leveraging the torque sensor’s real-time data, it becomes possible to implement adaptive control algorithms that dynamically adjust motor output based on measured loads. This integration should first be tested on a single joint in an isolated setup, allowing for detailed tuning and validation before being scaled up to the full robot. Once proven off the robot, it can be incorporated into Harpy’s onboard system, enabling more precise, responsive, and energy-efficient movement, especially during high-load maneuvers such as jumping, recovery, or dynamic terrain negotiation.

%% file: thesis.bbl
\begin{thebibliography}{10}
\providecommand{\url}[1]{#1}
\csname url@samestyle\endcsname
\providecommand{\newblock}{\relax}
\providecommand{\bibinfo}[2]{#2}
\providecommand{\BIBentrySTDinterwordspacing}{\spaceskip=0pt\relax}
\providecommand{\BIBentryALTinterwordstretchfactor}{4}
\providecommand{\BIBentryALTinterwordspacing}{\spaceskip=\fontdimen2\font plus
\BIBentryALTinterwordstretchfactor\fontdimen3\font minus \fontdimen4\font\relax}
\providecommand{\BIBforeignlanguage}[2]{{%
\expandafter\ifx\csname l@#1\endcsname\relax
\typeout{** WARNING: IEEEtran.bst: No hyphenation pattern has been}%
\typeout{** loaded for the language `#1'. Using the pattern for}%
\typeout{** the default language instead.}%
\else
\language=\csname l@#1\endcsname
\fi
#2}}
\providecommand{\BIBdecl}{\relax}
\BIBdecl

\bibitem{haldane_repetitive_2017}
\BIBentryALTinterwordspacing
D.~W. Haldane, J.~K. Yim, and R.~S. Fearing, ``Repetitive extreme-acceleration (14-g) spatial jumping with {Salto}-{1P},'' in \emph{2017 {IEEE}/{RSJ} {International} {Conference} on {Intelligent} {Robots} and {Systems} ({IROS})}, Sep. 2017, pp. 3345--3351, iSSN: 2153-0866. [Online]. Available: \url{https://ieeexplore.ieee.org/document/8206172}
\BIBentrySTDinterwordspacing

\bibitem{zeng_kou-iii_2024}
\BIBentryALTinterwordspacing
X.~Zeng, L.~Huang, G.~Zhang, and Y.~Li, ``Kou-iii: A bipedal robot with quadrotor-assisted locomotion,'' \emph{Mechatronics}, vol. 107, pp. 103--310, 2025. [Online]. Available: \url{https://papers.ssrn.com/sol3/papers.cfm?abstract_id=5056354}
\BIBentrySTDinterwordspacing

\bibitem{kanagamani_pdf_2023}
\BIBentryALTinterwordspacing
N.~Kanagamani, M.~Paulin, T.~Prabu, and M.~Kasim, ``({PDF}) {Design} and {Simulation} {Analysis} of {Spider} {Robot},'' \emph{International Journal on Applications in Information and communication Engineering}, vol.~10, no.~1, pp. 15--17, Oct. 2023. [Online]. Available: \url{https://www.researchgate.net/publication/381519728_Design_and_Simulation_Analysis_of_Spider_Robot}
\BIBentrySTDinterwordspacing

\bibitem{kim_leonardo_2021}
\BIBentryALTinterwordspacing
K.~Kim, P.~Spieler, E.~Lupu, A.~Ramezani, and S.-J. Chung, ``{LEONARDO} {Bipedal} {Robot} {With} {Thrusters},'' \emph{Science Robotics (AAAS)}, vol.~6, no.~59, Oct. 2021. [Online]. Available: \url{https://www.science.org/doi/10.1126/scirobotics.abf8136}
\BIBentrySTDinterwordspacing

\bibitem{noauthor_agility_2018}
\BIBentryALTinterwordspacing
``Agility {Robotics} {Raises} \${8M} to {Develop} {Legged} {Robots},'' 2018. [Online]. Available: \url{https://www.therobotreport.com/agility-robotics-8m-develop-legged-robots/}
\BIBentrySTDinterwordspacing

\bibitem{chae_ballu2_2021}
\BIBentryALTinterwordspacing
H.~Chae, M.~S. Ahn, D.~Noh, H.~Nam, and D.~Hong, ``{BALLU2}: {A} {Safe} and {Affordable} {Buoyancy} {Assisted} {Biped},'' \emph{Frontiers in Robotics and AI}, vol.~8, p. 730323, Dec. 2021. [Online]. Available: \url{https://www.frontiersin.orghttps://www.frontiersin.org/journals/robotics-and-ai/articles/10.3389/frobt.2021.730323/full}
\BIBentrySTDinterwordspacing

\bibitem{grandia_design_2024}
\BIBentryALTinterwordspacing
R.~Grandia, E.~Knoop, M.~A. Hopkins, G.~Wiedebach, J.~Bishop, S.~Pickles, D.~M{\"u}ller, and M.~B{\"a}cher, ``Design and control of a bipedal robotic character,'' \emph{arXiv preprint arXiv:2501.05204}, 2025. [Online]. Available: \url{https://arxiv.org/abs/2501.05204}
\BIBentrySTDinterwordspacing

\bibitem{villani_survey_2018}
\BIBentryALTinterwordspacing
V.~Villani, F.~Pini, F.~Leali, and C.~Secchi, ``Survey on human--robot collaboration in industrial settings: Safety, intuitive interfaces and applications,'' \emph{Mechatronics}, vol.~55, pp. 248--266, 2018. [Online]. Available: \url{https://www.researchgate.net/publication/323531679_Survey_on_human-robot_collaboration_in_industrial_settings_Safety_intuitive_interfaces_and_applications}
\BIBentrySTDinterwordspacing

\bibitem{abourachid_natural_2024}
\BIBentryALTinterwordspacing
A.~Abourachid and V.~Hugel, ``The natural bipeds, birds and humans: An inspiration for bipedal robots,'' in \emph{Biomimetic and Biohybrid Systems: 5th International Conference, Living Machines 2016, Edinburgh, UK, July 19-22, 2016. Proceedings 5}.\hskip 1em plus 0.5em minus 0.4em\relax Springer, 2016, pp. 3--15. [Online]. Available: \url{https://link.springer.com/chapter/10.1007/978-3-319-42417-0_1}
\BIBentrySTDinterwordspacing

\bibitem{honda_what_2018}
\BIBentryALTinterwordspacing
Honda, ``What {We} {Learned} {From} {ASIMO},'' 2018. [Online]. Available: \url{https://global.honda/en/robotics/}
\BIBentrySTDinterwordspacing

\bibitem{dyanmics_atlas_2024}
\BIBentryALTinterwordspacing
B.~Dyanmics, ``Atlas,'' 2024. [Online]. Available: \url{https://bostondynamics.com/atlas/}
\BIBentrySTDinterwordspacing

\bibitem{eth_anymal_2024}
\BIBentryALTinterwordspacing
ETH, ``{ANYmal},'' 2024. [Online]. Available: \url{https://rsl.ethz.ch/robots-media/anymal.html}
\BIBentrySTDinterwordspacing

\bibitem{bryant_development_2020}
\BIBentryALTinterwordspacing
A.~Bryant, ``Development of an empirical propeller thrust model using experimental measurements and computational fluid dynamics,'' Ph.D. dissertation, 2020. [Online]. Available: \url{https://shareok.org/server/api/core/bitstreams/e3302a18-d1a0-4ef4-b15c-1df5b9b8db24/content}
\BIBentrySTDinterwordspacing

\bibitem{ledoux_analysis_2021}
\BIBentryALTinterwordspacing
J.~Ledoux, S.~Riffo, and J.~Salomon, ``Analysis of the {Blade} {Element} {Momentum} {Theory},'' \emph{SIAM Journal on Applied Mathematics}, vol.~81, no.~6, pp. 2596--2621, Jan. 2021, arXiv:2004.11100 [math]. [Online]. Available: \url{http://arxiv.org/abs/2004.11100}
\BIBentrySTDinterwordspacing

\bibitem{lee_empirical_2020}
\BIBentryALTinterwordspacing
U.~H. Lee, C.-W. Pan, and E.~J. Rouse, ``Empirical characterization of a high-performance exterior-rotor type brushless dc motor and drive,'' in \emph{2019 IEEE/RSJ International Conference on Intelligent Robots and Systems (IROS)}.\hskip 1em plus 0.5em minus 0.4em\relax IEEE, 2019, pp. 8018--8025. [Online]. Available: \url{https://www.researchgate.net/publication/372948387_Energy_Efficiency_and_Performance_Evaluation_of_an_Exterior-Rotor_Brushless_DC_Motor_and_Drive_System_across_the_Full_Operating_Range}
\BIBentrySTDinterwordspacing

\bibitem{spong_robot_1989}
M.~Spong, S.~Hutchinson, and M.~Vidasagar, \emph{Robot {Modeling} and Control}, 1st~ed.\hskip 1em plus 0.5em minus 0.4em\relax USA: JOHN WILEY \& SONS, INC., Jan. 1989, vol.~1.

\bibitem{suh_development_2020}
\BIBentryALTinterwordspacing
J.~Suh, J.~Lee, and D.-E. Lee, ``Development and {Application} of {Motor}-{Equipped} {Reaction} {Torque} {Sensor},'' \emph{Applied Sciences}, 2020. [Online]. Available: \url{https://www.mdpi.com/2076-3417/10/24/8770?utm_}
\BIBentrySTDinterwordspacing

\bibitem{fadelli_robofly_2020}
\BIBentryALTinterwordspacing
Y.~M. Chukewad, J.~James, A.~Singh, and S.~Fuller, ``Robofly: An insect-sized robot with simplified fabrication that is capable of flight, ground, and water surface locomotion,'' \emph{IEEE Transactions on Robotics}, vol.~37, no.~6, pp. 2025--2040, 2021. [Online]. Available: \url{https://ieeexplore.ieee.org/abstract/document/9444546}
\BIBentrySTDinterwordspacing

\bibitem{pitroda_capture_2024}
\BIBentryALTinterwordspacing
S.~Pitroda, A.~Bondada, K.~Venkatesh, A.~Salagame, C.~Wang, T.~Liu, B.~Gupta, E.~Sihite, R.~Nemovi, A.~Ramezani, and M.~Gharib, ``Capture {Point} {Control} in {Thruster}-{Assisted} {Bipedal} {Locomotion},'' in \emph{2024 {IEEE} {International} {Conference} on {Advanced} {Intelligent} {Mechatronics} ({AIM})}, Jul. 2024, pp. 1139--1144. [Online]. Available: \url{https://ieeexplore.ieee.org/document/10637139}
\BIBentrySTDinterwordspacing

\bibitem{huang_jet-powered_2017}
\BIBentryALTinterwordspacing
Z.~Huang, B.~Liu, J.~Wei, L.~Qingsheng, J.~Ota, and Y.~Zhang, ``'{Jet}-{Powered}' {Feet} {Enable} {Robot} to {Balance} and {Step} {Over} {Large} {Gaps},'' \emph{2017 IEEE-RAS 17th International Conference on Humanoid Robotics (Humanoids)}, 2017. [Online]. Available: \url{https://ieeexplore.ieee.org/document/8246905}
\BIBentrySTDinterwordspacing

\bibitem{li_design_2022}
\BIBentryALTinterwordspacing
Y.~Li, Y.~Zhou, J.~Huang, Z.~Wang, S.~Zhu, K.~Wu, L.~Zheng, J.~Luo, R.~Cao, Y.~Zhang \emph{et~al.}, ``Design of a flying humanoid robot based on thrust vector control,'' \emph{arXiv preprint arXiv:2108.11557}, 2021. [Online]. Available: \url{https://arxiv.org/abs/2108.11557}
\BIBentrySTDinterwordspacing

\bibitem{iiot_ironcub_2024}
T.~Hui, A.~Paolino, G.~Nava, G.~L'Erario, F.~Di~Natale, F.~Bergonti, F.~Braghin, and D.~Pucci, ``Centroidal aerodynamic modeling and control of flying multibody robots,'' in \emph{2022 International Conference on Robotics and Automation (ICRA)}, 2022, pp. 2017--2023.

\bibitem{fadelli_quadrotors_2025}
\BIBentryALTinterwordspacing
I.~Fadelli, ``Quadrotors support enhanced locomotion in a new bipedal robot,'' Jan. 2025. [Online]. Available: \url{https://techxplore.com/news/2025-01-quadrotors-locomotion-bipedal-robot.html}
\BIBentrySTDinterwordspacing

\bibitem{gunel_modeling_2016}
\BIBentryALTinterwordspacing
O.~Gunel and A.~Ankarali, ``Modeling of {Basic} {Propeller} {Thrust} {Test} {System} and {Thrust} {Control} {Using} pid method,'' \emph{4th International Symposium on Innovative Technologies}, Nov. 2016. [Online]. Available: \url{https://www.isites.info/PastConferences/ISITES2016/ISITES2016/papers/A7-ISITES2016ID240.pdf}
\BIBentrySTDinterwordspacing

\bibitem{lam_propeller_2023}
J.~Lam, A.~Chen, A.~Bennett, and M.~Triantafyllou, ``Propeller {Characterization} {Testing} of a {Blue} {Robotics} {T200} {Thruster},'' in \emph{{OCEANS} 2023 - {Limerick}}, Jun. 2023.

\bibitem{kosa_experimental_2022}
\BIBentryALTinterwordspacing
P.~Kosa, M.~Kisev, L.~Vacho, and L.~Toth, ``Experimental {Measurement} of a {UAV} {Propeller}'s {Thrust},'' in \emph{Tehnicki vjesnik - {Technical} {Gazette} 29}, Slovakia, Jan. 2022. [Online]. Available: \url{https://www.researchgate.net/publication/358165888_Experimental_Measurement_of_a_UAV_Propeller's_Thrust}
\BIBentrySTDinterwordspacing

\bibitem{simmons_efficient_2022}
\BIBentryALTinterwordspacing
B.~M. Simmons, ``Efficient variable-pitch propeller aerodynamic model development for vectored-thrust evtol aircraft,'' in \emph{AIAA Aviation 2022 Forum}, 2022, p. 3817. [Online]. Available: \url{https://www.researchgate.net/publication/361459813_Efficient_Variable-Pitch_Propeller_Aerodynamic_Model_Development_for_Vectored-Thrust_eVTOL_Aircraft}
\BIBentrySTDinterwordspacing

\bibitem{zhang_investigation_2018}
\BIBentryALTinterwordspacing
Q.~Zhang, R.~K. Jaiman, P.~Ma, and J.~Liu, ``Investigation on the performance of a ducted propeller in oblique flow,'' \emph{Journal of Offshore Mechanics and Arctic Engineering}, vol. 142, no.~1, p. 011801, 2020. [Online]. Available: \url{https://arxiv.org/abs/1801.03211?utm_}
\BIBentrySTDinterwordspacing

\bibitem{jakbouski_analysis_2015}
\BIBentryALTinterwordspacing
A.~Jakbouski, A.~Kubaki, and B.~Minorowisz, ``Analysis {Thrust} for {Different} {Kind} of {Propellers},'' \emph{Advances in Intelligent Systems and Computing}, pp. 85--90, 2015. [Online]. Available: \url{https://www.researchgate.net/publication/283842267_Analysis_Thrust_for_Different_Kind_of_Propellers}
\BIBentrySTDinterwordspacing

\bibitem{lipovetsky_linear_2009}
\BIBentryALTinterwordspacing
S.~Lipovetsky, ``Linear regression with special coefficient features attained via parameterization in exponential, logistic, and multinomial–logit forms,'' \emph{Mathematic and Computer Modeling}, vol.~49, no. 7-8, pp. 1427--1435, Apr. 2009. [Online]. Available: \url{https://www.sciencedirect.com/science/article/pii/S0895717709000259}
\BIBentrySTDinterwordspacing

\bibitem{lee_how_2023}
\BIBentryALTinterwordspacing
U.~Lee, T.~Shepherd, K.~Sangbae, A.~De, H.~Su, R.~Gregg, L.~Mooney, and E.~Rouse, ``How to {Model} {Brushless} {Electric} {Motors} for the {Design} of {Lightweight} {Robotic} {Systems},'' \emph{arXiv preprint arXiv:2310.00080}, 2023. [Online]. Available: \url{https://arxiv.org/abs/2310.00080}
\BIBentrySTDinterwordspacing

\bibitem{lin_-depth_2010}
\BIBentryALTinterwordspacing
D.~Lin, P.~Zhou, and Z.~Cendes, ``In-depth study of the torque constant for permanent-magnet machines,'' \emph{IEEE Transactions on Magnetics}, vol.~45, no.~12, pp. 5383--5387, 2009. [Online]. Available: \url{https://www.researchgate.net/publication/224079306_In-Depth_Study_of_the_Torque_Constant_for_Permanent-Magnet_Machines}
\BIBentrySTDinterwordspacing

\bibitem{hbk_getting_2025}
\BIBentryALTinterwordspacing
HBK, ``Getting the {Back}-{Emf} {Voltage} {Constant} of {Permanent} {Magnet} {Motors} with {Just} {One} {Twist} of the {Wrist} using {Gen3i} {Data} {Recorder},'' 2025. [Online]. Available: \url{https://www.hbkworld.com/en/knowledge/resource-center/articles/getting-the-back-emf-voltage-constant-of-permanent-magnet-motors}
\BIBentrySTDinterwordspacing

\bibitem{schmidt_practical_2018}
\BIBentryALTinterwordspacing
A.~Schmidt, T.~Gumpert, S.~Schreiber, and A.~Schaffer, ``Practical {Approach} to {Characterize} {Realistic} {Motor} {Dynamics} for {Robotic} {Applications},'' 2018. [Online]. Available: \url{https://elib.dlr.de/197476/1/AIM22_0133_FI_copyright.pdf}
\BIBentrySTDinterwordspacing

\bibitem{colmenares_characterization_2020}
\BIBentryALTinterwordspacing
D.~Colmenares, R.~Kania, M.~Liu, and M.~Sitti, ``Characterization and thermal management of a dc motor-driven resonant actuator for miniature mobile robots with oscillating limbs,'' \emph{arXiv preprint arXiv:2002.00798}, 2020. [Online]. Available: \url{https://arxiv.org/abs/2002.00798}
\BIBentrySTDinterwordspacing

\bibitem{hwang_virtual_2018}
\BIBentryALTinterwordspacing
Y.~Hwang, Y.~Minami, and M.~Ishikawa, ``Virtual {Torque} {Sensor} for {Low}-{Cost} {RC} {Servo} {Motors} {Based} on {Dynamic} {System} {Identification} {Utilizing} {Parametric} {Constraints},'' \emph{Sensors}, vol.~18, no.~11, p. 3856, Nov. 2018. [Online]. Available: \url{https://www.mdpi.com/1424-8220/18/11/3856}
\BIBentrySTDinterwordspacing

\bibitem{wen_investigation_2020}
\BIBentryALTinterwordspacing
Y.~Wen, G.~Li, Q.~Wang, R.~Tang, Y.~Liu, and H.~Li, ``Investigation on the {Measurement} {Method} for {Output} {Torque} of a {Spherical} {Motor},'' \emph{Electrical, Electronics and Communications Engineering}, 2020. [Online]. Available: \url{https://www.mdpi.com/2076-3417/10/7/2510?utm_}
\BIBentrySTDinterwordspacing

\bibitem{dangol_control_2021}
\BIBentryALTinterwordspacing
P.~Dangol, E.~Sihite, and A.~Ramezani, ``Control of thruster-assisted, bipedal legged locomotion of the harpy robot,'' \emph{Frontiers in Robotics and AI}, vol.~8, p. 770514, 2021. [Online]. Available: \url{https://www.frontiersin.org/journals/robotics-and-ai/articles/10.3389/frobt.2021.770514/full}
\BIBentrySTDinterwordspacing

\bibitem{pitroda_dynamic_2023}
\BIBentryALTinterwordspacing
S.~Pitroda, ``Dynamic multimodal locomotion: A quick overview of hardware and control,'' Master's thesis, Northeastern University, 2023. [Online]. Available: \url{https://repository.library.northeastern.edu/files/neu:4f21z897z}
\BIBentrySTDinterwordspacing

\bibitem{kelly_design_2021}
\BIBentryALTinterwordspacing
P.~Kelly, ``Design of a {Thruster}-{Assisted} {Bipedal} {Robot},'' Master's thesis, Northeastern University, Aug. 2021. [Online]. Available: \url{https://repository.library.northeastern.edu/files/neu%3Abz60w8418?utm_}
\BIBentrySTDinterwordspacing

\bibitem{pitroda_enhanced_2024}
\BIBentryALTinterwordspacing
S.~Pitroda, E.~Sihite, T.~Liu, K.~V. Krishnamurthy, C.~Wang, A.~Salagame, R.~Nemovi, A.~Ramezani, and M.~Gharib, ``Enhanced capture point control using thruster dynamics and qp-based optimization for harpy,'' \emph{arXiv preprint arXiv:2411.17727}, 2024. [Online]. Available: \url{https://arxiv.org/abs/2411.17727}
\BIBentrySTDinterwordspacing

\bibitem{dangol_feedback_2020}
\BIBentryALTinterwordspacing
P.~Dangol and A.~Ramezani, ``{Feedback design for Harpy: a test bed to inspect thruster-assisted legged locomotion},'' in \emph{Unmanned Systems Technology XXII}, vol. 11425, International Society for Optics and Photonics.\hskip 1em plus 0.5em minus 0.4em\relax SPIE, 2020, p. 1142507. [Online]. Available: \url{https://doi.org/10.1117/12.2558284}
\BIBentrySTDinterwordspacing

\bibitem{dangol_unilateral_2021}
\BIBentryALTinterwordspacing
E.~Sihite, P.~Dangol, and A.~Ramezani, ``Unilateral ground contact force regulations in thruster-assisted legged locomotion,'' in \emph{2021 IEEE/ASME International Conference on Advanced Intelligent Mechatronics (AIM)}.\hskip 1em plus 0.5em minus 0.4em\relax IEEE, 2021, pp. 389--395. [Online]. Available: \url{https://ieeexplore.ieee.org/document/9517648}
\BIBentrySTDinterwordspacing

\bibitem{dangol_reduced-order-model-based_2021}
\BIBentryALTinterwordspacing
P.~Dangol and A.~Ramezani, ``Reduced-order-model-based feedback design for thruster-assisted legged locomotion,'' in \emph{11th International ESA Conference on Guidance, Navigation \& Control Systems (GNC)}, 2021. [Online]. Available: \url{https://arxiv.org/abs/2105.10082}
\BIBentrySTDinterwordspacing

\bibitem{sihite_posture_2024}
\BIBentryALTinterwordspacing
E.~Sihite, S.~Pitroda, T.~Liu, C.~Wang, K.~V. Krishnamurthy, A.~Salagame, R.~Nemovi, A.~Ramezani, and M.~Gharib, ``Posture manipulation of thruster-enhanced bipedal robot performing dynamic wall-jumping using model predictive control,'' in \emph{2024 {IEEE}-{RAS} 23rd {International} {Conference} on {Humanoid} {Robots} ({Humanoids})}, Nov. 2024, pp. 491--496. [Online]. Available: \url{https://ieeexplore.ieee.org/abstract/document/10769831}
\BIBentrySTDinterwordspacing

\bibitem{liang_rough-terrain_2021}
K.~Liang, E.~Sihite, P.~Dangol, A.~Lessieur, and A.~Ramezani, ``Rough-{Terrain} {Locomotion} and {Unilateral} {Contact} {Force} {Regulations} {With} a {Multi}-{Modal} {Legged} {Robot},'' in \emph{2021 {American} {Control} {Conference} ({ACC})}, May 2021, pp. 1762--1769.

\bibitem{de_oliveira_thruster-assisted_2020}
A.~C.~B. de~Oliveira and A.~Ramezani, ``Thruster-assisted {Center} {Manifold} {Shaping} in {Bipedal} {Legged} {Locomotion},'' in \emph{2020 {IEEE}/{ASME} {International} {Conference} on {Advanced} {Intelligent} {Mechatronics} ({AIM})}, Jul. 2020, pp. 508--513.

\bibitem{dhole_feedback_nodate}
\BIBentryALTinterwordspacing
A.~S. Dhole, ``Feedback design and implementation for integrated posture manipulation and thrust vectoring.'' [Online]. Available: \url{https://repository.library.northeastern.edu/files/neu:ms375p98v}
\BIBentrySTDinterwordspacing

\bibitem{dangol_hzd-based_2021}
P.~Dangol, A.~Lessieur, E.~Sihite, and A.~Ramezani, ``A {HZD}-based {Framework} for the {Real}-time, {Optimization}-free {Enforcement} of {Gait} {Feasibility} {Constraints},'' in \emph{2020 {IEEE}-{RAS} 20th {International} {Conference} on {Humanoid} {Robots} ({Humanoids})}, Jul. 2021, pp. 156--162.

\bibitem{dangol_performance_2020}
P.~Dangol, A.~Ramezani, and N.~Jalili, ``Performance satisfaction in {Midget}, a thruster-assisted bipedal robot,'' in \emph{2020 {American} {Control} {Conference} ({ACC})}, Jul. 2020, pp. 3217--3223.

\bibitem{dangol_towards_2020}
\BIBentryALTinterwordspacing
P.~Dangol and A.~Ramezani, ``Towards thruster-assisted bipedal locomotion for enhanced efficiency and robustness,'' \emph{IFAC-PapersOnLine}, vol.~53, no.~2, pp. 10\,019--10\,024, Jan. 2020. [Online]. Available: \url{https://www.sciencedirect.com/science/article/pii/S2405896320334844}
\BIBentrySTDinterwordspacing

\bibitem{pitroda_conjugate_2024}
\BIBentryALTinterwordspacing
S.~Pitroda, E.~Sihite, T.~Liu, K.~V. Krishnamurthy, C.~Wang, A.~Salagame, R.~Nemovi, A.~Ramezani, and M.~Gharib, ``Conjugate momentum based thruster force estimate in dynamic multimodal robot,'' Nov. 2024, issue: arXiv:2411.14596 arXiv: 2411.14596 [cs]. [Online]. Available: \url{http://arxiv.org/abs/2411.14596}
\BIBentrySTDinterwordspacing

\bibitem{pitroda_quadratic_2024}
\BIBentryALTinterwordspacing
S.~Pitroda, E.~Sihite, K.~V. Krishnamurthy, C.~Wang, A.~Salagame, R.~Nemovi, A.~Ramezani, and M.~Gharib, ``Quadratic {Programming} {Optimization} for {Bio}-{Inspired} {Thruster}-{Assisted} {Bipedal} {Locomotion} on {Inclined} {Slopes},'' Nov. 2024, issue: arXiv:2411.12968 arXiv: 2411.12968 [cs]. [Online]. Available: \url{http://arxiv.org/abs/2411.12968}
\BIBentrySTDinterwordspacing

\bibitem{krishnamurthy_narrow-path_2024}
\BIBentryALTinterwordspacing
K.~V. Krishnamurthy, C.~Wang, S.~Pitroda, A.~Salagame, E.~Sihite, R.~Nemovi, A.~Ramezani, and M.~Gharib, ``Narrow-{Path}, {Dynamic} {Walking} {Using} {Integrated} {Posture} {Manipulation} and {Thrust} {Vectoring},'' in \emph{2024 {IEEE} {International} {Conference} on {Advanced} {Intelligent} {Mechatronics} ({AIM})}, Jul. 2024, pp. 898--903. [Online]. Available: \url{https://ieeexplore.ieee.org/document/10637015}
\BIBentrySTDinterwordspacing

\bibitem{sihite_optimization-free_2021}
\BIBentryALTinterwordspacing
E.~Sihite, P.~Dangol, and A.~Ramezani, ``Optimization-free ground contact force constraint satisfaction in quadrupedal locomotion,'' in \emph{2021 60th IEEE Conference on Decision and Control (CDC)}.\hskip 1em plus 0.5em minus 0.4em\relax IEEE, 2021, pp. 713--719. [Online]. Available: \url{https://ieeexplore.ieee.org/abstract/document/9683155}
\BIBentrySTDinterwordspacing

\bibitem{ramezani_generative_2021}
\BIBentryALTinterwordspacing
A.~Ramezani, P.~Dangol, E.~Sihite, A.~Lessieur, and P.~Kelly, ``Generative design of nu’s husky carbon, a morpho-functional, legged robot,'' in \emph{2021 IEEE International Conference on Robotics and Automation (ICRA)}.\hskip 1em plus 0.5em minus 0.4em\relax IEEE, 2021, pp. 4040--4046. [Online]. Available: \url{https://ieeexplore.ieee.org/abstract/document/9561196}
\BIBentrySTDinterwordspacing

\bibitem{noauthor_progress_nodate}
\BIBentryALTinterwordspacing
A.~Salagame, S.~Manjikian, C.~Wang, K.~V. Krishnamurthy, S.~Pitroda, B.~Gupta, T.~Jacob, B.~Mottis, E.~Sihite, M.~Ramezani \emph{et~al.}, ``A letter on progress made on husky carbon: A legged-aerial, multi-modal platform,'' \emph{arXiv preprint arXiv:2207.12254}, 2022. [Online]. Available: \url{https://www.researchgate.net/publication/362248164_A_Letter_on_Progress_Made_on_Husky_Carbon_A_Legged-Aerial_Multi-modal_Platform}
\BIBentrySTDinterwordspacing

\bibitem{salagame_quadrupedal_2023}
\BIBentryALTinterwordspacing
A.~Salagame, M.~Gianello, C.~Wang, K.~Venkatesh, S.~Pitroda, R.~Rajput, E.~Sihite, M.~Leeser, and A.~Ramezani, ``Quadrupedal locomotion control on inclined surfaces using collocation method,'' in \emph{2024 American Control Conference (ACC)}.\hskip 1em plus 0.5em minus 0.4em\relax IEEE, 2024, pp. 2838--2843. [Online]. Available: \url{https://ieeexplore.ieee.org/document/10644231}
\BIBentrySTDinterwordspacing

\bibitem{krishnamurthy_towards_2023}
\BIBentryALTinterwordspacing
K.~Krishnamurthy, ``Towards {Dynamic} {Narrow} {Path} {Walking} on {NU}'s {Husky},'' Master's thesis, Northeastern University, Dec. 2023. [Online]. Available: \url{https://arxiv.org/abs/2312.12586?utm_}
\BIBentrySTDinterwordspacing

\bibitem{krishnamurthy_enabling_2024}
\BIBentryALTinterwordspacing
K.~V. Krishnamurthy, E.~Sihite, C.~Wang, S.~Pitroda, A.~Salagame, A.~Ramezani, and M.~Gharib, ``Enabling steep slope walking on {Husky} using reduced order modeling and quadratic programming,'' Nov. 2024, issue: arXiv:2411.11788 arXiv: 2411.11788 [cs]. [Online]. Available: \url{http://arxiv.org/abs/2411.11788}
\BIBentrySTDinterwordspacing

\bibitem{wang_legged_2023}
\BIBentryALTinterwordspacing
C.~Wang, ``Legged walking on inclined surfaces,'' Master's thesis, Northeastern University, 2023. [Online]. Available: \url{https://repository.library.northeastern.edu/files/neu:4f197h44r/fulltext.pdf}
\BIBentrySTDinterwordspacing

\bibitem{sihite_efficient_2022}
E.~Sihite, B.~Mottis, P.~Ghanem, A.~Ramezani, and M.~Gharib, ``Efficient {Path} {Planning} and {Tracking} for {Multi}-{Modal} {Legged}-{Aerial} {Locomotion} {Using} {Integrated} {Probabilistic} {Road} {Maps} ({PRM}) and {Reference} {Governors} ({RG}),'' in \emph{2022 {IEEE} 61st {Conference} on {Decision} and {Control} ({CDC})}, Dec. 2022, pp. 764--770.

\bibitem{sihite_multi-modal_2023}
\BIBentryALTinterwordspacing
E.~Sihite, A.~Kalantari, R.~Nemovi, A.~Ramezani, and M.~Gharib, ``Multi-{Modal} {Mobility} {Morphobot} ({M4}) with appendage repurposing for locomotion plasticity enhancement,'' \emph{Nature Communications}, vol.~14, no.~1, pp. 1--15, Jun. 2023. [Online]. Available: \url{https://www.nature.com/articles/s41467-023-39018-y}
\BIBentrySTDinterwordspacing

\bibitem{sihite_demonstrating_nodate}
\BIBentryALTinterwordspacing
E.~Sihite, F.~Slezak, I.~Mandralis, A.~Salagame, M.~Ramezani, A.~Kalantari, A.~Ramezani, and M.~Gharib, ``Demonstrating autonomous 3d path planning on a novel scalable ugv-uav morphing robot,'' in \emph{2023 IEEE/RSJ International Conference on Intelligent Robots and Systems (IROS)}.\hskip 1em plus 0.5em minus 0.4em\relax IEEE, 2023, pp. 3064--3069. [Online]. Available: \url{https://www.researchgate.net/publication/372827650_Demonstrating_Autonomous_3D_Path_Planning_on_a_Novel_Scalable_UGV-UAV_Morphing_Robot}
\BIBentrySTDinterwordspacing

\bibitem{krishnamurthy_thruster-assisted_2024}
\BIBentryALTinterwordspacing
K.~Krishnamurthy, C.~Wang, S.~Pitroda, A.~Salagame, E.~Sihite, R.~Nemovi, A.~Ramezani, and M.~Gharib, ``Thruster-{Assisted} {Incline} {Walking},'' \emph{arXiv preprint arXiv:2406.13118}, 2024. [Online]. Available: \url{https://arxiv.org/abs/2406.13118}
\BIBentrySTDinterwordspacing

\bibitem{salagame_letter_2022}
\BIBentryALTinterwordspacing
A.~Salagame, S.~Manjikian, C.~Wang, K.~V. Krishnamurthy, S.~Pitroda, B.~Gupta, T.~Jacob, B.~Mottis, E.~Sihite, M.~Ramezani, and A.~Ramezani, ``A {Letter} on {Progress} {Made} on {Husky} {Carbon}: {A} {Legged}-{Aerial}, {Multi}-modal {Platform},'' Jul. 2022, issue: arXiv:2207.12254 arXiv: 2207.12254 [cs, eess]. [Online]. Available: \url{http://arxiv.org/abs/2207.12254}
\BIBentrySTDinterwordspacing

\bibitem{gherold_self-supervised_2024-1}
\BIBentryALTinterwordspacing
V.~Gherold, I.~Mandralis, E.~Sihite, A.~Salagame, A.~Ramezani, and M.~Gharib, ``Self-supervised cost of transport estimation for multimodal path planning,'' Dec. 2024, issue: arXiv:2412.06101 arXiv: 2412.06101 [cs]. [Online]. Available: \url{http://arxiv.org/abs/2412.06101}
\BIBentrySTDinterwordspacing

\bibitem{mandralis_minimum_2023}
\BIBentryALTinterwordspacing
I.~Mandralis, E.~Sihite, A.~Ramezani, and M.~Gharib, ``Minimum {Time} {Trajectory} {Generation} for {Bounding} {Flight}: {Combining} {Posture} {Control} and {Thrust} {Vectoring},'' in \emph{2023 {European} {Control} {Conference} ({ECC})}, Jun. 2023, pp. 1--7. [Online]. Available: \url{https://ieeexplore.ieee.org/document/10178360}
\BIBentrySTDinterwordspacing

\bibitem{noauthor_this_nodate}
\BIBentryALTinterwordspacing
``This {Robot} {Has} {All} the {Moves}—{Eight}, to be {Precise} - {IEEE} {Spectrum}.'' [Online]. Available: \url{https://spectrum.ieee.org/robot-animals}
\BIBentrySTDinterwordspacing

\bibitem{sihite_multi-modal_2023-1}
\BIBentryALTinterwordspacing
E.~Sihite, A.~Kalantari, R.~Nemovi, A.~Ramezani, and M.~Gharib, ``Multi-{Modal} {Mobility} {Morphobot} ({M4}) with appendage repurposing for locomotion plasticity enhancement,'' \emph{Nat Commun}, vol.~14, no.~1, p. 3323, Jun. 2023. [Online]. Available: \url{https://www.nature.com/articles/s41467-023-39018-y}
\BIBentrySTDinterwordspacing

\bibitem{gherold_self-supervised_2024}
\BIBentryALTinterwordspacing
V.~Gherold, I.~Mandralis, E.~Sihite, A.~Salagame, A.~Ramezani, and M.~Gharib, ``Self-supervised cost of transport estimation for multimodal path planning,'' Dec. 2024, arXiv:2412.06101 [cs]. [Online]. Available: \url{http://arxiv.org/abs/2412.06101}
\BIBentrySTDinterwordspacing

\bibitem{sihite_dynamic_2024}
\BIBentryALTinterwordspacing
E.~Sihite, A.~Ramezani, and M.~Gharib, ``Dynamic modeling of wing-assisted inclined running with a morphing multi-modal robot,'' in \emph{2024 {IEEE} {International} {Conference} on {Robotics} and {Automation} ({ICRA})}, May 2024, pp. 2339--2345. [Online]. Available: \url{https://ieeexplore.ieee.org/document/10610678}
\BIBentrySTDinterwordspacing

\bibitem{circiu_pdf_2024}
\BIBentryALTinterwordspacing
I.~Circiu, ``({PDF}) {Theoretical} and practical aspects of the {Coandă} effect applied in aeronautics,'' in \emph{{MATEC} {Web} of {Conferences}}, Oct. 2024. [Online]. Available: \url{https://www.researchgate.net/publication/335334468_Theoretical_and_practical_aspects_of_the_Coanda_effect_applied_in_aeronautics}
\BIBentrySTDinterwordspacing

\bibitem{schubeler_schubeler_2025}
\BIBentryALTinterwordspacing
Schubeler, ``Schubeler,'' 2025. [Online]. Available: \url{https://www.schubeler.com/home/?v=0b3b97fa6688}
\BIBentrySTDinterwordspacing

\bibitem{ab_marine_ab_2025}
\BIBentryALTinterwordspacing
A.~Marine, ``{AB} {Marine},'' 2025. [Online]. Available: \url{https://ab-marine.com}
\BIBentrySTDinterwordspacing

\end{thebibliography}
